\documentclass[10pt,twocolumn,letterpaper]{article}

\usepackage{iccv}
\usepackage{times}
\usepackage{epsfig}
\usepackage{graphicx}
\usepackage{amsmath}
\usepackage{amssymb}
\usepackage{iccv}
\makeatletter
\@namedef{ver@everyshi.sty}{}
\makeatother
\usepackage{tikz}
\usepackage{pgfplots}
\usetikzlibrary{shapes,arrows}
\usetikzlibrary{decorations.text}
\usetikzlibrary{arrows,decorations.pathmorphing,backgrounds,positioning,fit,matrix}
\usepackage{times}
\usepackage{epsfig}
\usepackage{amsmath}
\usepackage{amssymb}
\usepackage{float}
\usepackage{graphicx}
\usepackage{commath}
\usepackage{tabularx}
\usepackage{lscape}
\usepackage{authblk}

\usepackage{subcaption}

\usepackage[pagebackref=true,breaklinks=true,letterpaper=true,colorlinks,bookmarks=false]{hyperref}

\iccvfinalcopy 

\def\iccvPaperID{****} 

\ificcvfinal\fi

\begin{document}

\title{I Bet You Are Wrong:\\ Gambling Adversarial Networks for Structured Semantic Segmentation}

\author[1,2]{Laurens Samson}
\author[1]{Nanne van Noord}
\author[2]{Olaf Booij}
\author[2]{Michael Hofmann}
\author[1]{Efstratios Gavves}
\author[2]{Mohsen Ghafoorian}
\affil[1]{University of Amsterdam, Amsterdam, Netherlands \authorcr Email: {\tt \{n.j.e.vannoord, e.gavves\}@uva.nl}\vspace{1.5ex}}
\affil[2]{TomTom, Amsterdam, Netherlands \authorcr Email: {\tt firstname.lastname@tomtom.com} \vspace{-2ex}}

\maketitle
\ificcvfinal\thispagestyle{empty}\fi

\begin{abstract}
   Adversarial training has been recently employed for realizing structured semantic segmentation, in which the aim is to preserve higher-level scene structural consistencies in dense predictions. However, as we show, value-based discrimination between the predictions from the segmentation network and ground-truth annotations can hinder the training process from learning to improve structural qualities as well as disabling the network from properly expressing uncertainties.
   
   In this paper, we rethink adversarial training for semantic segmentation and propose to formulate the fake/real discrimination framework with a correct/incorrect training objective. More specifically, we replace the discriminator with a ``gambler'' network that learns to spot and distribute its budget in areas where the predictions are clearly wrong, while the segmenter network tries to leave no clear clues for the gambler where to bet. Empirical evaluation on two road-scene semantic segmentation tasks shows that not only does the proposed method re-enable expressing uncertainties, it also improves pixel-wise and structure-based metrics.
   
\end{abstract}

\section{Introduction}
In the past years, deep neural networks have obtained substantial success in various visual recognition tasks including semantic segmentation~\cite{garcia2017review,ghosh2019understanding}. Despite the success of the frequently used (fully) convolutional neural networks~\cite{long2015fully} on semantic segmentation, they lack a built-in mechanism to enforce global structural qualities. For instance, if the task is to detect a single longest line among several linear structures in the image, then a CNN is probably not able to properly handle such global consistency and will likely give responses on other candidate structures. This stems from the fact that even though close-by pixels share a fair amount of receptive field, there is no designated mechanism to explicitly condition the prediction at a specific location on the predictions made at other related (close-by or far) locations, when training with a pixel-level loss.
\begin{figure*}[!t]
\centering
\begin{subfigure}[t]{0.195\linewidth}
    \includegraphics[width=\linewidth, trim={12cm 0cm 0cm 2cm},clip]{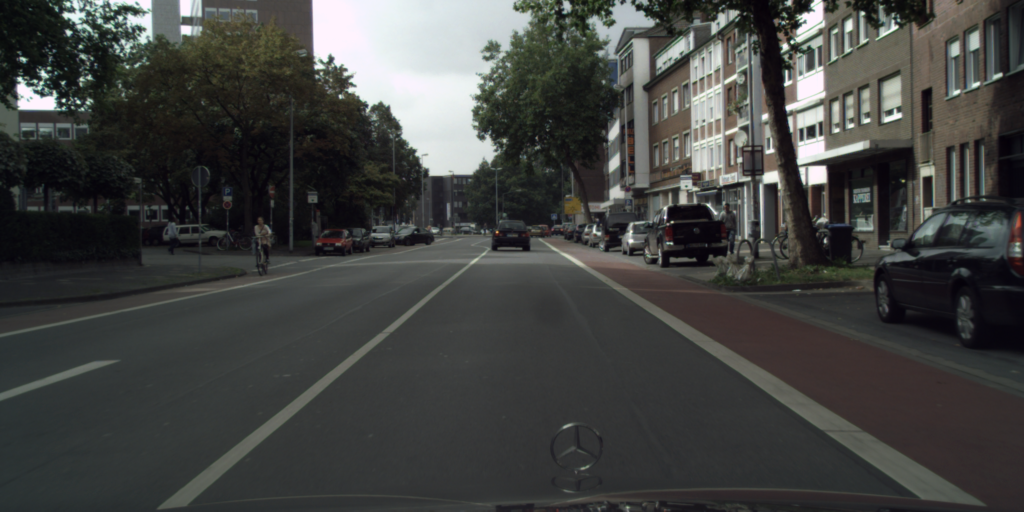}
\end{subfigure}
\hfill
\begin{subfigure}[t]{0.195\linewidth}
    \includegraphics[width=\linewidth, trim={12cm 0cm 0cm 2cm},clip]{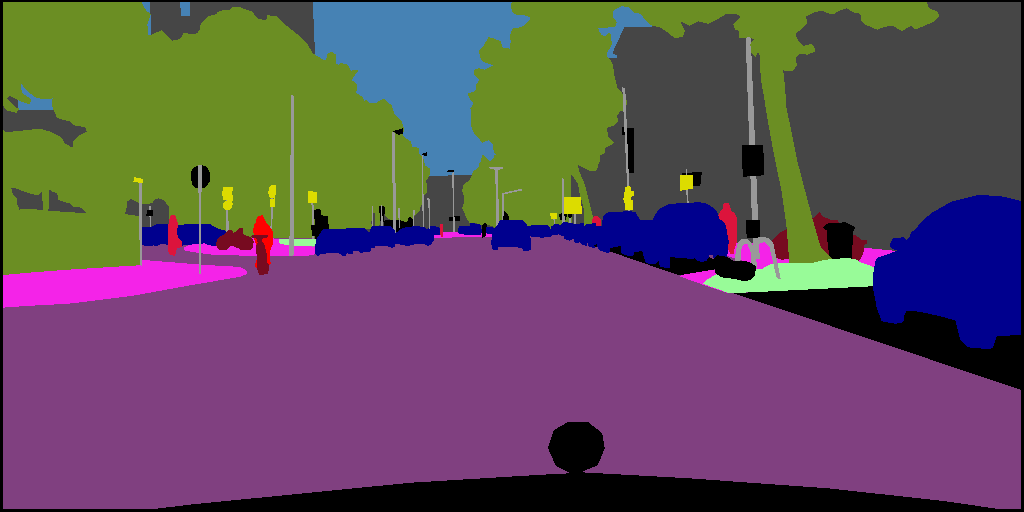}
\end{subfigure}
\hfill
\begin{subfigure}[t]{0.195\linewidth}
    \includegraphics[width=\linewidth, trim={12cm 0cm 0cm 2cm},clip]{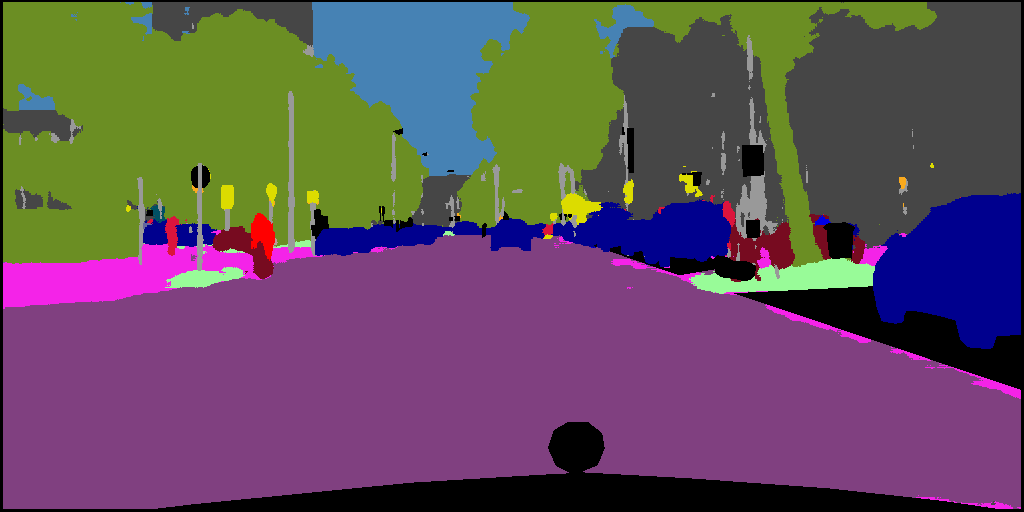}
\end{subfigure}
\hfill
\begin{subfigure}[t]{0.195\linewidth}
    \includegraphics[width=\linewidth, trim={12cm 0cm 0cm 2cm},clip]{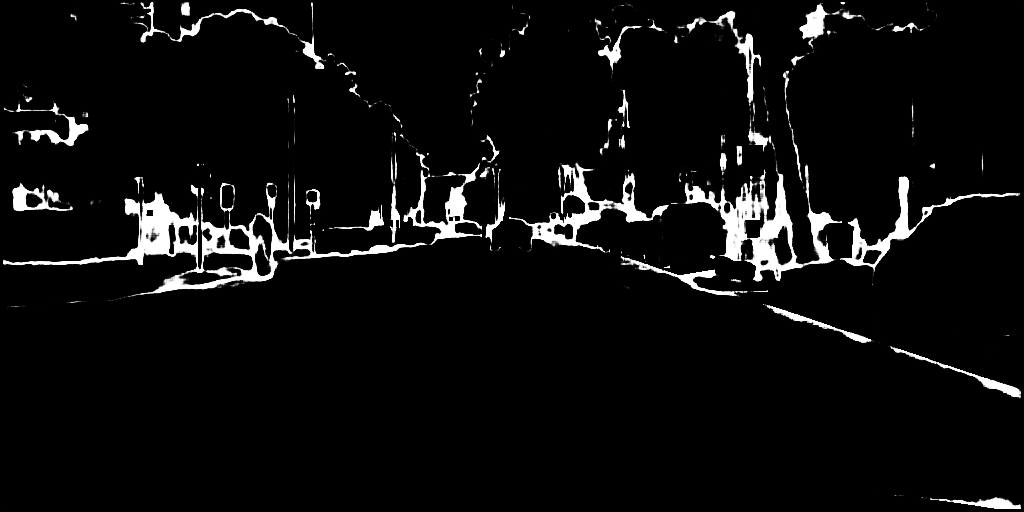}
\end{subfigure}
\hfill
\begin{subfigure}[t]{0.195\linewidth}
    \includegraphics[width=\linewidth, trim={12cm 0cm 0cm 2cm},clip]{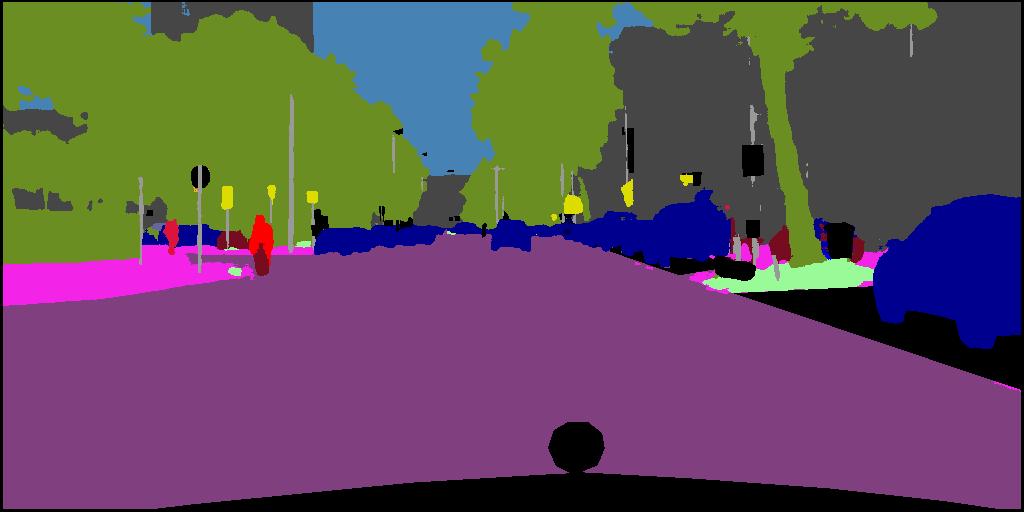}
\end{subfigure}
\caption{From left to right: sample image from Cityscapes~\cite{cordts2016cityscapes}, corresponding ground-truth image, predictions from U-Net trained with cross-entropy loss, betting map from the gambler network, predictions from the gambling adversarial nets. Notice e.g. spotted and resolved artefact in predictions from the cross-entropy trained U-Net in bottom right and right side of the road. Best visible zoomed-in on a screen.}
\label{fig:first_page_image}
\end{figure*}
To better preserve structural quality in semantic segmentation, several methods incorporate graphical models such as conditional random fields (CRF) ~\cite{krahenbuhl2011efficient, zheng2015conditional, tsogkas2015deep}, or use specific topology targeted engineered loss terms~\cite{bentaieb2016topology,oktay2017anatomically}. More recently, adversarial training~\cite{goodfellow2014generative} schemes are being explored~\cite{luc2016semantic,isola2017image,ghafoorian2018gan}, where a discriminator network learns to distinguish the distributions of the network-provided dense predictions (fake) and ground-truth labels (real), which directly encourages better inter-pixel consistencies in a learnable fashion. However, as we will show, the visual clues that the discriminator uses to distinguish the fake and real distributions are not always high-level geometrical properties.  For instance, a discriminator might be able to leverage the prediction values to contrast the fuzzy fake predictions with the crisp zero/one real prediction values to achieve an almost perfect discrimination accuracy. 

Such value-based discrimination results in two undesirable consequences: 1) The segmentation network (``segmenter'') is forced to push its predictions toward zeros and ones and pretend to be confident to mimic such a low-level property of real annotations. This prevents the network from expressing uncertainties. 2) In practice, the softmax probability vectors can not get to exact zeros/ones that requires infinitely large logits. This leaves a permanent possibility for the discriminator to scrutinize the small - but still remaining- value gap between the two distributions, making it needless to learn the more complicated geometrical discrepancies. This hinders such adversarial training procedures to reach their full potential in learning the scene structure.

The value-based discrimination inherently stems from the fake/real discrimination scheme employed in adversarial structured semantic segmentation. Therefore, we aim to study a surrogate adversarial training scheme that still models the higher level prediction consistencies, but is not trained to directly contrast the real and fake distributions. In particular, we replace the discriminator with a ``gambler'' network, that given the predictions of the segmenter and a limited budget, learns to spot and invest in areas where the predictions of the network are likely wrong. Put another way, we reformulate the fake/real discrimination problem into a correct/incorrect distinction task. This prevents the segmenter network from faking certainty, since a wrong confident prediction caught by the gambler, highly penalizes the segmenter. See Figures \ref{fig:first_page_image} and \ref{fig:adv_loss_expl} for getting an overview.\\ \\
Following are the main contributions of the paper:
\begin{itemize}
    \item We propose gambling adversarial networks as a novel adversarial training scheme for structured semantic segmentation.
    \item We show that the proposed method resolves the usual adversarial semantic segmentation training issue with faking confidence.
    \item We demonstrate that this reformulation in the adversarial training improves the semantic segmentation quality over the baselines, both in pixel-wise and structural metrics on two semantic segmentation datasets, namely the Cityscapes~\cite{cordts2016cityscapes} and Camvid~\cite{brostow2009semantic} datasets.
\end{itemize}

\tikzstyle{block} = [draw, rectangle, 
    minimum height=2em, minimum width=4em, node distance=2cm]
\tikzstyle{sum} = [draw, circle, minimum size=1cm, node distance=2cm]
\tikzstyle{dia} = [diamond, draw, text badly centered, inner sep=2t]
\tikzstyle{square} = [draw, rectangle, 
    minimum height=2em, minimum width=2em, node distance=2cm]
\tikzstyle{input} = [coordinate]
\tikzstyle{output} = [coordinate]
\tikzstyle{pinstyle} = [pin edge={to-,thin,black}]
\begin{figure*}[!t]
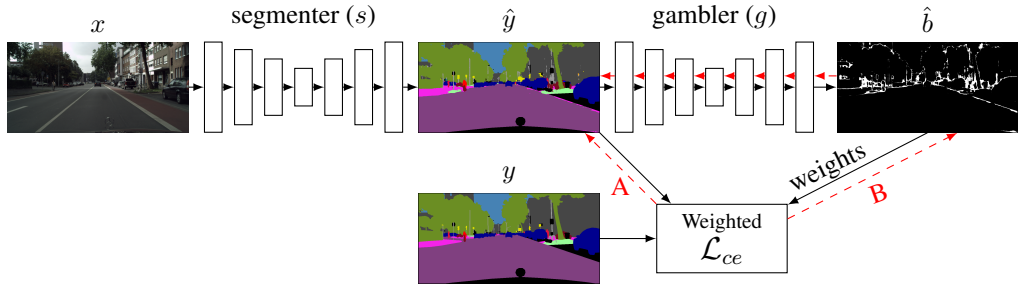

\centering
\begin{tikzpicture}[auto, node distance=1.4cm,>=latex']
    \node[inner sep=0pt, name=x, label={[yshift=-0cm]$x$}](x) {\includegraphics[height=1.2cm ]{Figures/27_rgb.png}};
    \node[draw, rectangle, minimum height=1.2cm, minimum width = 0.05cm, right = 0.2cm of x, name=enc1]{};

    \node[draw, rectangle, node distance=0.4cm, minimum height=1.0cm, minimum width = 0.01cm, right of=enc1, name=enc2]{};
    \node[draw, rectangle, node distance=0.4cm, minimum height=0.75cm, minimum width = 0.01cm, right of=enc2, name=enc3]{};
    \node[draw, rectangle, node distance=0.4cm, minimum height=0.5cm, minimum width = 0.01cm, right of=enc3, name=enc4, label={[label distance=0.4cm]segmenter ($s$)}]{};
    \node[draw, rectangle, node distance=0.4cm, minimum height=0.75cm, minimum width = 0.01cm, right of=enc4, name=dec1]{};
    \node[draw, rectangle, node distance=0.4cm, minimum height=1.0cm, minimum width = 0.01cm, right of=dec1, name=dec2]{};
    \node[draw, rectangle, node distance=0.4cm and 1cm, minimum height=1.2cm, minimum width = 0.01cm, right of=dec2, name=dec3]{};
    \node[inner sep=0pt, name=pred, right= 0.2 cm of dec3,  label={[yshift=-0cm]$\hat{y}$}](pred) {\includegraphics[height=1.2cm ]{Figures/27_seg.png}};
    \node[draw, rectangle, minimum height=1.2cm, minimum width = 0.05cm,  right = 0.2 cm of pred, name=enc2_1]{};
    \node[draw, rectangle, node distance=0.4cm, minimum height=1.0cm, minimum width = 0.01cm, right of=enc2_1, name=enc2_2]{};
    \node[draw, rectangle, node distance=0.4cm, minimum height=0.75cm, minimum width = 0.01cm, right of=enc2_2, name=enc2_3]{};
    \node[draw, rectangle, node distance=0.4cm, minimum height=0.5cm, minimum width = 0.01cm, right of=enc2_3, name=enc2_4, label={[label distance=0.4cm]gambler ($g$)}]{};
    \node[draw, rectangle, node distance=0.4cm, minimum height=0.75cm, minimum width = 0.01cm, right of=enc2_4, name=dec2_1]{};
    \node[draw, rectangle, node distance=0.4cm, minimum height=1.0cm, minimum width = 0.01cm, right of=dec2_1, name=dec2_2]{};
    \node[draw, rectangle, node distance=0.4cm, minimum height=1.2cm, minimum width = 0.01cm, right of=dec2_2, name=dec2_3]{};
    \node[inner sep=0pt, name=bet_map, right = 0.3cm of dec2_3,  label={[yshift=-0cm]$\hat{b}$}](bet_map) {\includegraphics[height=1.2cm ]{Figures/27_betmap.png}};
    \node[inner sep=0pt, name=gt, below = 0.8 cm of  pred, label={[yshift=-0cm]$y$}](gt) {\includegraphics[height=1.2cm ]{Figures/27_GT.png}};
    \node[square, text width = 1.5cm, align=center, right = 0.75cm of gt, name=adv](adv) {\footnotesize{Weighted} \large{$\mathcal{L}_{ce}$}};

    \draw[-latex] (bet_map.south) to node[above,rotate=26, yshift=-3, xshift=-10] {weights} (adv);
    \path [draw=black, -latex] 
            ([xshift=0cm]pred.south east) to node[below,rotate=20] {} ([xshift=0.2cm]adv.north west);
    \draw[-latex] (gt) to node[above,rotate=0] {} (adv);
    \draw[-latex] (x) to node[above,rotate=0] {} (enc1);
    \draw[-latex] (enc1) to node[above,rotate=0] {} (enc2);
    \draw[-latex] (enc2) to node[above,rotate=0] {} (enc3);
    \draw[-latex] (enc3) to node[above,rotate=0] {} (enc4);
    \draw[-latex] (enc4) to node[above,rotate=0] {} (dec1);
    \draw[-latex] (dec1) to node[above,rotate=0] {} (dec2);
    \draw[-latex] (dec2) to node[above,rotate=0] {} (dec3);
    \draw[-latex] (dec3) to node[above,rotate=0] {} (pred);
    
    \draw[-latex] (pred) to node[above,rotate=0] {} (enc2_1);
    \draw[-latex] (enc2_1) to node[above,rotate=0] {} (enc2_2);
    \draw[-latex] (enc2_2) to node[above,rotate=0] {} (enc2_3);
    \draw[-latex] (enc2_3) to node[above,rotate=0] {} (enc2_4);
    \draw[-latex] (enc2_4) to node[above,rotate=0] {} (dec2_1);
    \draw[-latex] (dec2_1) to node[above,rotate=0] {} (dec2_2);
    \draw[-latex] (dec2_2) to node[above,rotate=0] {} (dec2_3);
    \draw[-latex] (dec2_3) to node[above,rotate=0] {} (bet_map);

    

    \path [draw=red, -latex, dashed] 
            ([yshift=-.2cm]adv.north east) to node[below,rotate=20] {\color{red}{B}} ([xshift=.4cm]bet_map.south);
    \path [draw=red, -latex, dashed] 
            ([yshift=.15cm]bet_map.west) to node[above,rotate=0] {} ([yshift=.15cm]dec2_3.east);
    \path [draw=red, -latex, dashed] 
            ([yshift=.15cm]dec2_3.west) to node[above,rotate=0] {} ([yshift=.15cm]dec2_2.east);
    \path [draw=red, -latex, dashed] 
            ([yshift=.15cm]dec2_2.west) to node[above,rotate=0] {} ([yshift=.15cm]dec2_1.east);
\color{red}{    }\path [draw=red, -latex, dashed] 
            ([yshift=.15cm]dec2_1.west) to node[above,rotate=0] {} ([yshift=.15cm]enc2_4.east);
    \path [draw=red, -latex, dashed] 
            ([yshift=.15cm]enc2_4.west) to node[above,rotate=0] {} ([yshift=.15cm]enc2_3.east);
    \path [draw=red, -latex, dashed] 
            ([yshift=.15cm]enc2_3.west) to node[above,rotate=0] {} ([yshift=.15cm]enc2_2.east);
    \path [draw=red, -latex, dashed] 
            ([yshift=.15cm]enc2_2.west) to node[above,rotate=0] {} ([yshift=.15cm]enc2_1.east);
    \path [draw=red, -latex, dashed] 
            ([yshift=.15cm]enc2_1.west) to node[above,rotate=0] {} ([yshift=.15cm]pred.east);
    \path [draw=red, -latex, dashed] 
            ([yshift=-.0cm]adv.north west) to node[xshift=0.cm, yshift=-0.0cm, below,rotate=-0] {\color{red}{A}} ([xshift=-0.2cm]pred.south east);
    
 
\end{tikzpicture}
\caption{An overview of gambling adversarial networks. The solid black arrows indicate the forward pass. The red dashed arrows represent the two gradient flows of the weighted cross-entropy loss. Gradient flow A provides pixel-level feedback independent of other pixel predictions. Gradient flow B, going through the gambler network, enables feedback reflecting the inter-pixel and structural consistency.} 
\label{fig:adv_loss_expl}
\end{figure*}

\section{Related work}

\noindent 
\textbf{Structure-preserving semantic segmentation.}
Several methods have been proposed that use specific loss terms which are targeted at preserving  topologies~\cite{bentaieb2016topology,oktay2017anatomically,mirikharaji2018star} or use graphical models ~\cite{joy2019efficient, krahenbuhl2011efficient, chen2017deeplab, zheng2015conditional, schwing2015fully, namin2015multi, larsson2018revisiting, larsson4learning, jaderberg2014deep, tsogkas2015deep} such as CRFs that model unary, pairwise and/or higher-order potentials, either as a post-processing at the inference time or as integrated training refinement steps. Hand-engineering differentiable targeted loss terms for every desirable structural property is not always feasible in practice. On the other hand, using graphical models either confines the consistency improvements to model low-level features in a local context or imposes high computational costs.

\noindent
\\\textbf{Adversarial semantic segmentation}.
Adversarial training schemes have been extensively employed in the literature to impose structural consistencies for semantic segmentation~\cite{isola2017image,luc2016semantic, hung2018adversarial,dai2017scan,huo2017splenomegaly,kohl2017adversarial,moeskops2017adversarial,yang2017automatic,li2017brain,sadanandan4spheroid,nguyen2017shadow}.
Luc et al.~\cite{luc2016semantic} incorporate a discriminator network trained to distinguish the real labels and network-produced predictions. Involving the segmenter in a minimax game with the discriminator motivates the network to bridge the gap between the two distributions and consequently having higher-level consistencies in predicted labels. 

More recently, it has been shown that the training dynamics of paired image-to-image translation in general~\cite{wang2018high, wang2018perceptual} and adversarial semantic segmentation specifically ~\cite{ghafoorian2018gan, hwang2018adversarial, xue2018segan,zanjani2019deep}, can be improved using paired real/fake embedding losses.

Our method is similar to the aforementioned adversarial formulations in the sense that it also employs a critic network that perceives the whole prediction map, consequently enabling it to model inter-pixel dependencies, and is similarly involved in a minimax game with the segmentation network. Similar to the embedding loss adversarial training, we also leverage the pairing between predictions and ground-truth in our adversarial training, with the difference that we incorporate the ground-truth not as an input but as a supervision for our gambler network. More generally, our method differs in the defined minimax game formulation; the gambler is trained to learn to spot the likely incorrect predictions, while the segmenter is trained to leave as little (structural) clues as possible for the gambler to make an easily profitable investment. 


Luc et al.~\cite{luc2016semantic} also discuss the value-based discrimination issue, which they attempt to alleviate by feeding the discriminator with a Cartesian product of the prediction maps and the input image channels. However, their followed strategy resulted in no improvements as reported. This can be attributed to remaining value-based evidence based on values distribution granularity. For instance, a very tiny response to a first-layer edge detector, in this case, can already signify a fake data sample.

\noindent
\\\textbf{Hard-sample mining}.
Our method is also closely related to the literature on class-imbalance/hard-sample mining. Class-imbalance is another inherent difficulty that needs to be properly tackled when dealing with problems/datasets with imbalanced semantic classes, as is often the case in semantic segmentation. Synthetic minority over-sampling technique (SMOTE)~\cite{chawla2002smote} and Mean/median-frequency balancing~\cite{eigen2015predicting} are common simple strategies that over-sample or scale the loss terms corresponding to the under-represented classes. More recently, focal loss~\cite{lin2017focal} and loss max-pooling~\cite{bulo2017loss} improve over the aforementioned by distinguishing between rarity and difficulty; not all samples from a frequent class are easy and not all samples belonging to infrequent classes are difficult. Therefore, focal loss and loss max-pooling address the more generic problem of hard-sample mining. However, the main issue with both is their inherent limitation dealing with label noise and/or ambiguities in the underlying semantics. We can view our gambling adversarial networks as an adversarially learned version of focal loss; the gambler learns to bet on (i.e. up-weight) the samples that it perceives as more difficult and/or more likely to be wrong in predictions from the segmenter. Such a learned approach can alleviate the label noise problem, as a learned network over noisy labels may be able to generalize beyond the noise level in the training dataset~\cite{natarajan2013learning,ghafoorian2018student} or at least soften the erroneous strong weight-increase for noisy samples. Furthermore, focal loss is derived in a pixel-wise manner and therefore cannot provide structural feedback to the segmentation network. 
\section{Method}
In this section, the proposed method, gambling adversarial networks, is described. First, we present the usual adversarial training formulation for structured semantic segmentation and discuss the potential issues with it. Thereafter, we describe gambling adversarial networks and its reformulation of the former. 

\subsection{Conventional adversarial training}
In the usual adversarial training formulation, the discriminator learns to discriminate the ground-truth (real) from the predictions provided by the network (fake). By involving the segmenter in a minimax game, it is challenged to improve its predictions to provide realistic-looking predictions to fool the discriminator~\cite{luc2016semantic}. In semantic segmentation, such an adversarial training framework is often employed with the aim to improve the higher-level structural qualities, such as connectivity, inter-pixel (local and non-local) consistencies and smoothness. The minimax game is set-up by forming the following loss terms for the discriminator and segmenter:
\begin{multline}
    \mathcal{L}_{d}(x, y; \theta_{s}, \theta_{d}) =  \mathcal{L}_{bce}(d(x, s(x;\theta_{s});\theta_{d}), 0) \\ +   \mathcal{L}_{bce}(d(x, y;\theta_{d}), 1),
\label{eq:dis_loss}
\end{multline}
where $x$ and $y$ are the input image and the corresponding label-map, $s(x;\theta_{s})$ is the segmenter's mapping of the input image $x$ to a dense segmentation map parameterized by $\theta_{s}$, $d(x, y; \theta_{d})$ represents the discriminator operating on segmentations $y$, conditioned on input image $x$ and the binary cross-entropy is defined as $\mathcal{L}_{bce}(\hat{y}, y) = - (y \log \hat{y} + (1-y) \log(1-\hat{y}))$, where $\hat{y}$ and $y$ are the prediction and label respectively.

Typically, the loss function for the segmenter is a combination of low-level (pixel-wise) and high-level (adversarial) loss terms~\cite{isola2017image, luc2016semantic}:

\begin{multline}
    \mathcal{L}_{s}(x, y; \theta_{s}, \theta_{d}) =  \mathcal{L}_{ce}(s(x;\theta_{s}), y) \\  +  \lambda \mathcal{L}_{bce}(d(x, s(x;\theta_{s});\theta_{d}), 1),
\label{eq:gen_loss}
\end{multline}
where $\lambda$ is the importance weighting of the adversarial loss, being the recommended non-saturating reformulation of the original minimax loss term to prevent vanishing gradients~\cite{goodfellow2014generative, fedus2017many}. The pixel-level cross-entropy loss $\mathcal{L}_{ce}$ optimizes all the pixels independently of each other by minimizing $\mathcal{L}_{ce}(\hat{y}, y) = -\frac{1}{wh} \sum_{i, j}^{w, h} \sum_k^c y_{i, j, k} \log \hat{y}_{i, j, k}$, where $w$ and $h$ are the width and the height of image $x$ and $c$ is the number of classes in the dataset. 

Recently, the usual adversarial training for structured semantic segmentation was suggested to be modified~\cite{ghafoorian2018gan, xue2018segan, wang2018perceptual} by replacing the binary cross-entropy loss as the adversarial loss term for the segmenter, with a fake/real paired embedding difference loss, where the embeddings are extracted from the adversarially trained discriminator. To be more specific, the adversarial loss term in Equation~(\ref{eq:gen_loss}) is replaced by the following embedding loss:
\begin{equation}
\begin{aligned}
    &\mathcal{L}_{emb}(x, \hat{y}, y; \theta_{d}) = \norm{d_e(x, \hat{y};\theta_{d}) - d_e(x, y;\theta_{d})}_2,
\end{aligned}
\end{equation}
where the function $d_e(x,y;\theta)$ represents the extracted features from a particular layer in the discriminator. As shown in the EL-GAN method, this could significantly stabilize training ~\cite{ghafoorian2018gan}.

Ideally, the discriminator's decisions are purely based on the structural differences between the real and the fake predictions. However, in semantic segmentation, it is often possible for the discriminator to perfectly distinguish the labels from the predictions based on the values. The output of the segmenter is a softmax vector per pixel, which assigns a probability to every class that ranges between zero and one. In contrast, the values in the ground-truth are either zeros or ones due to the one-hot encoding. Such value-based discrepancy can yield unsatisfactory gradient feedback, since the segmenter might be forced to mimic the one-hot encoding of the ground-truth instead of the global structures. Additionally, the value-based discrimination is a never-ending problem since realizing exact ones and zeros requires infinite large logits, however, in practise, the segmenter always leaves a small value-based gap that can be exploited by the discriminator. Another undesired outcome is the loss of ability to express uncertainties, since all the predictions will converge towards a one-hot representation to bridge the value-based gap between the one-hot labels and probabilistic predictions. 

\subsection{Gambling Adversarial Networks}
To prevent the adversarial network from utilizing the value-based discrepancy, we propose gambling adversarial networks, which focuses solely on improving the structural inconsistencies. Instead of the usual real/fake adversarial training task, we propose to modify the task to learn to distinguish incorrect predictions given the whole prediction map. Different from a discriminator, the critic network (gambler) does not observe the ground-truth labels, but solely the RGB-image in combination with the prediction of the segmentation network (segmenter). Given a limited investment budget, the gambler predicts an image-sized betting map, where high bets indicate pixels that are likely incorrectly classified, given the contextual prediction clues around it. Since the gambler receives the entire prediction, structurally ill-formed predictions, such as non-smoothness, disconnectivities and shape-anomalies are clear visual clues for profitable investments for the gambler. An overview of gambling adversarial networks is provided in Figure \ref{fig:adv_loss_expl}.

Similar to conventional adversarial training, the gambler and segmenter play a minimax game; The gambler maximizes the expected weighted pixel-wise cross-entropy where the weights are determined by its betting map, while the segmenter attempts to improve its predictions such that the gambler does not have clues where to bet:

\begin{equation}
\begin{split}
    &\mathcal{L}_{g}(x, y; \theta_s, \theta_g) = \\
    &-\frac{1}{wh} \sum_{i,j}^{w,h} g(x, s(x;\theta_s);\theta_g)_{i, j} \mathcal{L}_{ce}(s(x;\theta_s)_{i, j}, y_{i, j}),
\end{split}
\end{equation}
where $g(x, s(x;\theta_s);\theta_g)_{i, j}$ is the amount of budget the gambler invests on position ($i, j$).
\\The segmenter network minimizes the opposite:
\begin{equation}
\begin{aligned}
    \mathcal{L}_{s}(x, y;\theta_s, \theta_g) = \mathcal{L}_{ce}(s(x;\theta_s), y) - \mathcal{L}_{g}(x, y; \theta_s, \theta_g).
\end{aligned}
\end{equation}
Similar to conventional adversarial training, the segmentation network optimizes a combination of loss terms: a per-pixel cross-entropy loss and an inter-pixel adversarial loss. 
It should be noted that the gambler can easily maximize this loss by betting infinite amounts on all the pixels. Therefore, it is necessary to limit the budget the gambler can spend. We accommodate this by turning the betting map into a smoothed probability distribution:
\begin{equation}
\begin{aligned}
    g(x, \hat{y}; \theta_g)_{i, j} = \frac{ g_{\sigma}(x, \hat{y};\theta_g)_{i, j} + \beta}{\sum_{k,l}^{w,h} g_{\sigma}(x, \hat{y};\theta_g)_{k, l} + \beta},
\end{aligned}
\end{equation}
where $\beta$ is a smoothing factor and $g_{\sigma}(x, \hat{y};\theta_g)_{i, j}$ represents the sigmoid output of the gambler network for pixel with the indices $i, j$. Smoothing the  betting map regularizes the model to spread its bets over multiple pixels instead of focusing on a single location.  

The adversarial loss causes two different gradient streams for the segmentation network, as shown in Figure \ref{fig:adv_loss_expl}, where the solid black and dashed red arrows indicate the forward pass and backward gradient flows respectively. In the backward pass, the gradient flow A pointing directly towards the prediction provides pixel-wise feedback independent of the other pixel predictions. Meanwhile, the gradient flow B, going through the gambler network, provides feedback reflecting inter-pixel and structural consistencies.

\section{Experimental results}
In this section, we discuss the datasets and metrics for the evaluation of gambling adversarial networks. Thereafter, we describe the different network architectures for the segmenter and gambler networks and provide details for training. Finally, we report the results of our experiments.

\subsection{Experimental setup}
\paragraph{Datasets.} We conduct experiments on two different urban road-scene semantic segmentation datasets, but hypothesize that the method is generic and can be applied to any segmentation dataset. 

\textit{Cityscapes.} The Cityscapes~\cite{cordts2016cityscapes} dataset contains 2975 training images, 500 validation images and 1525 test images with a resolution of 2048 $\times$ 1024 consisting of 19 different classes, such as cars, persons and road signs. For preprocessing of the data, we down-scale the images to 1024 $\times$ 512, perform random flipping and take random crops of 512 $\times$ 512 for training. Furthermore, we perform intensity jittering on the RGB-images.

\textit{Camvid.} The urban scene Camvid~\cite{brostow2009semantic} dataset consists of 429 training images, 101 validation images and 171 test images with a resolution of 960 $\times$ 720. We apply the same data augmentations as described above, except that we do not perform any down-scaling.

\paragraph{Metrics.} In addition to the mean intersection over union (IoU), we also quantify the structural consistency of the segmentation maps. Firstly, we compute the BF-score~\cite{csurka2013good}, which measures whether the contours of objects in the predictions match with the contours of the ground-truth. A point on the contour line is a match if the distance between the ground-truth and prediction lies within a toleration distance $\tau$, which we set to 0.75 \% of the image diagonal as suggested in~\cite{csurka2013good}. Furthermore, we utilize a modified Hausdorff distance to quantitatively measure the structural correctness~\cite{dubuisson1994modified}. We slightly modify the original Hausdorff distance, to prevent it from being overwhelmed by outliers:
\def\Inf{\operatornamewithlimits{inf\vphantom{p}}}

\begin{equation}
\begin{aligned}
 \text{d}_{H}(X,Y)  = \frac{1}{2} \sum  \left\{ \frac{1}{|X|} \right. &\sum_{x \in X} \Inf_{y\in Y}  \text{d}(x, y),  \\   \frac{1}{|Y|} &\sum_{y \in Y} \left. \Inf_{x \in X}\text{d}(x,y)\right\},
 \end{aligned}
\end{equation}
where $X$ and $Y$ are the contours of the predictions and labels from a particular class and $d(x,y)$ is the Euclidean distance. We average the score over all the classes that are present in the prediction and the ground-truth.

\paragraph{Network architectures.}
For comparison, we experiment with two well-known baseline segmentation network architectures. Firstly, a U-Net~\cite{ronneberger2015u} based architecture as implemented in Pix2Pix~\cite{isola2017image}, which is an encoder-decoder structure with skip connections. The encoder consists of nine down-sampling blocks containing a convolutional layer with batch normalization and ReLu. The decoder blocks are the same, except that the convolutions are replaced by transposed convolutions. Furthermore, we conduct experiments with PSPNet~\cite{zhao2017pyramid}, which utilizes a pyramid pooling module to capture more contextual information. Similar to~\cite{zhao2017pyramid}, we utilize an ImageNet pre-trained ResNet-101~\cite{he2016deep} as backbone.

For the gambler network, we utilize the same networks as the segmentation network. When training with the U-Net based architecture, the gambler network is identical except that it contains only six down-sampling blocks. For the PSPNet, the architecture of the gambler and segmenter are identical. For the baseline adversarial methods, we utilize the PatchGAN discriminator from Pix2Pix~\cite{isola2017image}.

\paragraph{Training.}
For training the models, we utilize the Adam optimizer~\cite{kingma2014adam} with a linearly decaying learning rate over time. Similar to the conventional adversarial training, the gambler and segmenter are trained in an alternating fashion where the gambler is frozen when updating the segmenter and vice versa. Furthermore, we learned that as opposed to conventional adversarial training, our network does not required separate pre-training and in general, we observe that the training is less sensitive to hyperparameters. Details of the hyperparameters can be found in the supplementary material.

\subsection{Results}
\begin{figure}[!t]
\vspace{-0.8cm}
  \includegraphics[width=1.1\linewidth]{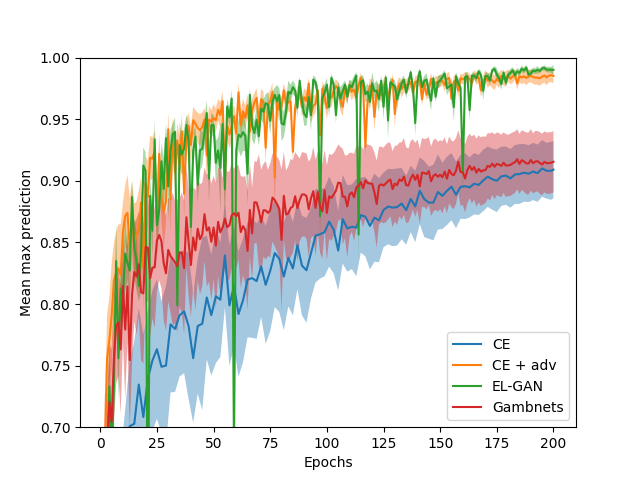}
   \vspace{-0.7cm}\caption{Mean maximum class-likelihoods (mean confidence) over time on the Cityscapes~\cite{cordts2016cityscapes} validation set. Solid central curves and the surrounding shaded area represent the mean and standard deviation respectively. }
\label{fig:mean_max_scratch}
\end{figure}

\paragraph{Confidence expression.}
As discussed before, value-based discrimination encourages the segmentation network to mimic the one-hot vectors of the ground-truth, resulting in loss of ability to express uncertainty. We hypothesize that reformulating the fake/real discrimination in the adversarial training to a correct/incorrect distinction scheme will mitigate the issue. To verify this, the mean and standard deviation of the maximum class-likelihood value in every softmax vector for each pixel is tracked on the validation set over different training epochs and the results are depicted in Figure \ref{fig:mean_max_scratch}. We conducted this experiment with the U-Net based architecture on Cityscapes, but we observed the same phenomena with the other segmentation network and on the other dataset.  One can observe that for both the standard adversarial training and EL-GAN that discriminate the real from the fake predictions, the predictions are converging towards one, with barely any standard deviation. For the gambling adversarial networks, the uncertainty of the predictions is well-preserved. In Table \ref{final_mean_max}, the average mean maximum over the last 10 epochs is shown, which confirms that the gambling adversarial networks maintain the ability to express the uncertainty similar to the cross-entropy model, while the existing adversarial methods attempt to converge to a one-hot vector.

\begin{table}[t]
\begin{tabular}{lll}
\hline
Method                                           & Mean max  \\ \hline
\multicolumn{1}{l|}{Cross-entropy}               & $90.7 \pm 2.3$ \%  \\
\multicolumn{1}{l|}{Cross-entropy + adversarial} & $98.4 \pm 0.5$ \%  \\
\multicolumn{1}{l|}{EL-GAN}                      & $98.9 \pm 0.2$ \%   \\
\hline
\multicolumn{1}{l|}{Gambling nets}               & $91.4 \pm 2.4$ \%  \\
\end{tabular}
\caption{Mean maximum value in every softmax vector on the Cityscapes \cite{csurka2013good} validation set averaged over the last 10 epochs.}
\label{final_mean_max}
\end{table}

\paragraph{U-Net based segmenter.}
First, we compare the baselines with the gambling adversarial networks on the Cityscapes~\cite{cordts2016cityscapes} validation set with the U-Net based architecture. The results in Table \ref{Unet_city} show the gambling adversarial networks perform better on the pixel-wise metric (IoU), but also on the structural metrics. In Table \ref{iou_per_class},  the IoU per class is provided for the same experiments. The gambling adversarial networks perform better on most of the classes. Moreover, performance particularly improves on the classes with finer structures, such as traffic light and person. In Table \ref{bf_per_class}, we report the BF-score per class, where the gambling adversarial networks outperform the other methods on almost all classes. Moreover, similar to the IoU, we observe the most significant improvements on the more fine-grained classes, such as rider and pole. 

In Figure \ref{fig:Unet_city}, one qualitative sample is depicted, in the supplementary material more samples are provided. The adversarial methods resolve some of the artifacts, such as the odd pattern in the car on the left. Moreover, the boundaries of the pedestrians on the sidewalk become more precise. We also provide an example betting map predicted by the gambler, given the predictions from the baseline model trained with cross-entropy in combination with the RGB-image. Note that the gambler bets on the badly shaped building in the prediction and responds to the artifacts in the car. 

\begin{figure*}[!th]
\centering
\captionsetup[subfigure]{labelformat=empty}

\begin{subfigure}[t]{0.32\linewidth}
    \includegraphics[width=\linewidth]{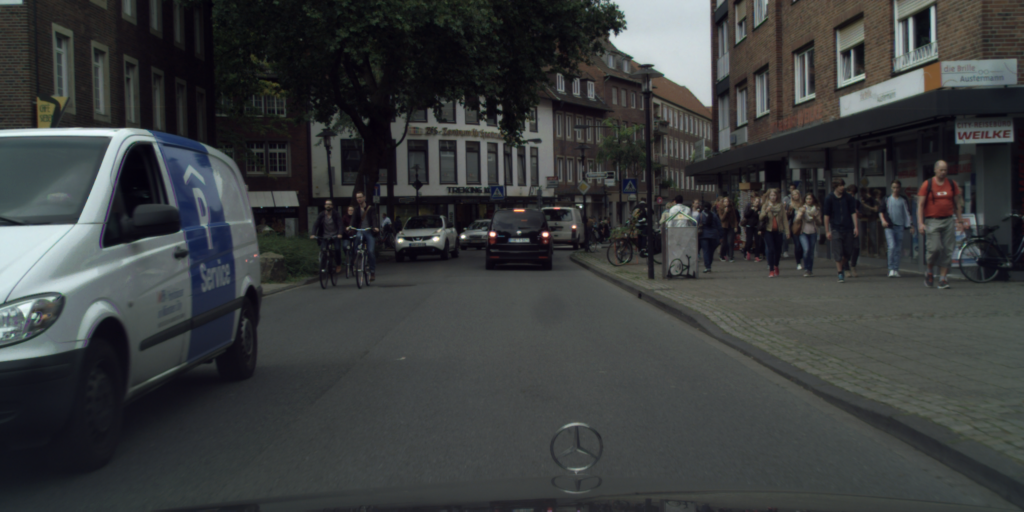}
    \caption{RGB-image}
\end{subfigure}
\hfill
\begin{subfigure}[t]{0.32\linewidth}
    \includegraphics[width=\linewidth]{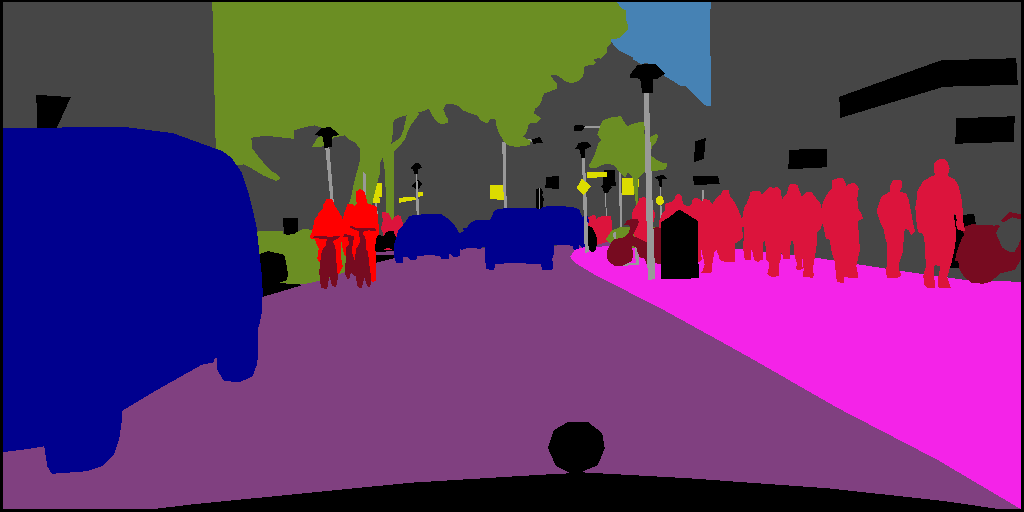}
    \caption{Ground-truth}
\end{subfigure}
\hfill
\begin{subfigure}[t]{0.32\linewidth}
    \includegraphics[width=\linewidth]{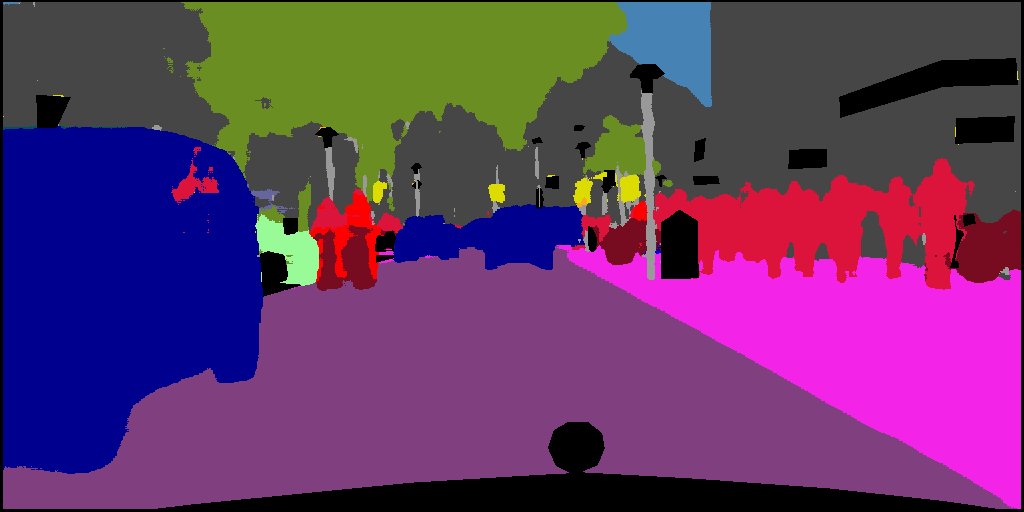}
    \caption{CE}

\end{subfigure}
\hfill
\begin{subfigure}[t]{0.32\linewidth}
    \includegraphics[width=\linewidth]{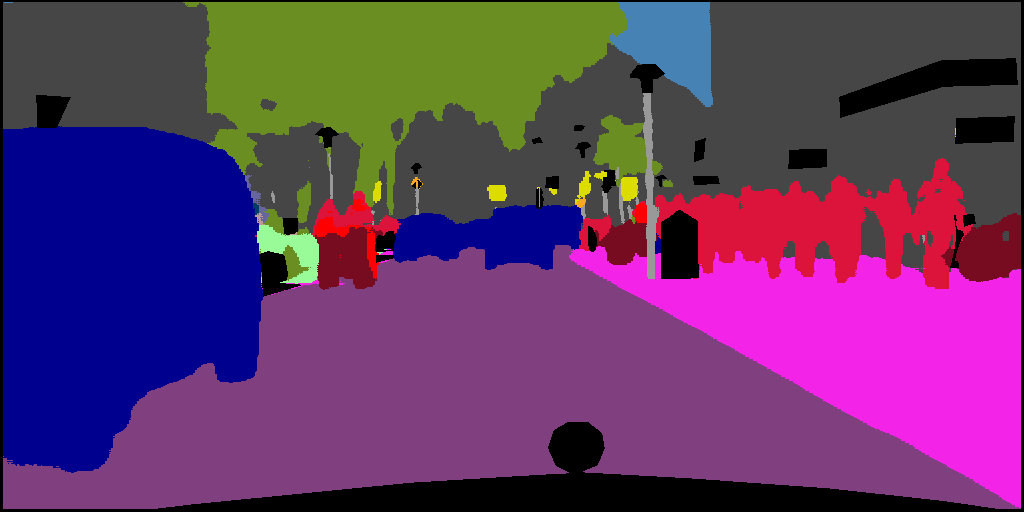}
    \caption{Focal loss}
\end{subfigure}
\hfill
\begin{subfigure}[t]{0.32\linewidth}
    \includegraphics[width=\linewidth]{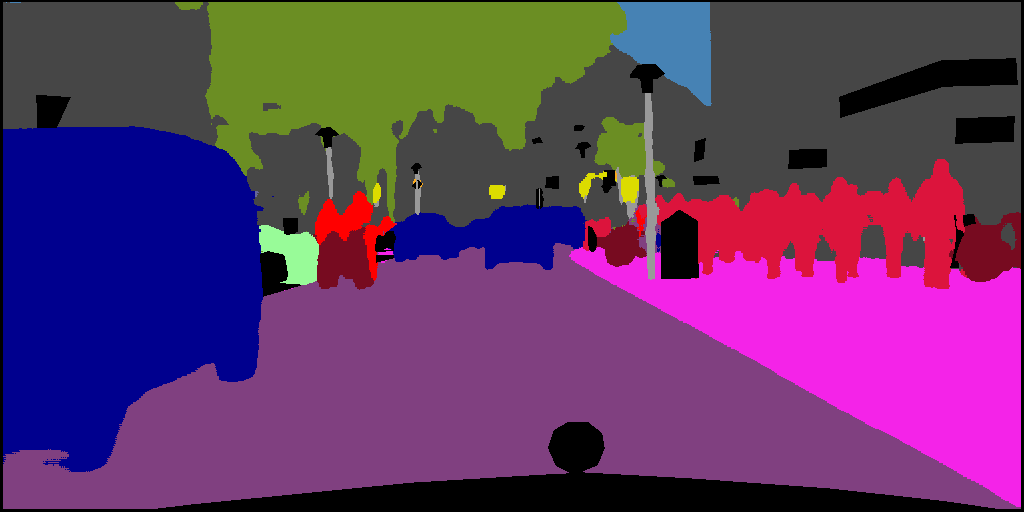}
    \caption{CE + adv}
\end{subfigure}
\hfill
\begin{subfigure}[t]{0.32\linewidth}
    \includegraphics[width=\linewidth]{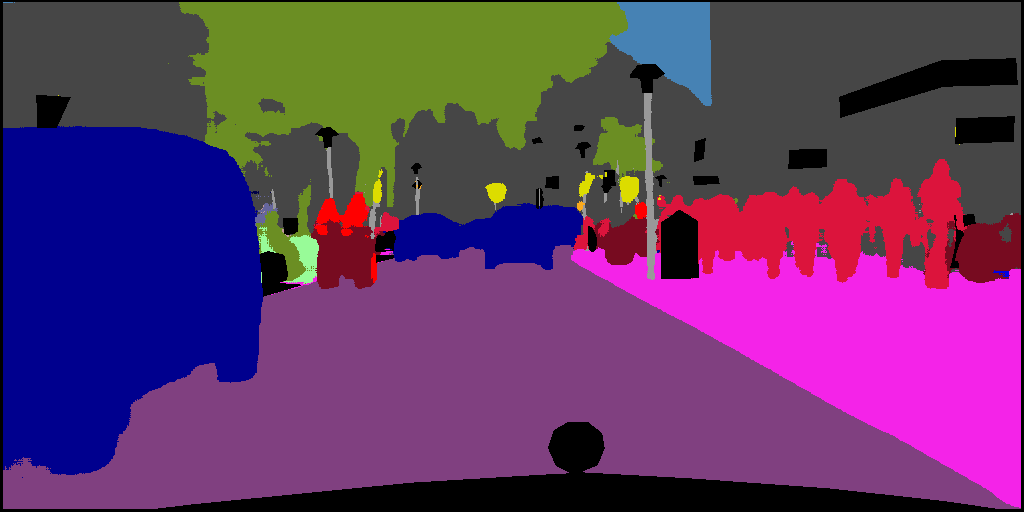}
    \caption{EL-GAN}
\end{subfigure}
\hfill
\begin{subfigure}[t]{0.49\linewidth}
    \includegraphics[width=\linewidth]{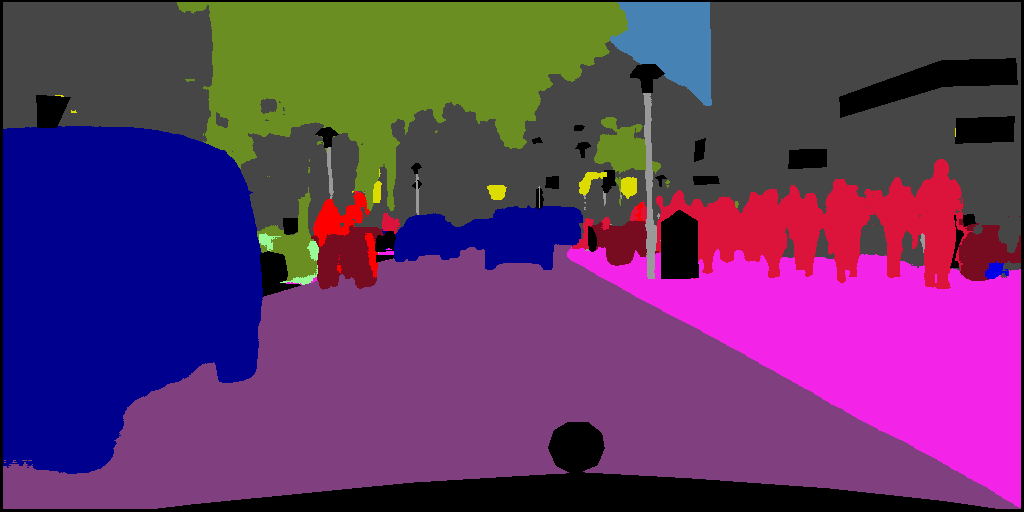}
    \caption{Gambling nets}
\end{subfigure}
\hfill
\begin{subfigure}[t]{0.49\linewidth}
    \includegraphics[width=\linewidth]{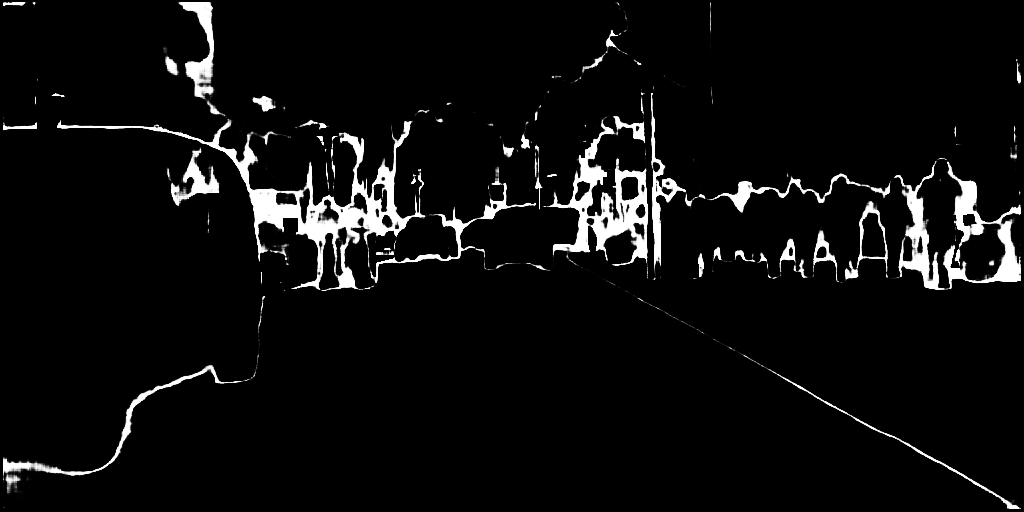}
    \caption{Betting map}
\end{subfigure}
\caption{Qualitative results on Cityscapes~\cite{cordts2016cityscapes} with the U-Net based architecture~\cite{isola2017image}.}
\label{fig:Unet_city}
\end{figure*}

\begin{table}[t]
\begin{tabular}{lllll}
\hline
Method                                           & Mean IoU  & BF-score & Hausdorff  \\ \hline
\multicolumn{1}{l|}{CE}                     & 52.7         & 49.0          & 36.8           \\
\multicolumn{1}{l|}{Focal loss~\cite{lin2017focal}}             & 56.2          & 55.3          & 30.2           \\
\multicolumn{1}{l|}{CE + adv~\cite{luc2016semantic}}               & 56.3         & 57.3          & 31.3           \\
\multicolumn{1}{l|}{EL-GAN~\cite{ghafoorian2018gan}}                      & 55.4         & 54.2          & 31.6  
\\ \hline
\multicolumn{1}{l|}{Gambling nets}            & \textbf{57.9}  & \textbf{58.5} & \textbf{27.6}  \\
     
\end{tabular}
\caption{Results on Cityscapes~\cite{cordts2016cityscapes} with  U-Net based architecture~\cite{isola2017image} as segmentation network.}
\label{Unet_city}
\end{table}

\begin{table*}[]
\small
\resizebox{\textwidth}{!}{
\begin{tabular}{lllllllllllllllllllll}
\hline
Method                             & \small road          & \small swalk        & \small build        & \small wall          & \small fence         & \small pole          & \small tlight        &  \small sign          & \small veg.          & \small ter.       & \small sky           & \small pers        & \small rider         & \small car           & \small truck         & \small bus           & \small train         & \small mbike         & \small bike          & \small mean \\ \hline
\multicolumn{1}{l|}{CE}            & 95.2          & 68.4          & 84.4          & 26.0          & 30.9          & 43.0          & 38.9          & 51.3          & 87.2          & 50.3          & 91.5          & 59.0          & 32.6          & 85.5          & 22.8          & 43.2          & \textbf{19.2} & 15.4          & 57.4         & 57.2\\
\multicolumn{1}{l|}{Focal loss~\cite{lin2017focal}}    & 96.0          & 71.3          & 87.1          & 32.2          & 34.9          & 48.6          & 47.6          & 57.8          & 88.9          & 54.2          & 92.7          & 62.9          & 33.5          & 87.2          & 28.5          & \textbf{47.5} & 18.3          & 19.3          & 60.0       & 56.2   \\
\multicolumn{1}{l|}{CE + adv~\cite{luc2016semantic}}      & 95.9          & 72.7          & 83.5          & 28.9          & 35.2          & 49.8          & 47.8          & 59.3          & 89.0          & 54.8          & 92.3          & 66.4          & 38.4          & 87.2          & 27.8          & 41.4          & 15.3          & 20.3          & 62.5    &  56.3 \\
\multicolumn{1}{l|}{EL-GAN~\cite{ghafoorian2018gan}}        & 96.1          & 71.1          & 86.8          & \textbf{33.5} & 37.0          & 48.7          & 46.6          & 57.3          & 88.9          & 53.6          & 92.9          & 62.6          & 34.4          & 87.1          & 26.0          & 38.3          & 16.3          & 17.8          & 58.9      & 55.4    \\ \hline
\multicolumn{1}{l|}{Gambling} & \textbf{96.3} & \textbf{73.0} & \textbf{87.6} & 33.4          & \textbf{39.1} & \textbf{52.9} & \textbf{51.3} & \textbf{61.9} & \textbf{89.7} & \textbf{55.8} & \textbf{93.1} & \textbf{68.1} & \textbf{38.9} & \textbf{88.7} & \textbf{30.3} & 40.2          & 11.5          & \textbf{24.8} & \textbf{63.2} & \textbf{57.9}
\end{tabular}}
\caption{IoU per class on the validation set of Cityscapes~\cite{cordts2016cityscapes} with U-Net based architecture~\cite{isola2017image} as segmentation network}
\label{iou_per_class}
\end{table*}

\begin{table*}[t!]
\small
\resizebox{\textwidth}{!}{
\begin{tabular}{llllllllllllllllllll}
\hline
Method                          & road          & swalk         & build         & wall          & fence         & pole          & tlight        & sign          & veg.          & ter.          & sky           & pers          & rider         & car           & truck         & bus  & train        & mbike         & bike          \\ \hline
\multicolumn{1}{l|}{CE}         & 84.8          & 69.0          & 77.3          & 15.6          & 13.7          & 66.4          & 31.3          & 53.7          & 82.3          & 28.7          & 82.0          & 47.5          & 29.2          & 76.0          & 8.3           & 12.2 & 2.6          & 8.9           & 44.1          \\
\multicolumn{1}{l|}{Focal loss ~\cite{lin2017focal}} & 87.2          & 72.4          & 80.7          & 19.7          & 16.0          & 71.0          & 40.1          & 62.3          & 86.1          & \textbf{35.8} & 84.8          & 51.6          & 32.0          & 79.4          & 9.2           & 18.0 & 4.3          & 12.1          & 50.5          \\
\multicolumn{1}{l|}{CE + adv ~\cite{luc2016semantic}}   & 82.6          & 72.3          & 79.8          & 16.2          & 16.2          & 72.1          & 43.6          & 65.7          & 86.2          & 34.5          & 83.3          & 54.8          & 34.4          & 78.8          & 8.7           & 17.5 & 4.4          & 14.0          & 52.0          \\
\multicolumn{1}{l|}{EL-GAN ~\cite{ghafoorian2018gan}}     & 86.9          & 72.3          & 79.9          & 19.3          & 16.4          & 70.7          & 38.2          & 63.4          & 85.5          & 32.7          & 84.0          & 51.2          & 32.7          & 78.1          & 9.5           & 16.8 & \textbf{4.8} & 8.8           & 47.0          \\ \hline
\multicolumn{1}{l|}{Gambling}   & \textbf{87.4} & \textbf{74.3} & \textbf{81.3} & \textbf{20.7} & \textbf{18.6} & \textbf{74.0} & \textbf{45.7} & \textbf{67.8} & \textbf{87.2} & 35.4          & \textbf{85.4} & \textbf{57.0} & \textbf{38.8} & \textbf{80.0} & \textbf{11.2} & \textbf{19.3} & 4.4          & \textbf{15.6} & \textbf{52.9}
\end{tabular}}
\caption{BF-score \cite{csurka2013good} per class on the validation set of Cityscapes~\cite{cordts2016cityscapes} with U-Net based architecture~\cite{isola2017image} as segmentation network}
\label{bf_per_class}
\end{table*}


\begin{table}[]
\begin{tabular}{lllll}
\hline
Method                                           & Mean IoU  & BF-score & Hausdorff  \\ \hline
\multicolumn{1}{l|}{CE}                          & 72.4         & 69.0          & 19.4        \\
\multicolumn{1}{l|}{Focal loss~\cite{lin2017focal}}                  & 71.5         & 67.4          & 21.2       \\
\multicolumn{1}{l|}{CE + adv~\cite{luc2016semantic}}                    & 68.0         & 67.0          & 20.9     \\
\multicolumn{1}{l|}{EL-GAN~\cite{ghafoorian2018gan}}                      & 71.3         & 67.0          & 21.2  \\ \hline
\multicolumn{1}{l|}{Gambling nets}               & \textbf{73.1}  & \textbf{70.1} & \textbf{18.7}  \\
     
\end{tabular}
\caption{Results on Cityscapes~\cite{cordts2016cityscapes} with PSPNet~\cite{zhao2017pyramid}  as segmentation network.}
\label{PSP_city}
\end{table}

\begin{table}[]
\begin{tabular}{lllll}
\hline
Method                                           & Mean IoU  & BF-score & Hausdorff  \\ \hline
\multicolumn{1}{l|}{CE}                          & 72.5         & 71.8          & 17.9        \\
\multicolumn{1}{l|}{Focal loss~\cite{lin2017focal}}                  & 70.8         & 71.4          & 17.7       \\
\multicolumn{1}{l|}{CE + adv~\cite{luc2016semantic}}                    & \textbf{72.7}         & 72.7          & 17.1    \\
\multicolumn{1}{l|}{EL-GAN~\cite{ghafoorian2018gan}}                      &  70.1        &  69.6         & 19.1  \\ \hline
\multicolumn{1}{l|}{Gambling nets}               & 72.1  & \textbf{73.8} & \textbf{16.0}  \\
     
\end{tabular}
\caption{Results on Camvid~\cite{brostow2009semantic} with PSPNet~\cite{zhao2017pyramid}  as segmentation network.}
\label{PSP_camvid}
\end{table}

\paragraph{PSPNet segmenter.}
We conduct experiments with the PSPNet~\cite{zhao2017pyramid} segmenter on the Camvid~\cite{brostow2009semantic} and Cityscapes~\cite{cordts2016cityscapes} datasets. In Table \ref{PSP_city}, the results are shown on the Cityscapes validation set. Again, the gambling adversarial networks perform better than the existing methods, on both of the structure-based metrics as well as the mean IoU. In Figure \ref{fig:psp_city}, a qualitative sample is shown, more can be found in the supplementary material. The gambling adversarial networks provides more details to the traffic lights. Also, the structure of the sidewalk shows significant improvements over the predictions from the model trained with standard segmentation loss. 
\begin{figure}[]
\captionsetup[subfigure]{labelformat=empty}
\centering
\begin{subfigure}[t]{0.49\linewidth}
    \includegraphics[width=\linewidth, trim={12cm 5cm 0cm 1cm},clip]{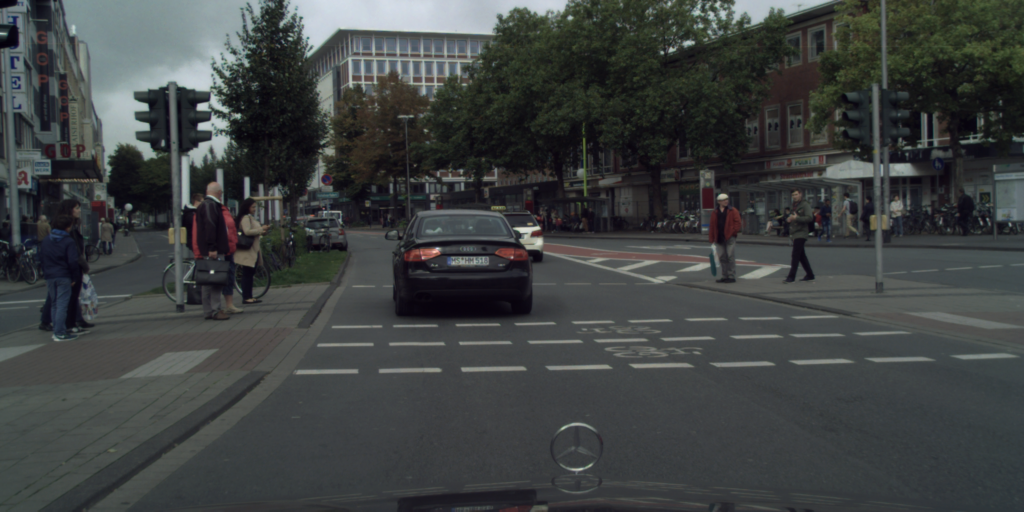}
    \caption{RGB-image}

\end{subfigure}
\hfill
\begin{subfigure}[t]{0.49\linewidth}
    \includegraphics[width=\linewidth, trim={12cm 5cm 0cm 1cm},clip]{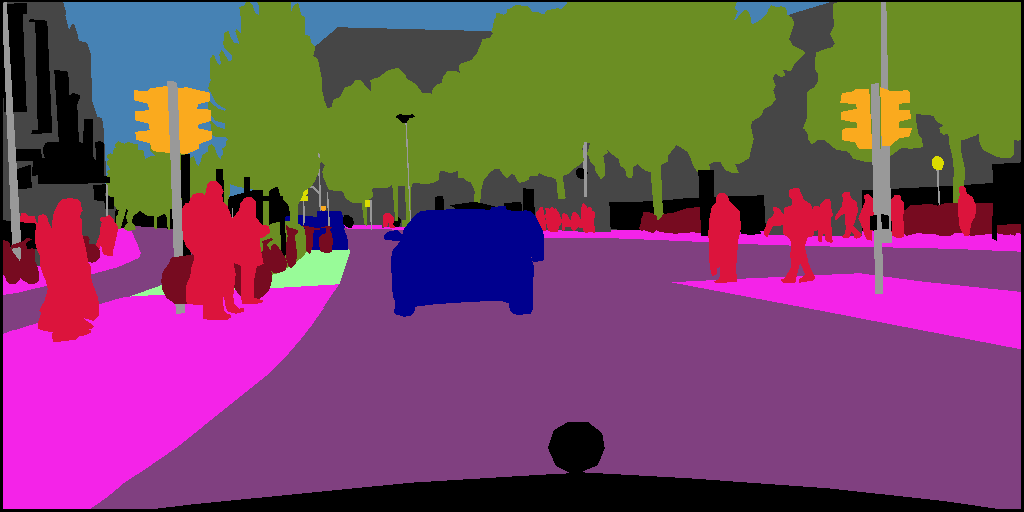}
    \caption{Ground-truth}
\end{subfigure}
\hfill
\begin{subfigure}[t]{0.49\linewidth}
    \includegraphics[width=\linewidth, trim={12cm 5cm 0cm 1cm},clip]{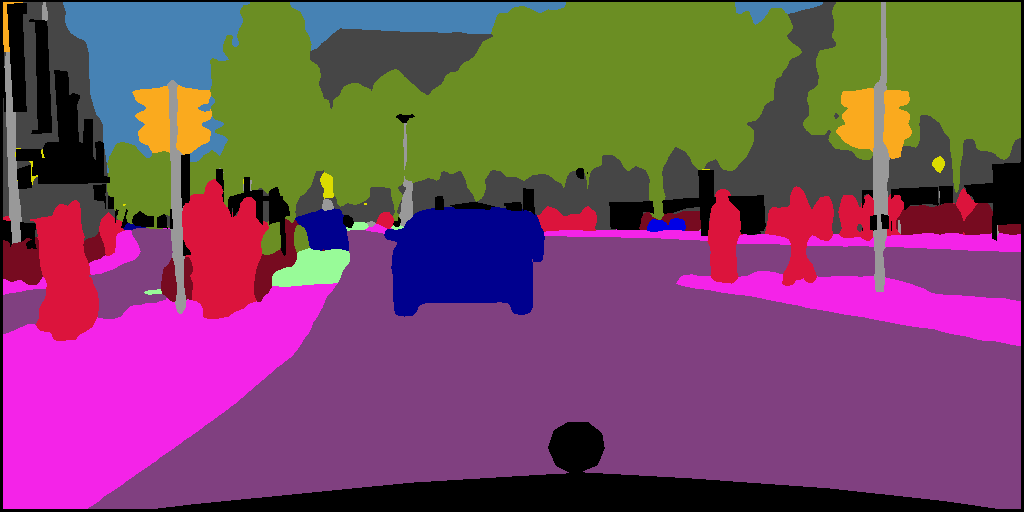}
    \caption{Cross-entropy}
\end{subfigure}
\hfill
\begin{subfigure}[t]{0.49\linewidth}
    \includegraphics[width=\linewidth, trim={12cm 5cm 0cm 1cm},clip]{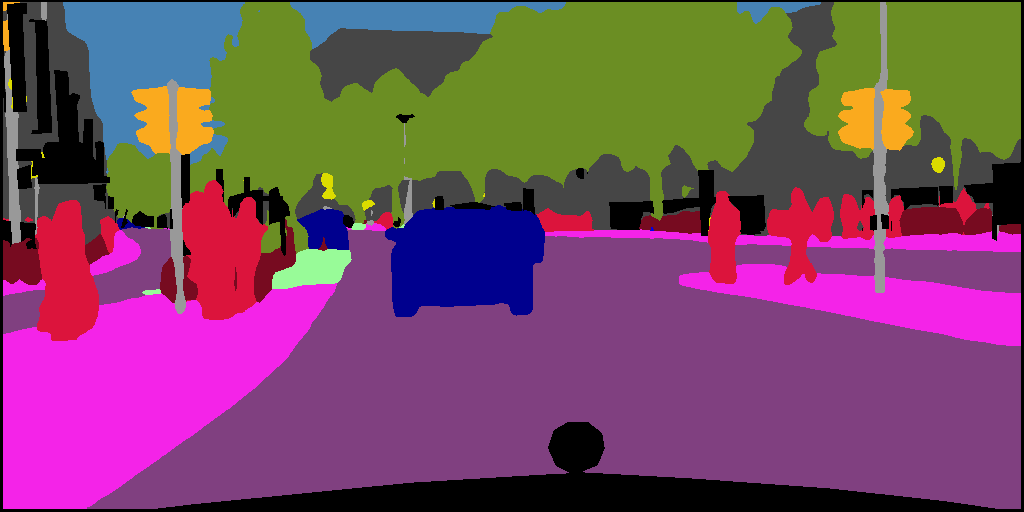}
    \caption{Gambling nets}
\end{subfigure}
\caption{Qualitative results on Cityscapes~\cite{cordts2016cityscapes} with  PSPNet~\cite{zhao2017pyramid} as segmentation network}
\label{fig:psp_city}
\end{figure}

The quantitative results on the Camvid~\cite{brostow2009semantic} test set are shown in Table \ref{PSP_camvid}. The gambling adversarial networks achieve the highest score on the structure-based metrics, but the standard adversarial training~\cite{luc2016semantic} performs best on the IoU. In the supplementary material, we provide qualitative results for the Camvid ~\cite{brostow2009semantic} test set and extra images for the aforementioned experiments.

\section{Discussion}
\paragraph{Correct/incorrect versus real/fake discrimination.}
We reformulated the adversarial real/fake discrimination task into training a critic network to learn to spot the likely incorrect predictions. As shown in Figure \ref{fig:mean_max_scratch}, the discrimination of real and fake causes undesired gradient feedback, since all the softmax vectors converge to a one-hot vector. We empirically showed that this behavior is caused by a value-based discrimination of the adversarial network. Moreover, modifying the adversarial task to correct/incorrect discrimination solves several problems. First of all, the reason to apply adversarial training to semantic segmentation is to improve on the high-level structures. However, the value-based discriminator is not only providing feedback based on the visual difference between the predictions and the labels, but also an undesirable value-based feedback. Moreover, updating the weights in a network with the constraint that the output must be a one-hot vector complicates training unnecessarily. Finally, the value-based discriminator hinders the network from properly disclosing uncertainty. Both the structured prediction and expressing uncertainty can be of great value for semantic segmentation, e.g. in autonomous driving and medical imaging applications. However, changing the adversarial task to discriminating the correct from the incorrect predictions resolves the aforementioned issues. The segmentation network is not forced to imitate the one-hot vectors, which preserves the uncertainty in the predictions and simplifies the training. Although we still notice that the gambler sometimes utilizes the prediction values by betting on pixels where the segmenter is uncertain, we also obtain improvements on the structure-based metrics compared to the existing adversarial methods. 

\paragraph{Gambling adversarial networks vs. focal loss}
The adversarial loss in gambling adversarial networks resembles the focal loss~\cite{lin2017focal}, since both methods up-weight the harder samples that contain more useful information for the update. The focal loss is defined as: $\mathcal{L}_{foc}(y, \hat{y}, p_t) = -(1-p_t)^{\gamma} y \log \hat{y}$, where $p_t$ is the probability of the correct class and $\gamma$ is a focusing factor, which indicates how much the easier samples are down-weighted. The advantage of the focal loss is that the ground-truth is exploited to choose the weights, however, the downside is that the focal loss might be over-pronouncing the ambiguous or incorrectly labeled pixels. The adversarial loss in gambling adversarial networks learns the weighting map, which can mitigate the noise effect. Moreover, the adversarial loss generates an extra flow of gradients (flow B), as observable in Figure \ref{fig:adv_loss_expl}. Gradient stream A provides information to the segmentation network independent of other pixel predictions similar to the focal loss, whereas gradient stream B provides gradients reflecting structural qualities, which is lacking in case of the focal loss.

\paragraph{Insights into betting maps}
Inspecting the betting maps (see for instance Figure~\ref{fig:Unet_city}), we observe that some of the bets correspond to the class borders, especially the ones that seemingly do not match the visual evidence in the underlying image or the expected shape of the object. We should note that even though there are chances that the ground-truth labels on the borders are different from the predictions, blindly betting on all the borders is not even close to the optimal policy. The clear bad structures in the predictions, e.g. the weird prediction of rider inside the car or the badly formed wall on the left side, are still more rewarding investments that are also being spotted by the gambler.

\section{Conclusion}
In this paper, we studied a novel reformulation of adversarial training for semantic segmentation, in which we replace the discriminator with a gambler network that learns to use the inter-pixel consistency clues to spot the wrong predictions. We showed that involving the segmenter in a minimax game with such a gambler results in notable improvements in structural and pixel-wise metrics, as measured on two road-scene semantic segmentation datasets.




\newpage
\onecolumn
\begin{center}
\huge
Supplementary Material 
\end{center}




\noindent In this section, we provide the hyperparameters for gambling adversarial networks and extra qualitative results for the different experiments.

\subsection*{Hyperparameters}
\noindent In the following paragraphs, the hyperparameters for the experiments in the results section are described.

\paragraph{U-Net based architecture on Cityscapes.}
Training details for the experiment on Cityscapes \cite{cordts2016cityscapes} with U-Net based architecture \cite{isola2017image}. We trained the segmenter and gambler in alternating fashion of 200 and 400 iterations respectively over 300 epochs with a batch size of 4. The betting maps are calculated with a smoothing factor $\beta$ of 0.02. Details for the segmenter and gambler are as following:
\begin{itemize}
    \item \textbf{Segmenter}: optimizer: Adam \cite{kingma2014adam}, learning rate: 1e-4, beta1: 0.5, beta2: 0.99, adversarial coefficient $\lambda$: 1.0, weight decay: 5e-4.
    \item \textbf{Gambler}: optimizer: Adam \cite{kingma2014adam}, learning rate 1e-4, beta1: 0.5, beta2: 0.99, weight decay: 5e-4.
\end{itemize}
\paragraph{PSPnet on Cityscapes.}
Training details for the experiment on Cityscapes \cite{cordts2016cityscapes} with PSPNet \cite{zhao2017pyramid}. We trained the segmenter and gambler in alternating fashion of 800 and 800 iterations respectively over 200 epochs with a batch size of 3. The betting maps are calculated with a smoothing factor $\beta$ of 0.02. Details for the segmenter and gambler are as following:
\begin{itemize}
    \item \textbf{Segmenter}: optimizer: Adam \cite{kingma2014adam}, learning rate: 2.5e-5, beta1: 0.5, beta2: 0.99, adversarial coefficient $\lambda$: 1.0, weight decay: 5e-4.
    \item \textbf{Gambler}: optimizer: Adam \cite{kingma2014adam}, learning rate 2.5e-5, beta1: 0.5, beta2: 0.99, weight decay: 5e-4.
\end{itemize}

\paragraph{PSPNet  on Camvid.}
Training details for the experiment on Camvid \cite{brostow2009semantic} with PSPNet \cite{zhao2017pyramid}. We trained the segmenter and gambler in alternating fashion of 100 and 200 iterations respectively over 100 epochs with a batch size of 2. The betting maps are calculated with a smoothing factor $\beta$ of 0.02. Details for the segmenter and gambler are as following:
\begin{itemize}
    \item \textbf{Segmenter}: optimizer: Adam \cite{kingma2014adam}, learning rate: 5e-5, beta1: 0.5, beta2: 0.99, adversarial coefficient $\lambda$: 0.5, weight decay: 5e-4.
    \item \textbf{Gambler}: optimizer: Adam \cite{kingma2014adam}, learning rate 5e-5, beta1: 0.5, beta2: 0.99, weight decay: 5e-4.
\end{itemize}

\subsection*{Qualitative results}
\noindent In Figures ~\ref{extra_qual_city_unet} and ~\ref{extra_qual_city_psp}, extra qualitative results are depicted for the experiments on the Cityscapes \cite{cordts2016cityscapes} validation set for the U-Net based architecture \cite{isola2017image} and PSPNet \cite{zhao2017pyramid} as segmentation network. In Figure ~\ref{extra_qual_cam_psp}, some samples are shown on the test set of Camvid \cite{brostow2009semantic} with PSPNet as segmentation network.

\clearpage
\begin{landscape}
\begin{figure}[!t]
\centering
\begin{subfigure}[t]{0.12\linewidth}
    \includegraphics[width=\linewidth]{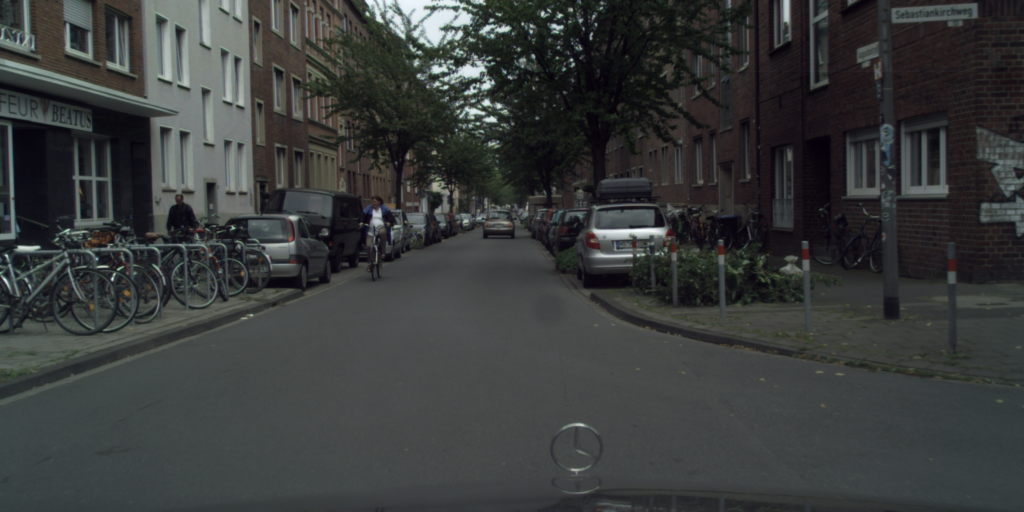}
\end{subfigure}
\begin{subfigure}[t]{0.12\linewidth}
    \includegraphics[width=\linewidth]{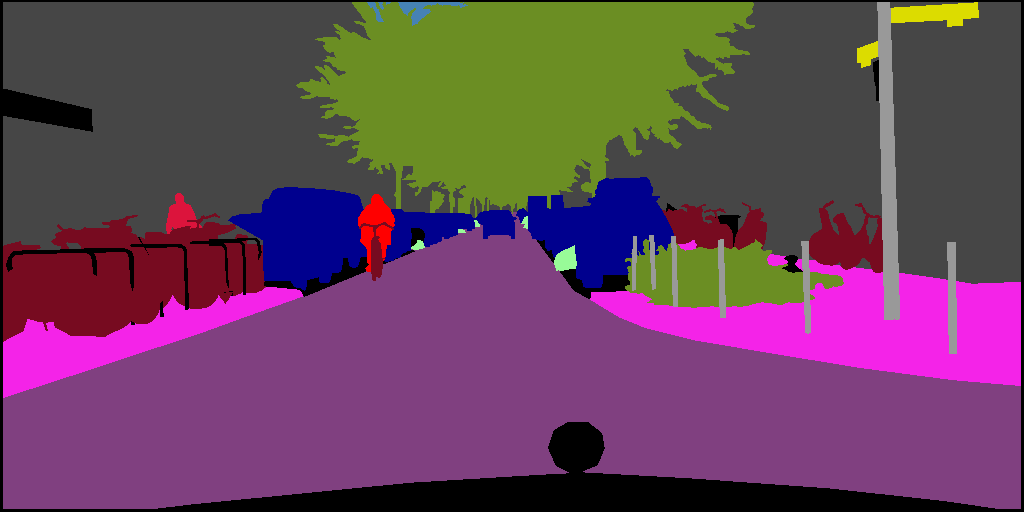}
\end{subfigure}
\begin{subfigure}[t]{0.12\linewidth}
    \includegraphics[width=\linewidth]{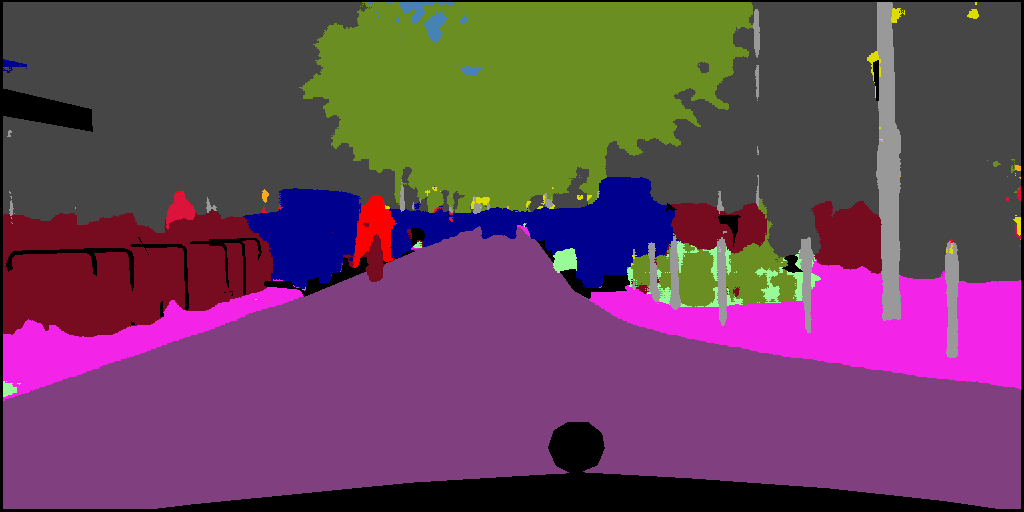}
\end{subfigure}
\begin{subfigure}[t]{0.12\linewidth}
    \includegraphics[width=\linewidth]{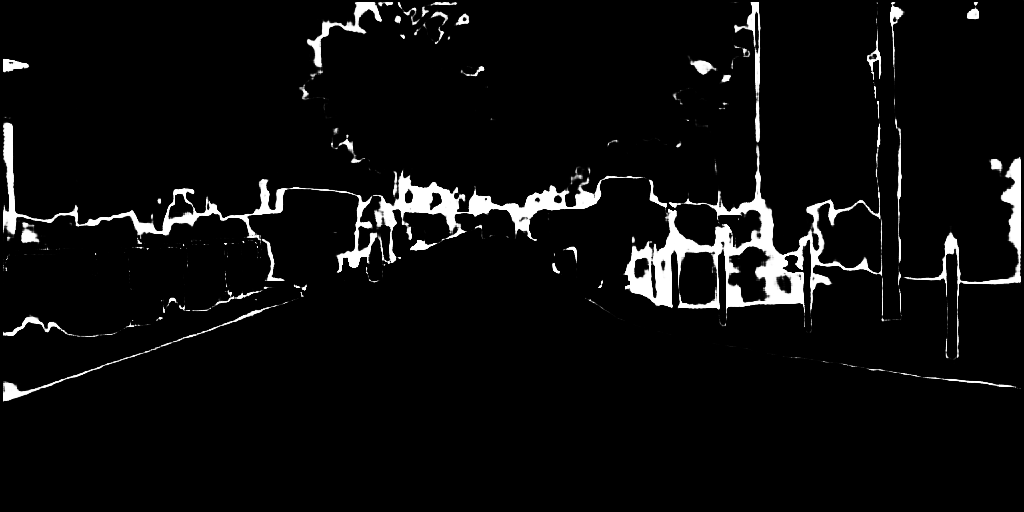}
\end{subfigure}
\begin{subfigure}[t]{0.12\linewidth}
    \includegraphics[width=\linewidth]{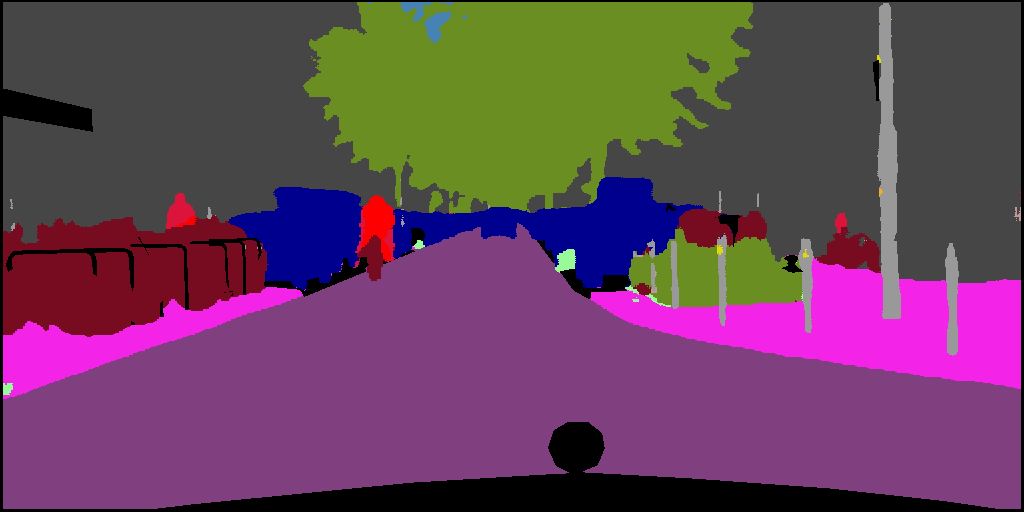}
\end{subfigure}
\begin{subfigure}[t]{0.12\linewidth}
    \includegraphics[width=\linewidth]{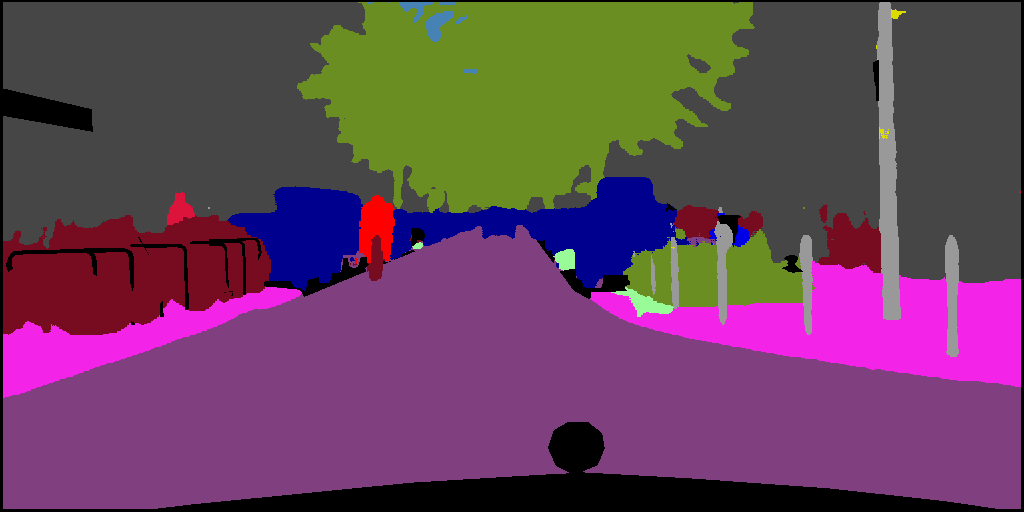}
\end{subfigure}
\begin{subfigure}[t]{0.12\linewidth}
    \includegraphics[width=\linewidth]{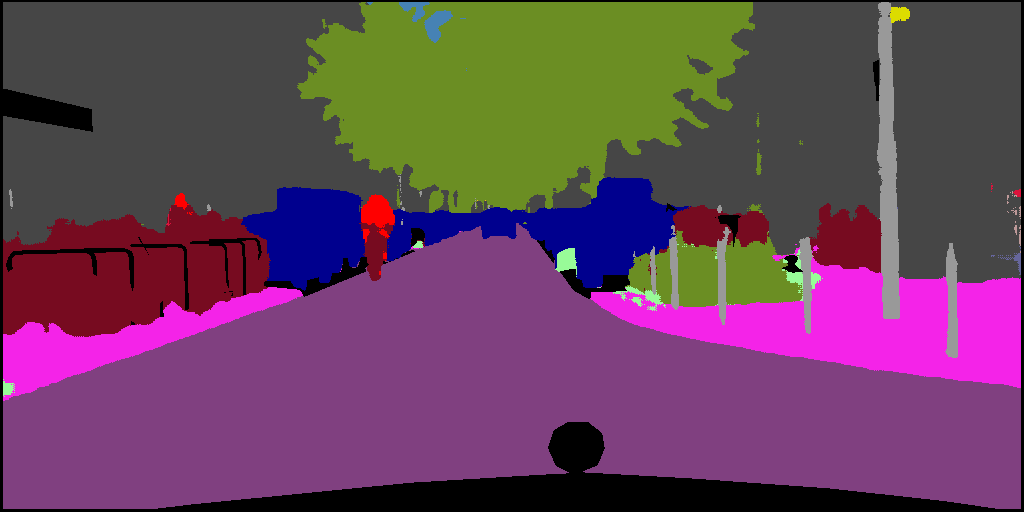}
\end{subfigure}
\begin{subfigure}[t]{0.12\linewidth}
    \includegraphics[width=\linewidth]{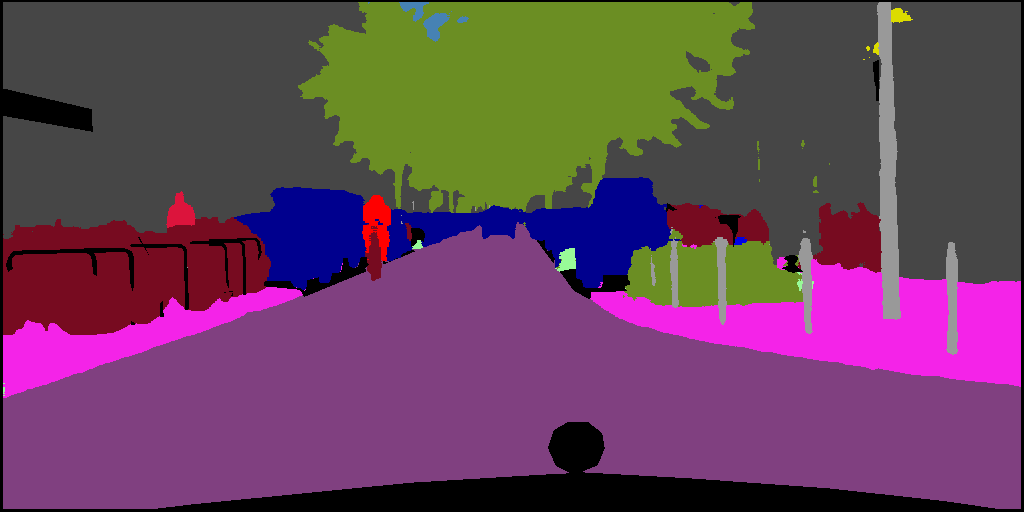}
\end{subfigure}
\begin{subfigure}[t]{0.12\linewidth}
    \includegraphics[width=\linewidth]{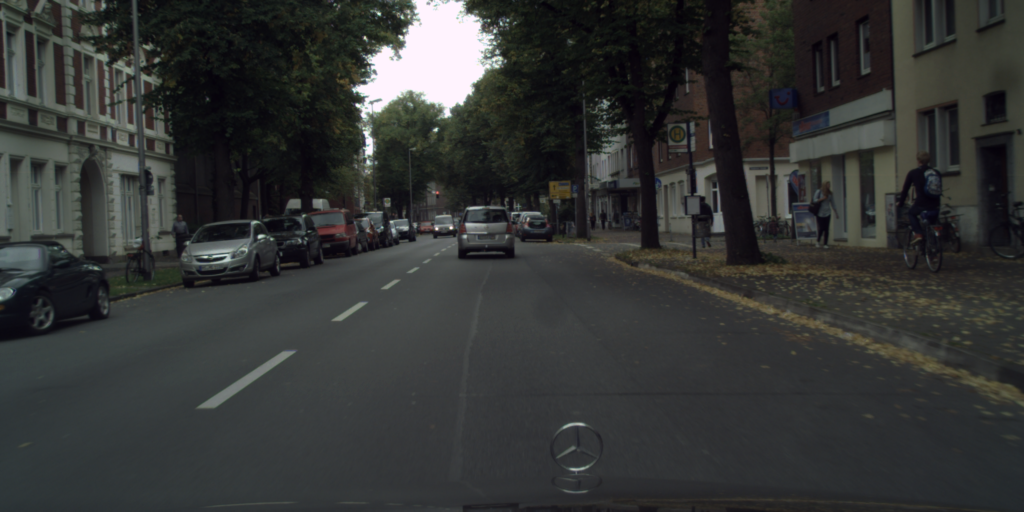}
\end{subfigure}
\begin{subfigure}[t]{0.12\linewidth}
    \includegraphics[width=\linewidth]{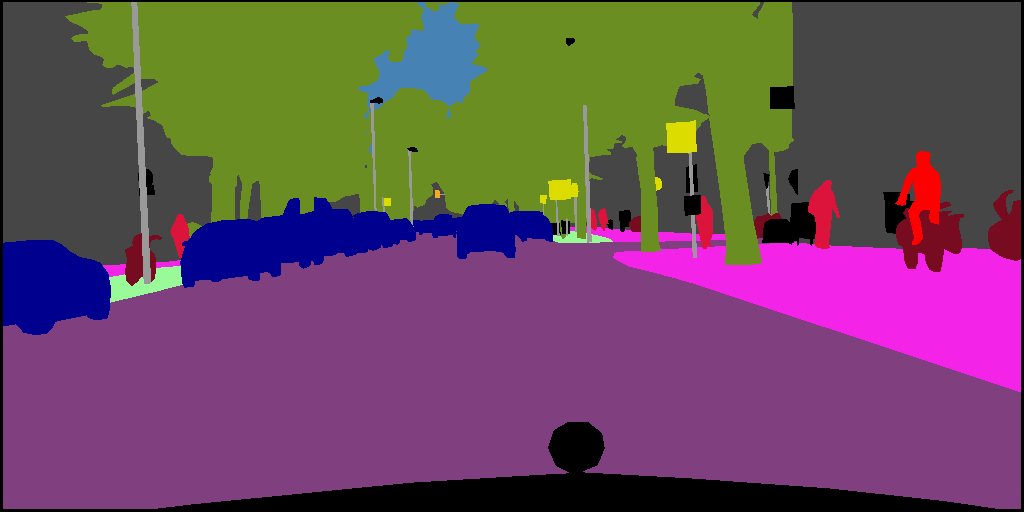}
\end{subfigure}
\begin{subfigure}[t]{0.12\linewidth}
    \includegraphics[width=\linewidth]{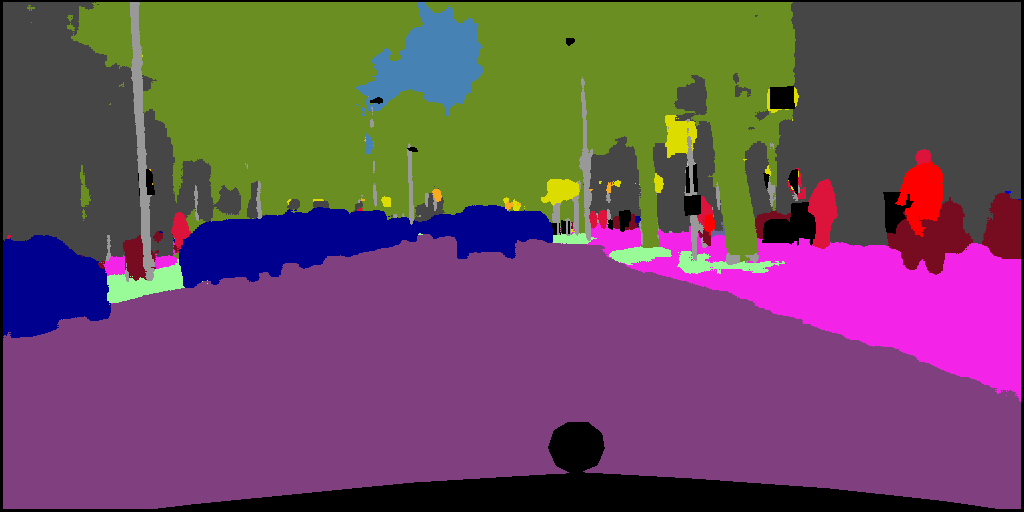}
\end{subfigure}
\begin{subfigure}[t]{0.12\linewidth}
    \includegraphics[width=\linewidth]{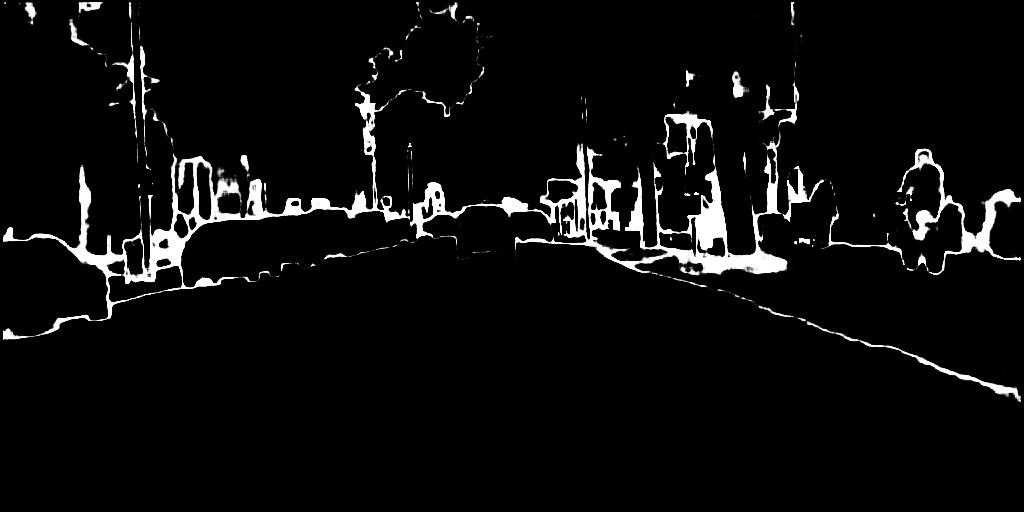}
\end{subfigure}
\begin{subfigure}[t]{0.12\linewidth}
    \includegraphics[width=\linewidth]{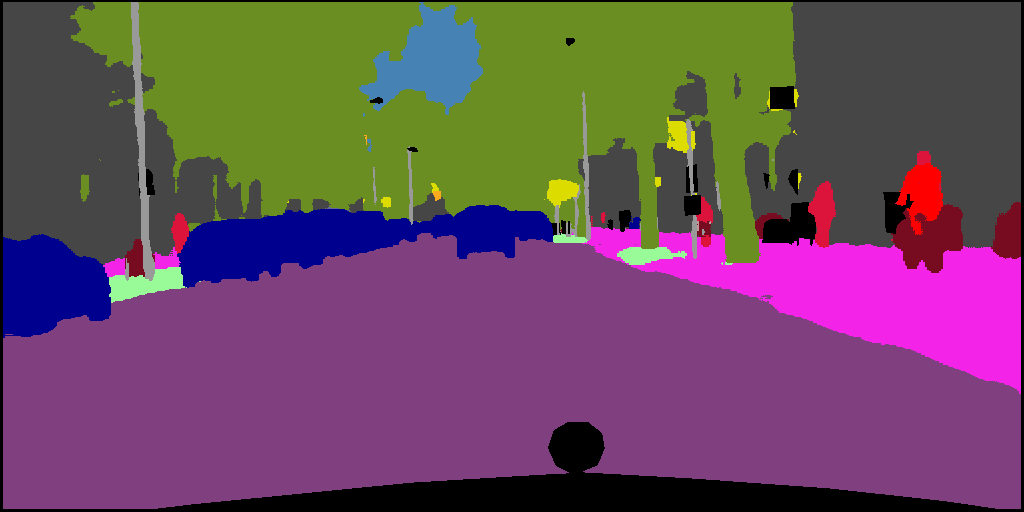}
\end{subfigure}
\begin{subfigure}[t]{0.12\linewidth}
    \includegraphics[width=\linewidth]{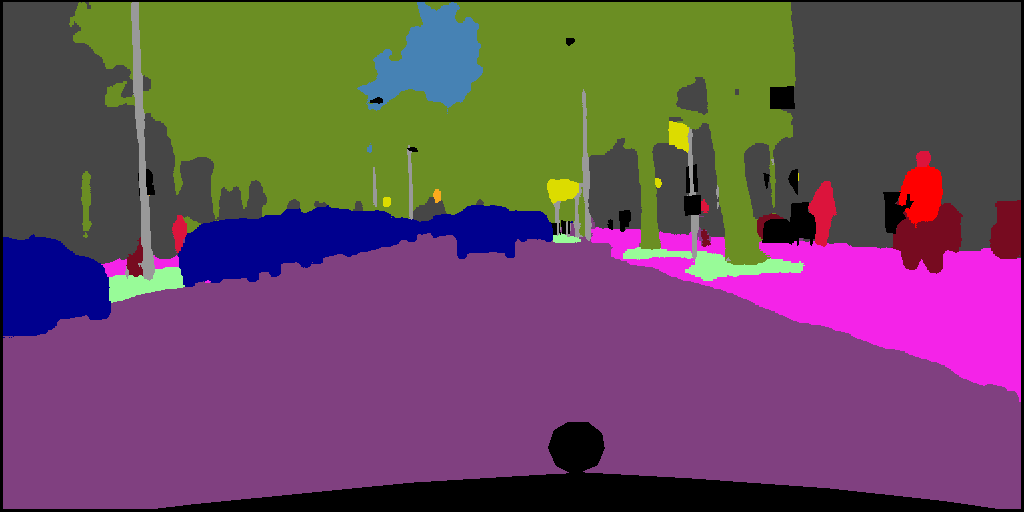}
\end{subfigure}
\begin{subfigure}[t]{0.12\linewidth}
    \includegraphics[width=\linewidth]{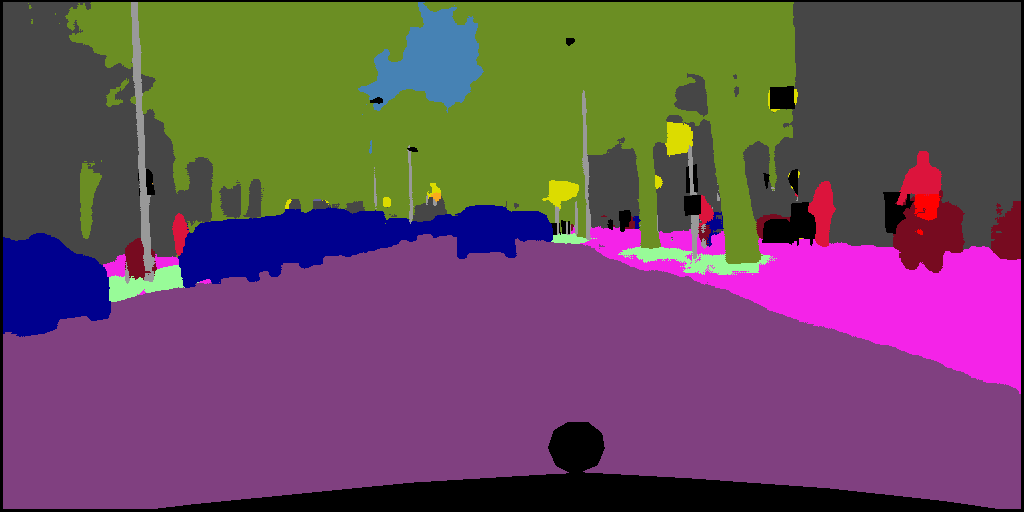}
\end{subfigure}
\begin{subfigure}[t]{0.12\linewidth}
    \includegraphics[width=\linewidth]{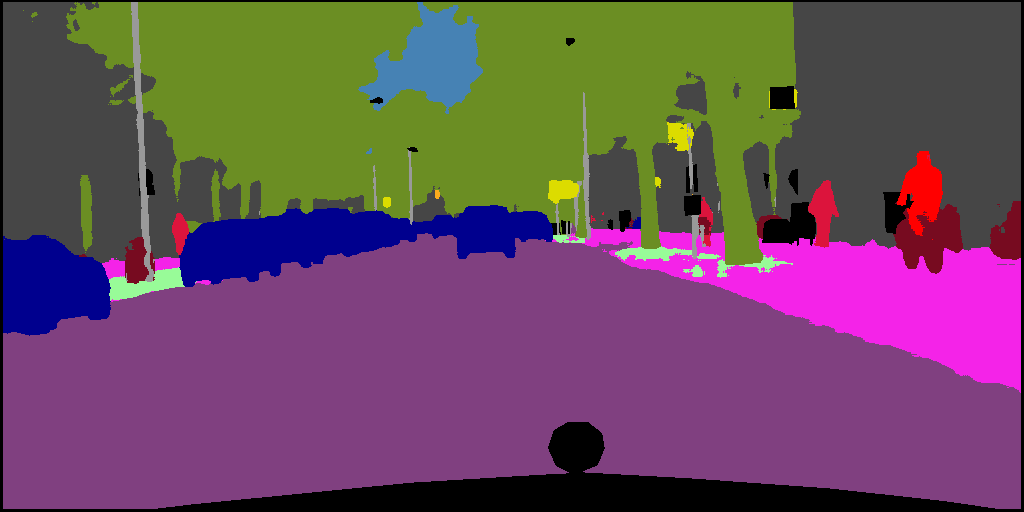}
\end{subfigure}
\begin{subfigure}[t]{0.12\linewidth}
    \includegraphics[width=\linewidth]{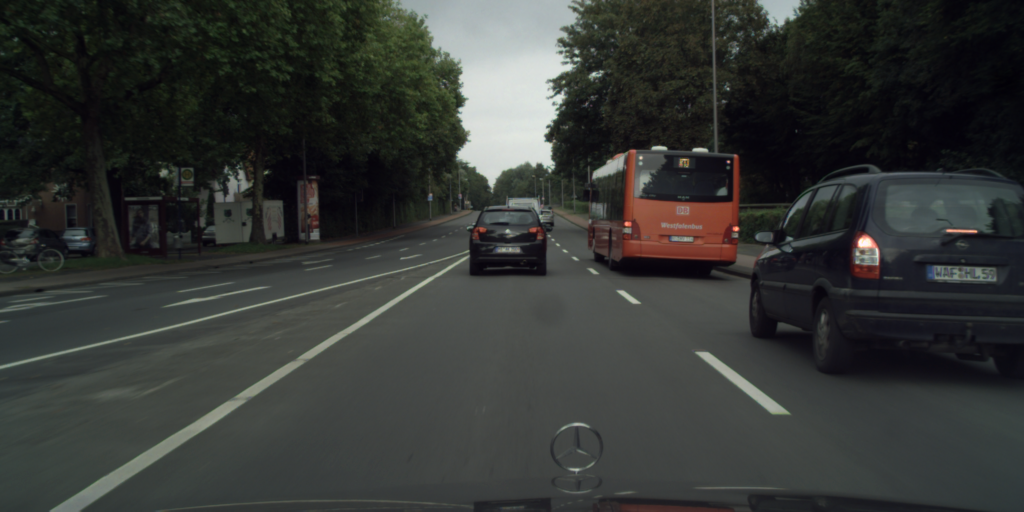}
\end{subfigure}
\begin{subfigure}[t]{0.12\linewidth}
    \includegraphics[width=\linewidth]{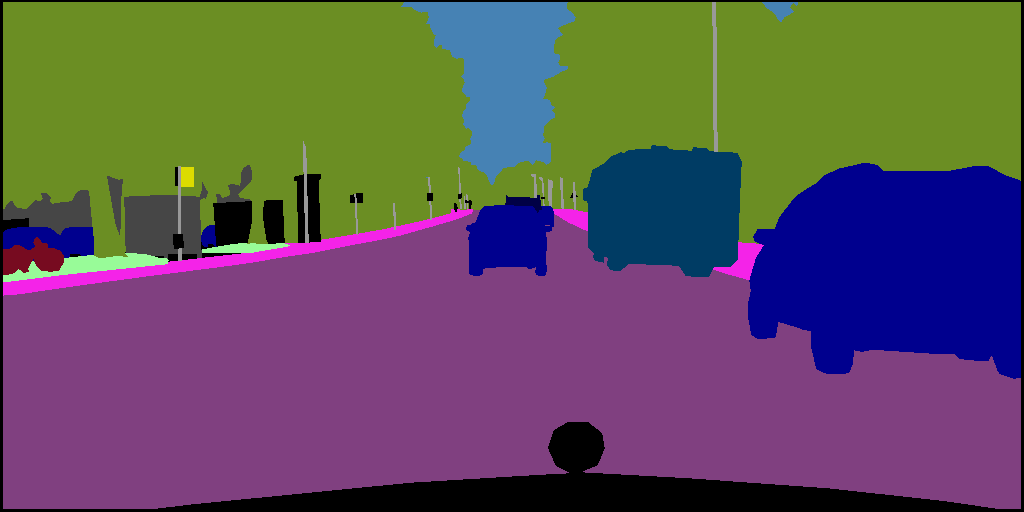}
\end{subfigure}
\begin{subfigure}[t]{0.12\linewidth}
    \includegraphics[width=\linewidth]{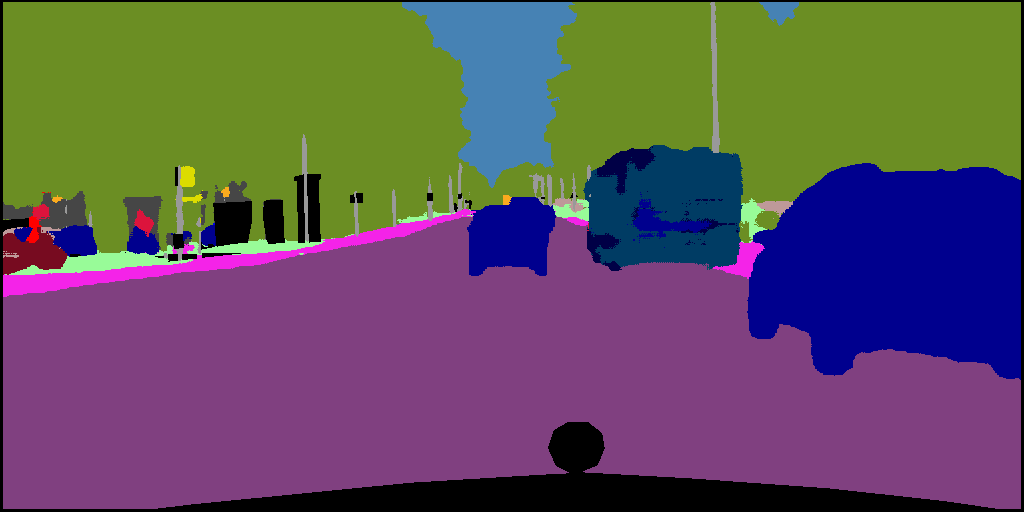}
\end{subfigure}
\begin{subfigure}[t]{0.12\linewidth}
    \includegraphics[width=\linewidth]{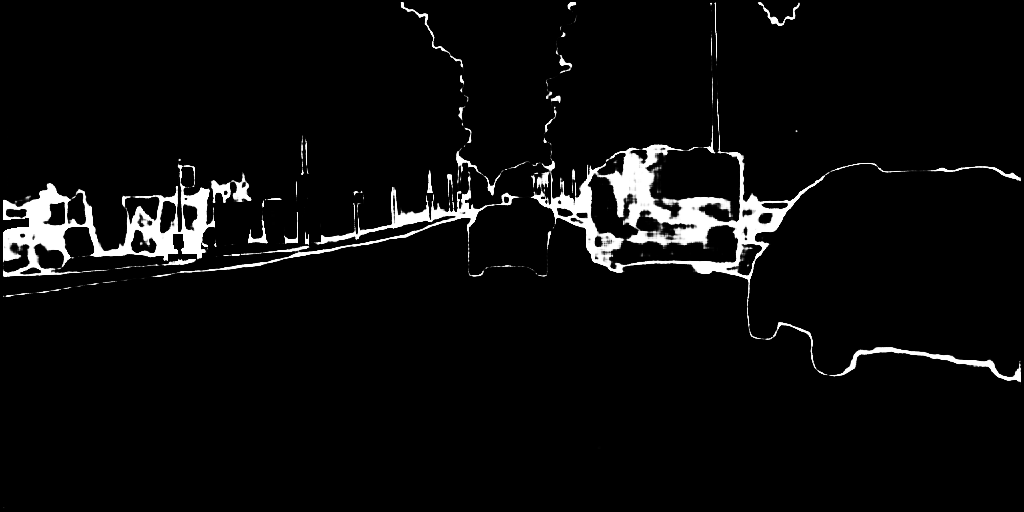}
\end{subfigure}
\begin{subfigure}[t]{0.12\linewidth}
    \includegraphics[width=\linewidth]{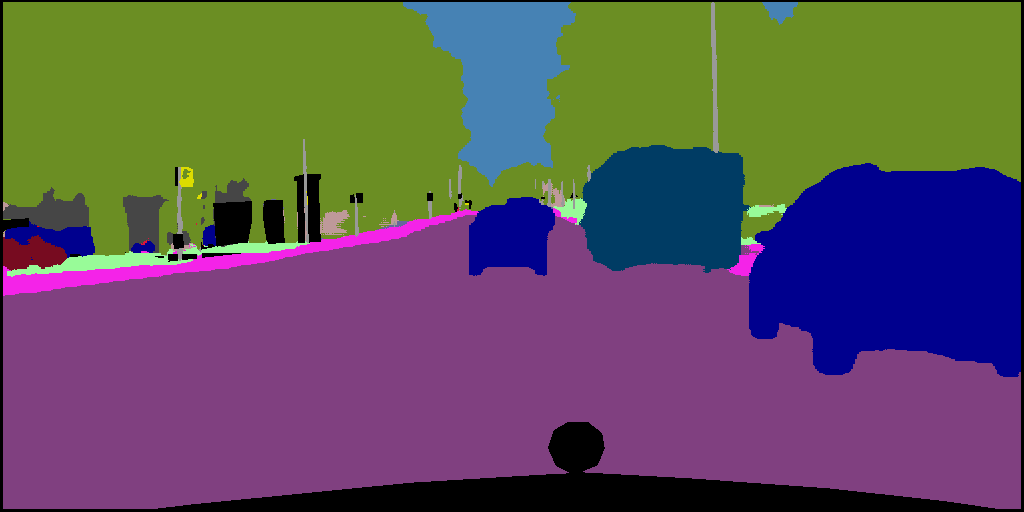}
\end{subfigure}
\begin{subfigure}[t]{0.12\linewidth}
    \includegraphics[width=\linewidth]{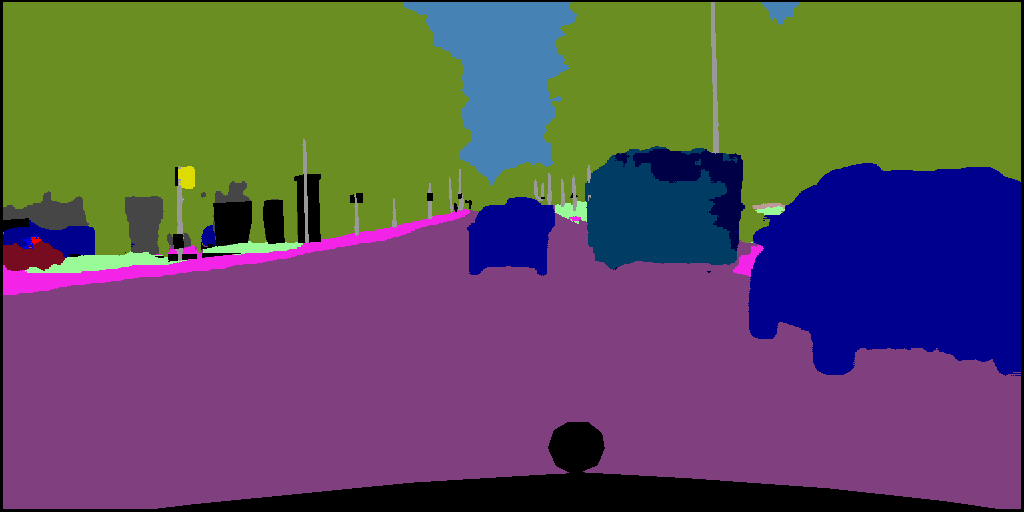}
\end{subfigure}
\begin{subfigure}[t]{0.12\linewidth}
    \includegraphics[width=\linewidth]{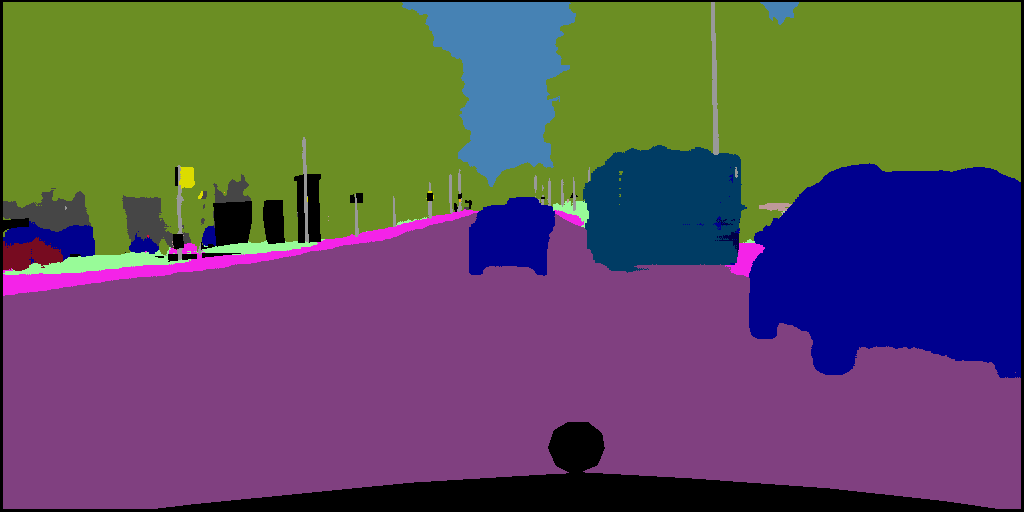}
\end{subfigure}
\begin{subfigure}[t]{0.12\linewidth}
    \includegraphics[width=\linewidth]{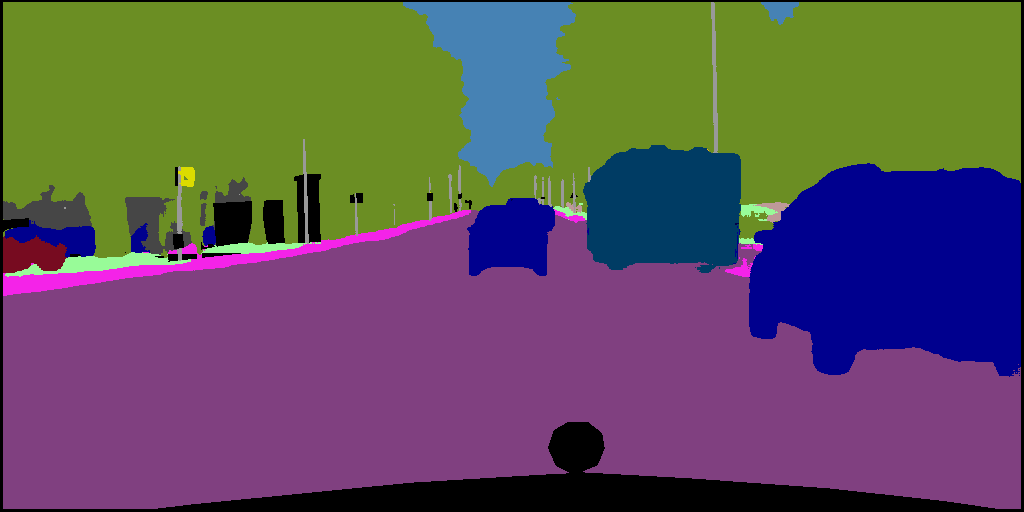}
\end{subfigure}
\caption{From left to right: RGB-image, ground-truth, CE, betting-map, focal loss, CE + adv, EL-GAN, gambling nets. The betting map is a prediction with as input the RGB image and the CE prediction. Results are for the Cityscapes \cite{cordts2016cityscapes} validation set with the U-Net based architecture \cite{isola2017image}. Best visible zoomed-in on a screen.}
\label{extra_qual_city_unet}
\end{figure}

\begin{figure}[!t]
\centering
\begin{subfigure}[t]{0.12\linewidth}
    \includegraphics[width=\linewidth]{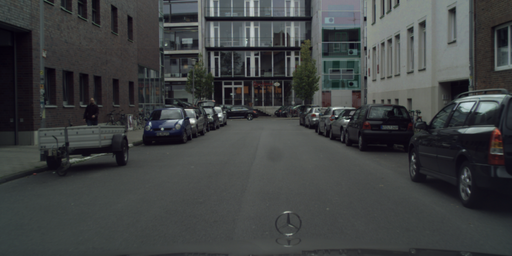}
\end{subfigure}
\begin{subfigure}[t]{0.12\linewidth}
    \includegraphics[width=\linewidth]{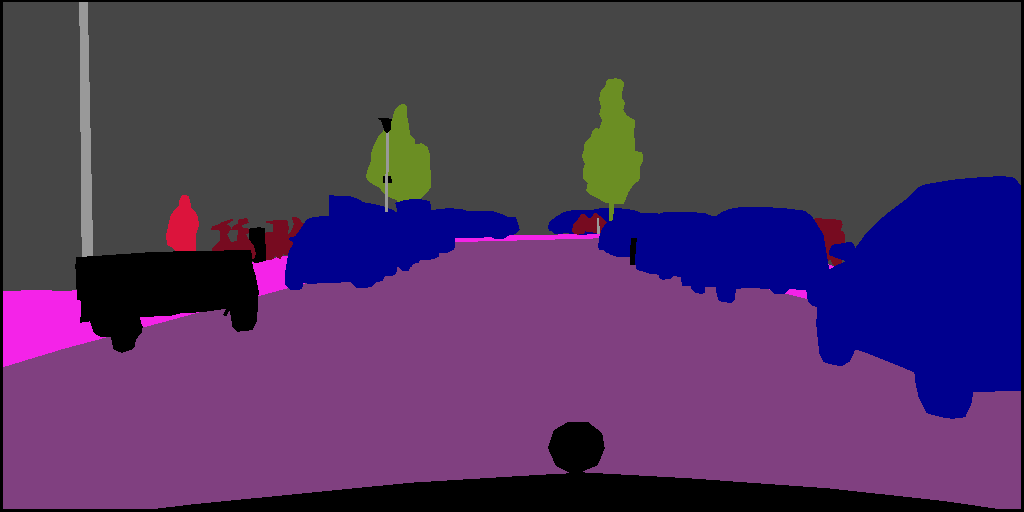}
\end{subfigure}
\begin{subfigure}[t]{0.12\linewidth}
    \includegraphics[width=\linewidth]{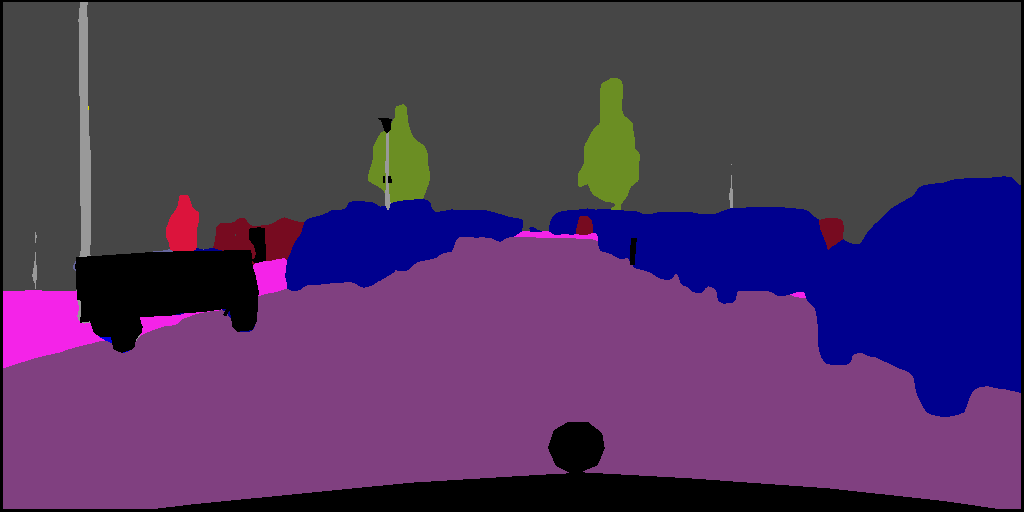}
\end{subfigure}
\begin{subfigure}[t]{0.12\linewidth}
    \includegraphics[width=\linewidth]{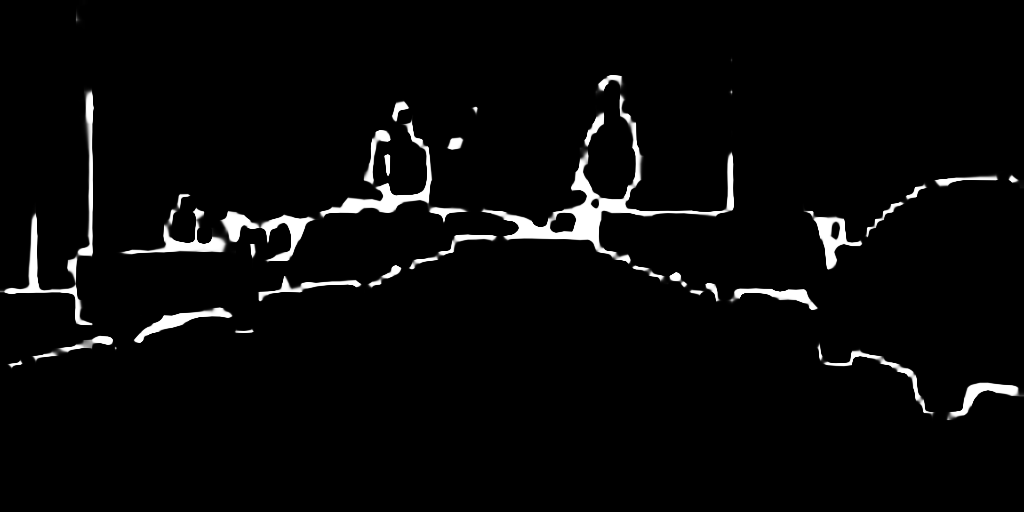}
\end{subfigure}
\begin{subfigure}[t]{0.12\linewidth}
    \includegraphics[width=\linewidth]{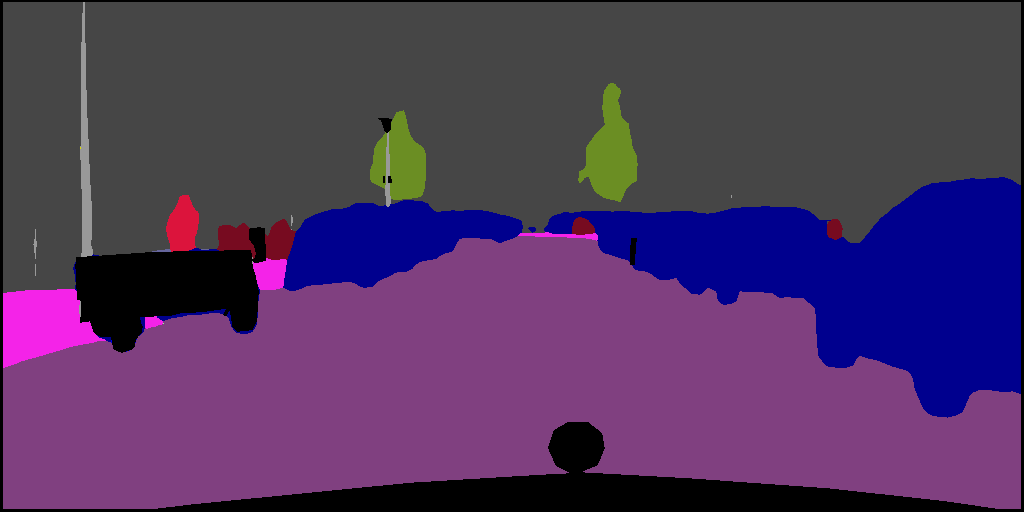}
\end{subfigure}
\begin{subfigure}[t]{0.12\linewidth}
    \includegraphics[width=\linewidth]{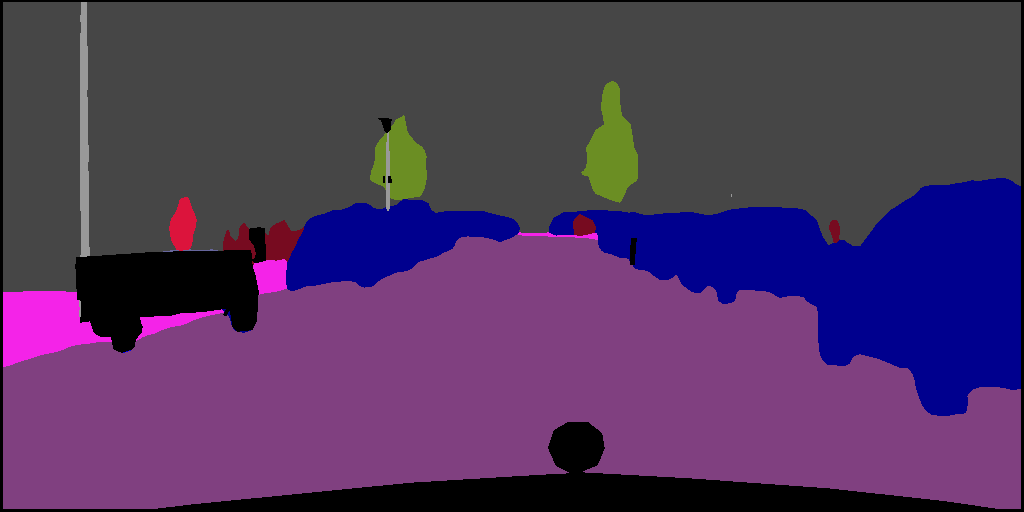}
\end{subfigure}
\begin{subfigure}[t]{0.12\linewidth}
    \includegraphics[width=\linewidth]{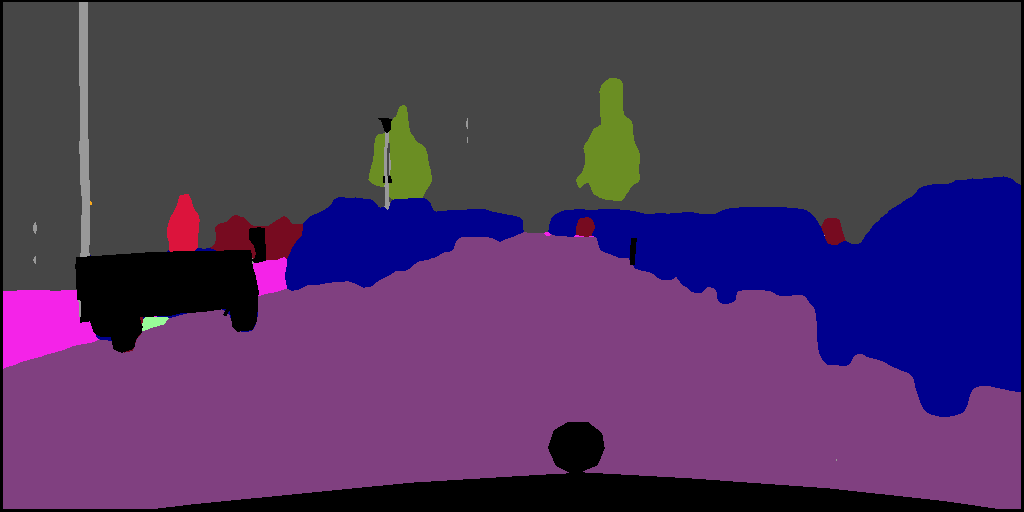}
\end{subfigure}
\begin{subfigure}[t]{0.12\linewidth}
    \includegraphics[width=\linewidth]{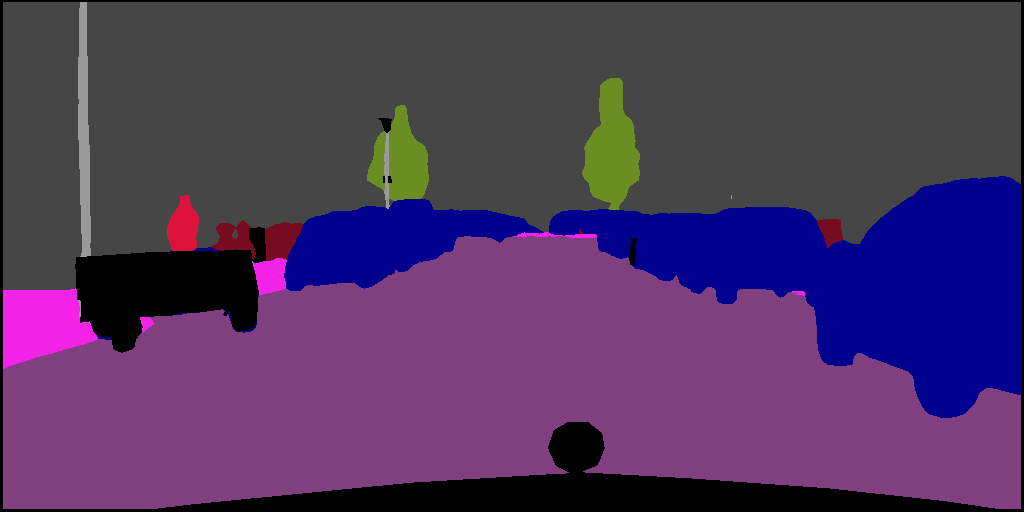}
\end{subfigure}
\begin{subfigure}[t]{0.12\linewidth}
    \includegraphics[width=\linewidth]{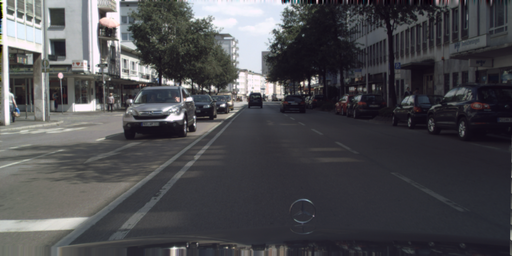}
\end{subfigure}
\begin{subfigure}[t]{0.12\linewidth}
    \includegraphics[width=\linewidth]{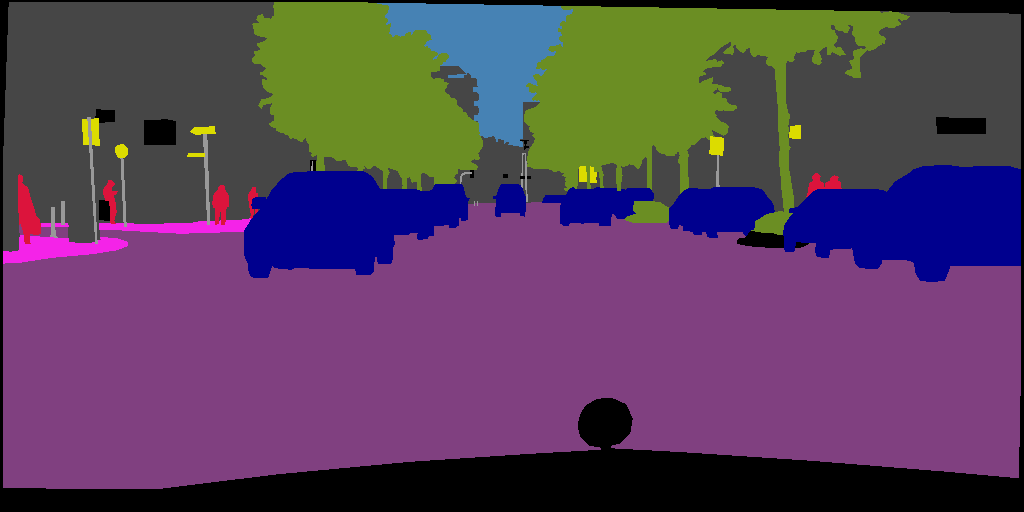}
\end{subfigure}
\begin{subfigure}[t]{0.12\linewidth}
    \includegraphics[width=\linewidth]{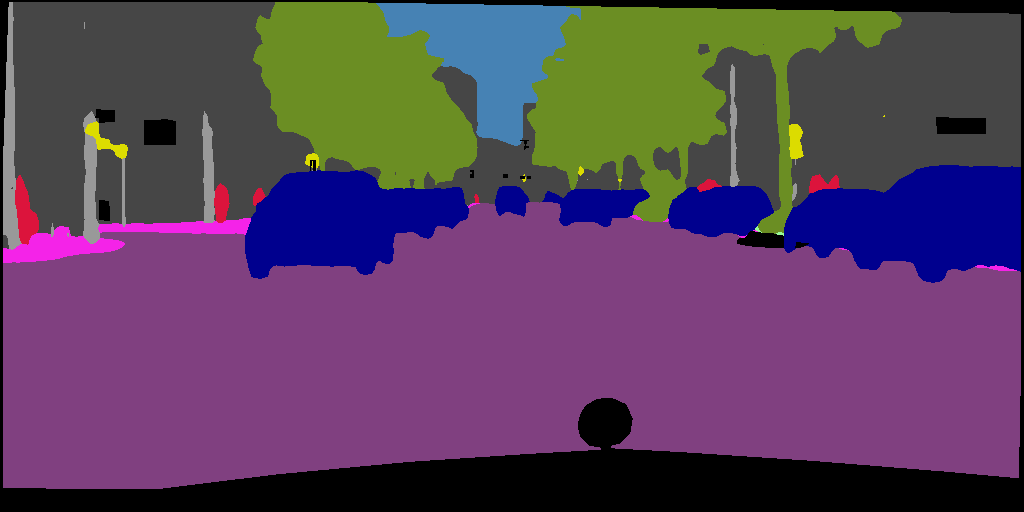}
\end{subfigure}
\begin{subfigure}[t]{0.12\linewidth}
    \includegraphics[width=\linewidth]{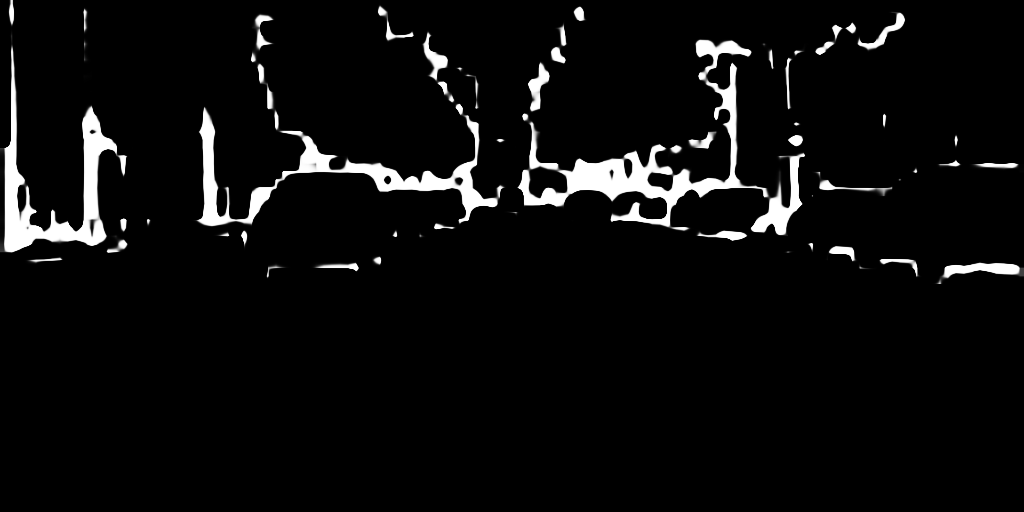}
\end{subfigure}
\begin{subfigure}[t]{0.12\linewidth}
    \includegraphics[width=\linewidth]{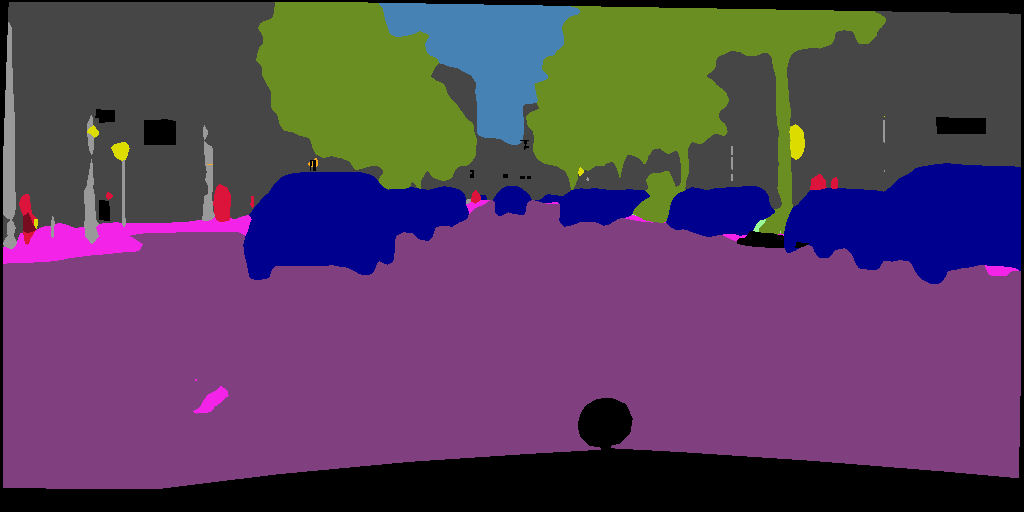}
\end{subfigure}
\begin{subfigure}[t]{0.12\linewidth}
    \includegraphics[width=\linewidth]{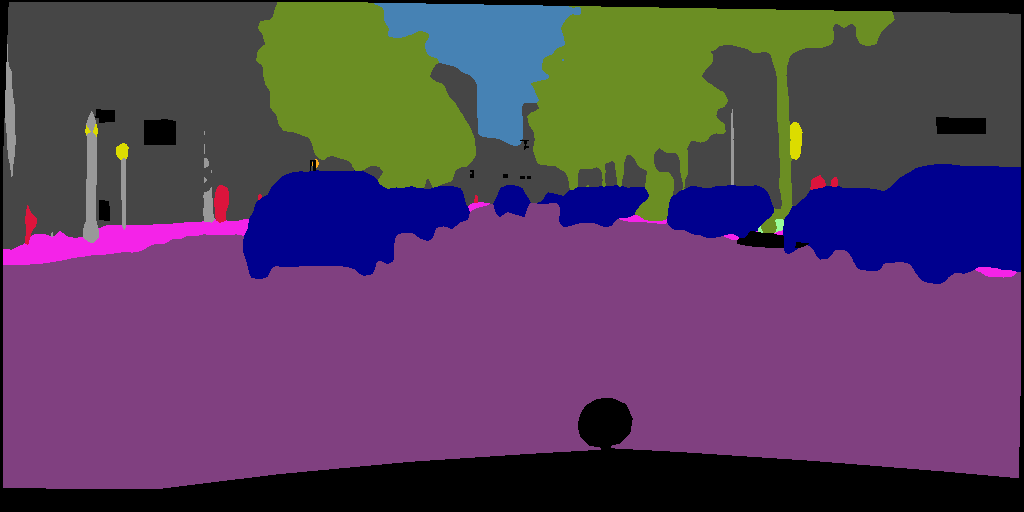}
\end{subfigure}
\begin{subfigure}[t]{0.12\linewidth}
    \includegraphics[width=\linewidth]{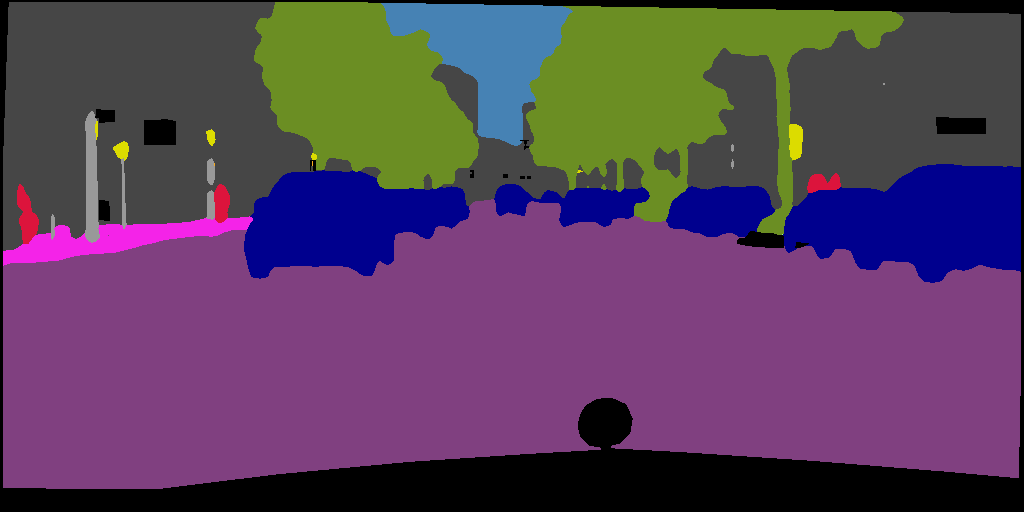}
\end{subfigure}
\begin{subfigure}[t]{0.12\linewidth}
    \includegraphics[width=\linewidth]{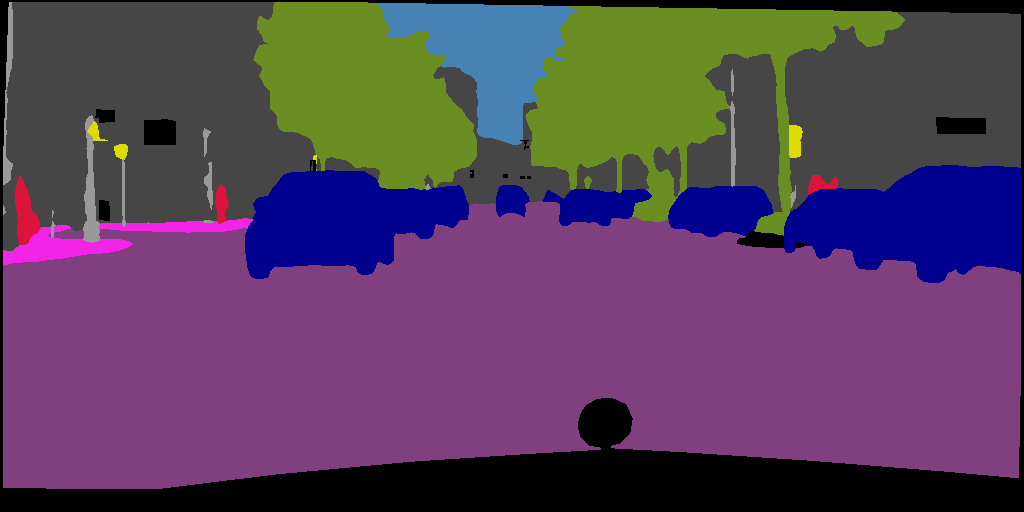}
\end{subfigure}
\begin{subfigure}[t]{0.12\linewidth}
    \includegraphics[width=\linewidth]{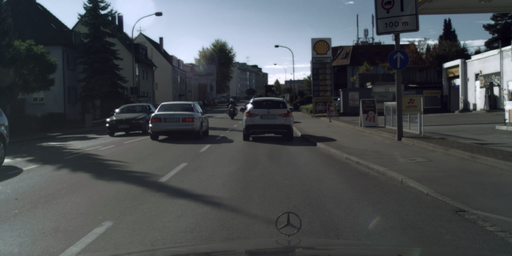}
\end{subfigure}
\begin{subfigure}[t]{0.12\linewidth}
    \includegraphics[width=\linewidth]{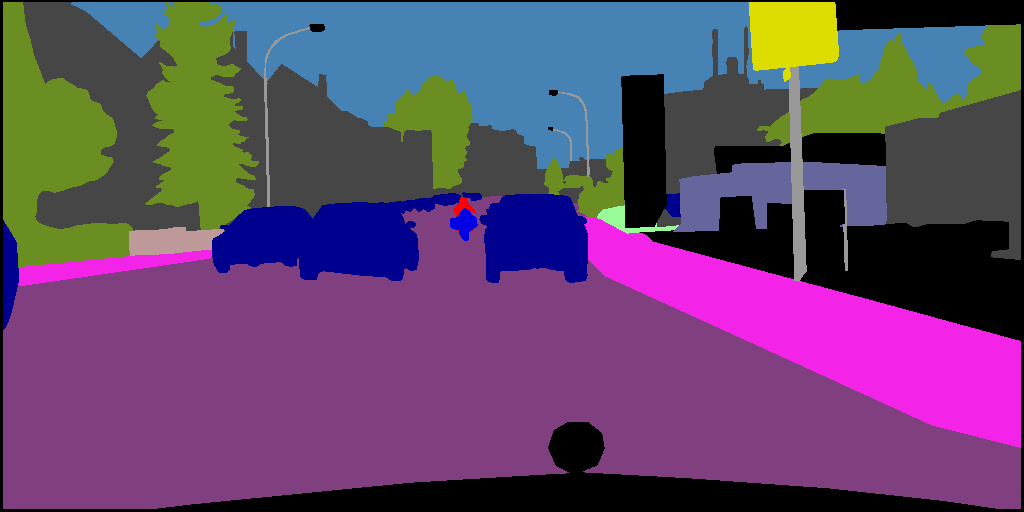}
\end{subfigure}
\begin{subfigure}[t]{0.12\linewidth}
    \includegraphics[width=\linewidth]{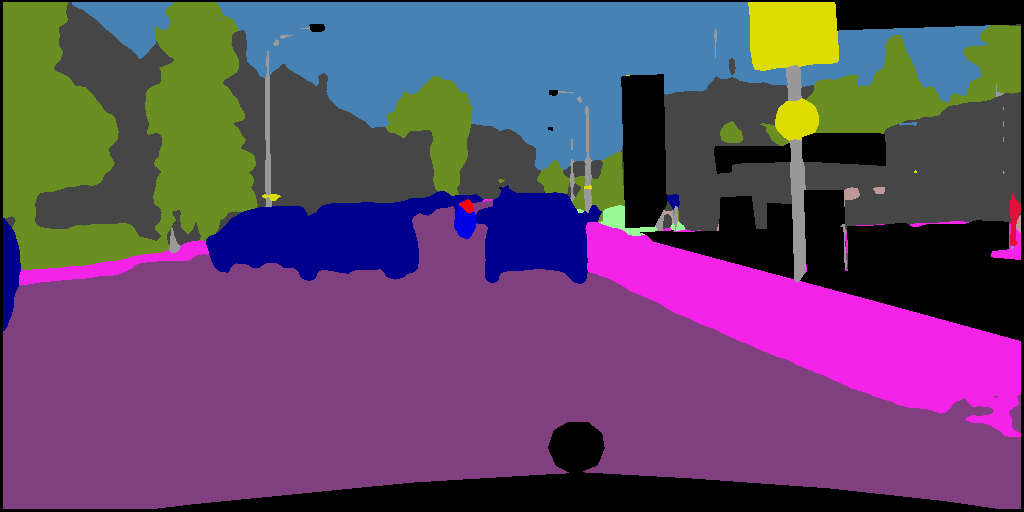}
\end{subfigure}
\begin{subfigure}[t]{0.12\linewidth}
    \includegraphics[width=\linewidth]{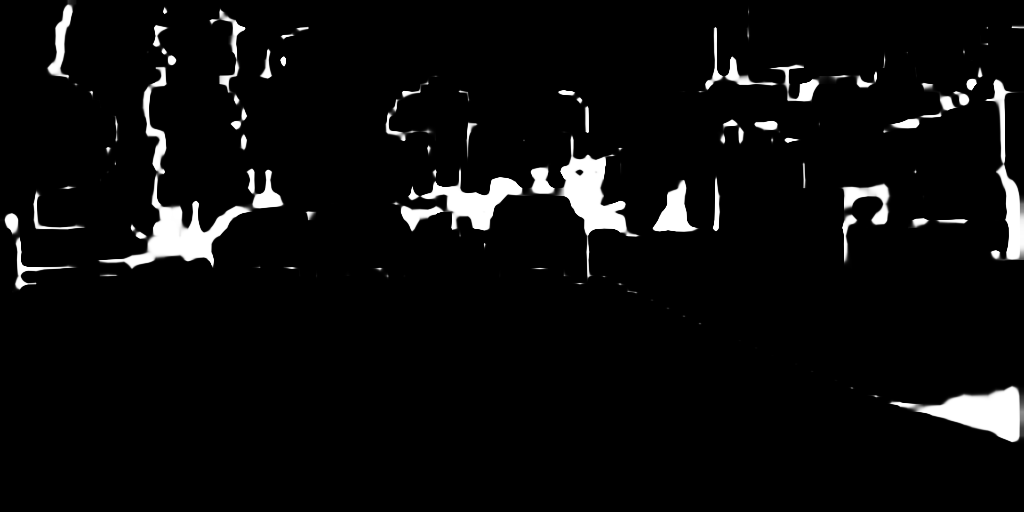}
\end{subfigure}
\begin{subfigure}[t]{0.12\linewidth}
    \includegraphics[width=\linewidth]{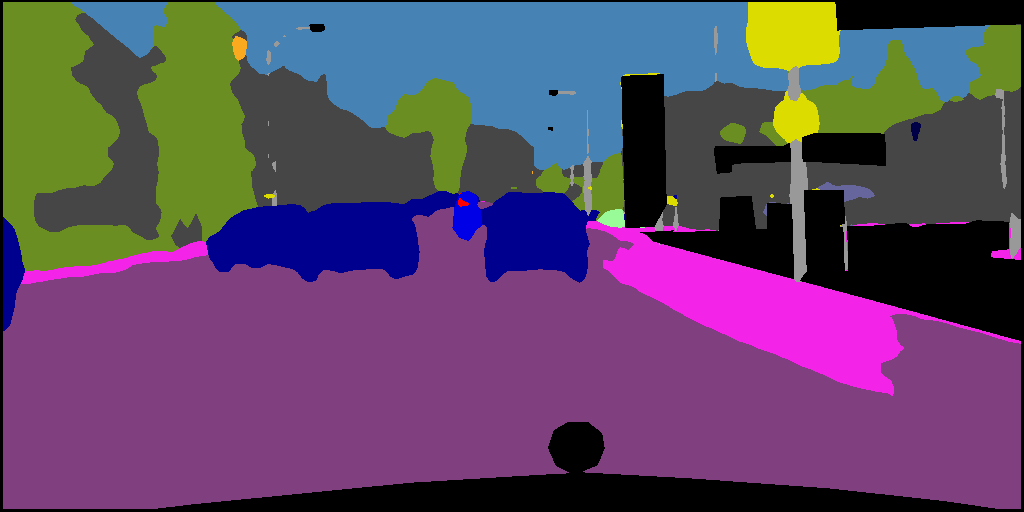}
\end{subfigure}
\begin{subfigure}[t]{0.12\linewidth}
    \includegraphics[width=\linewidth]{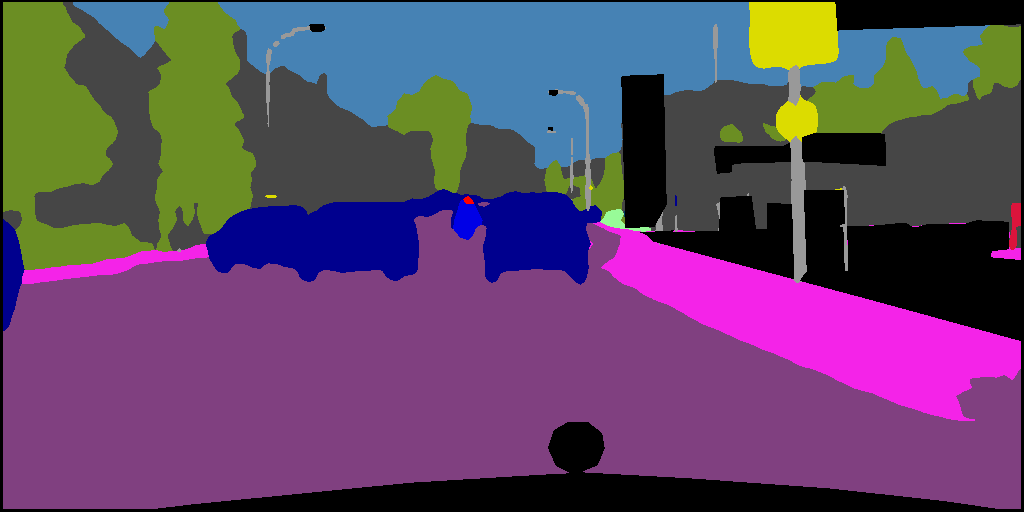}
\end{subfigure}
\begin{subfigure}[t]{0.12\linewidth}
    \includegraphics[width=\linewidth]{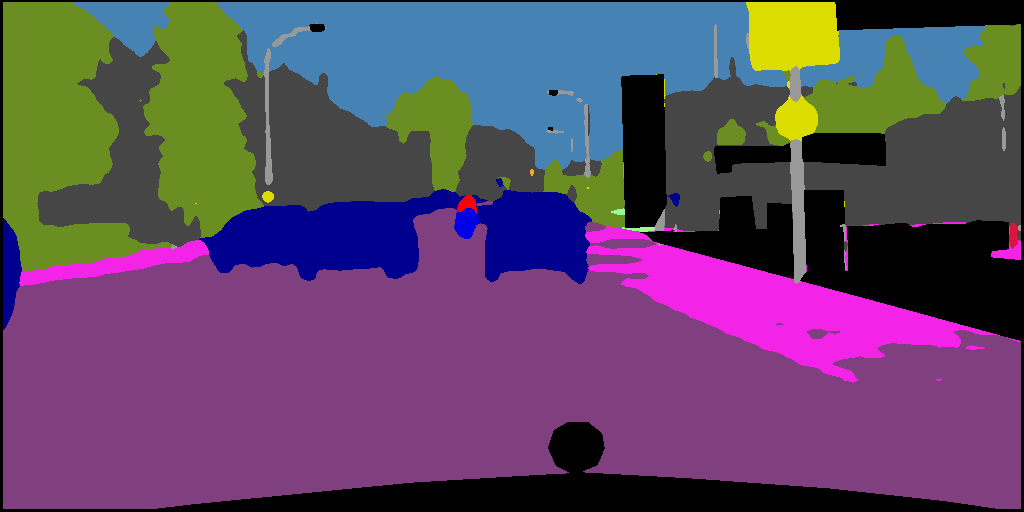}
\end{subfigure}
\begin{subfigure}[t]{0.12\linewidth}
    \includegraphics[width=\linewidth]{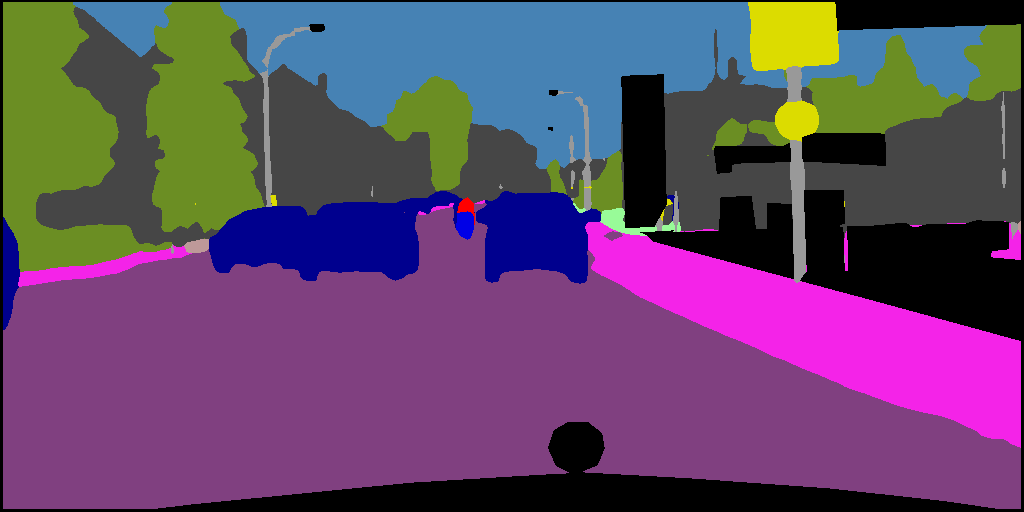}
\end{subfigure}

\caption{From left to right: RGB-image, ground-truth, CE, betting-map, focal loss, CE + adv, EL-GAN, gambling nets. The betting map is a prediction with as input the RGB image and the CE prediction. Results are for the Cityscapes \cite{cordts2016cityscapes} validation set with PSPNet  \cite{zhao2017pyramid}. Best visible zoomed-in on a screen.}
\label{extra_qual_city_psp}
\end{figure}
\clearpage
\begin{figure}[!t]
\centering
\begin{subfigure}[t]{0.12\linewidth}
    \includegraphics[width=\linewidth]{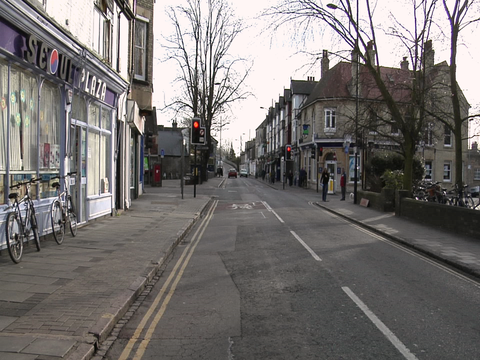}
\end{subfigure}
\begin{subfigure}[t]{0.12\linewidth}
    \includegraphics[width=\linewidth]{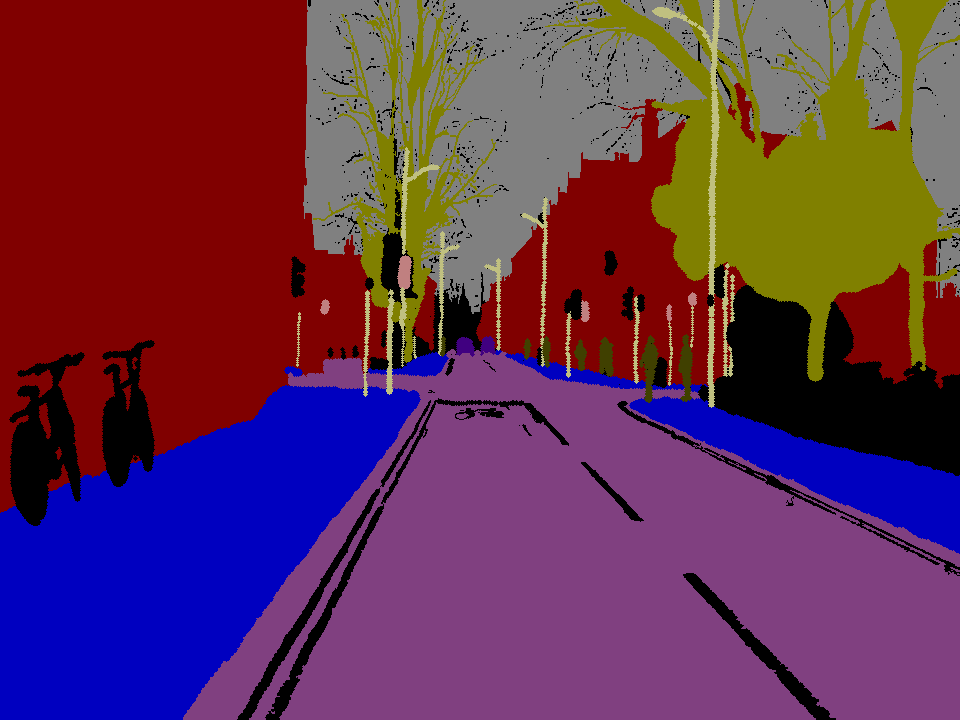}
\end{subfigure}
\begin{subfigure}[t]{0.12\linewidth}
    \includegraphics[width=\linewidth]{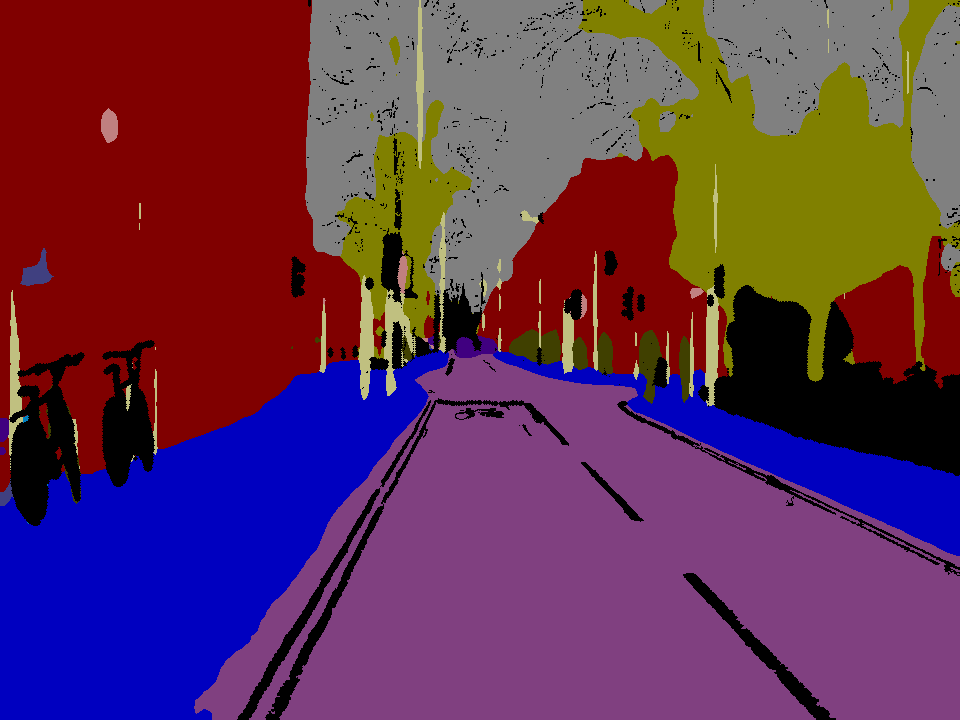}
\end{subfigure}
\begin{subfigure}[t]{0.12\linewidth}
    \includegraphics[width=\linewidth]{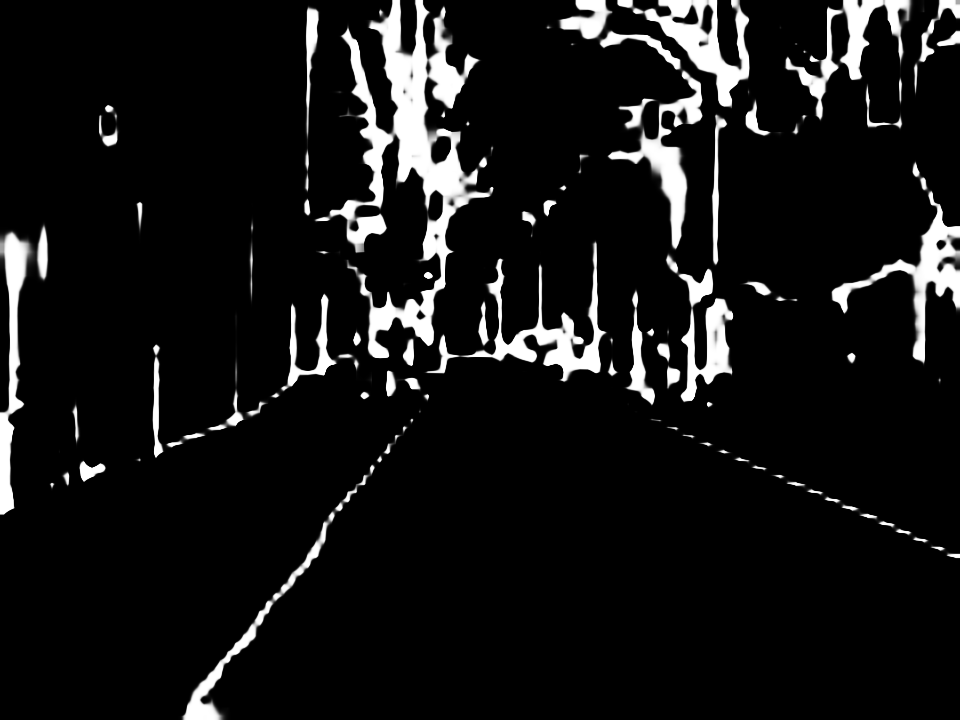}
\end{subfigure}
\begin{subfigure}[t]{0.12\linewidth}
    \includegraphics[width=\linewidth]{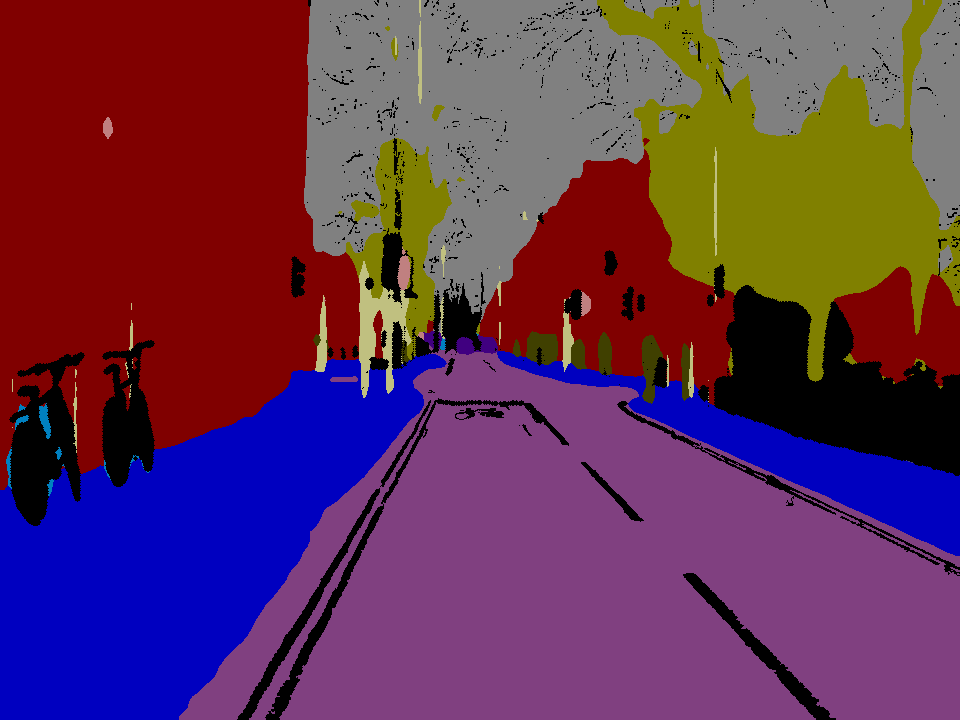}
\end{subfigure}
\begin{subfigure}[t]{0.12\linewidth}
    \includegraphics[width=\linewidth]{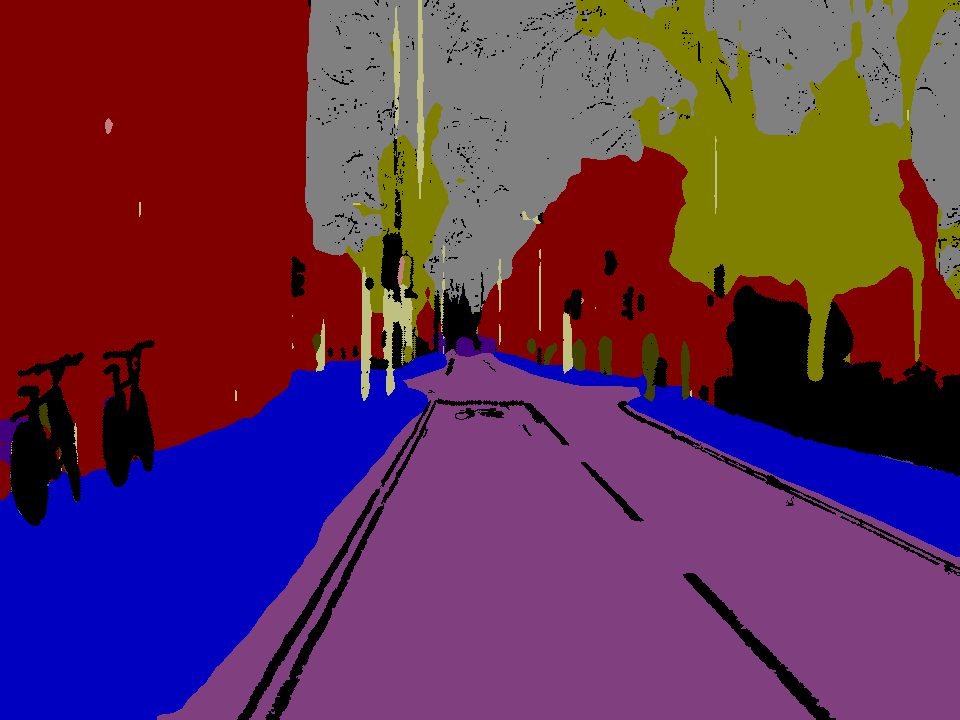}
\end{subfigure}
\begin{subfigure}[t]{0.12\linewidth}
    \includegraphics[width=\linewidth]{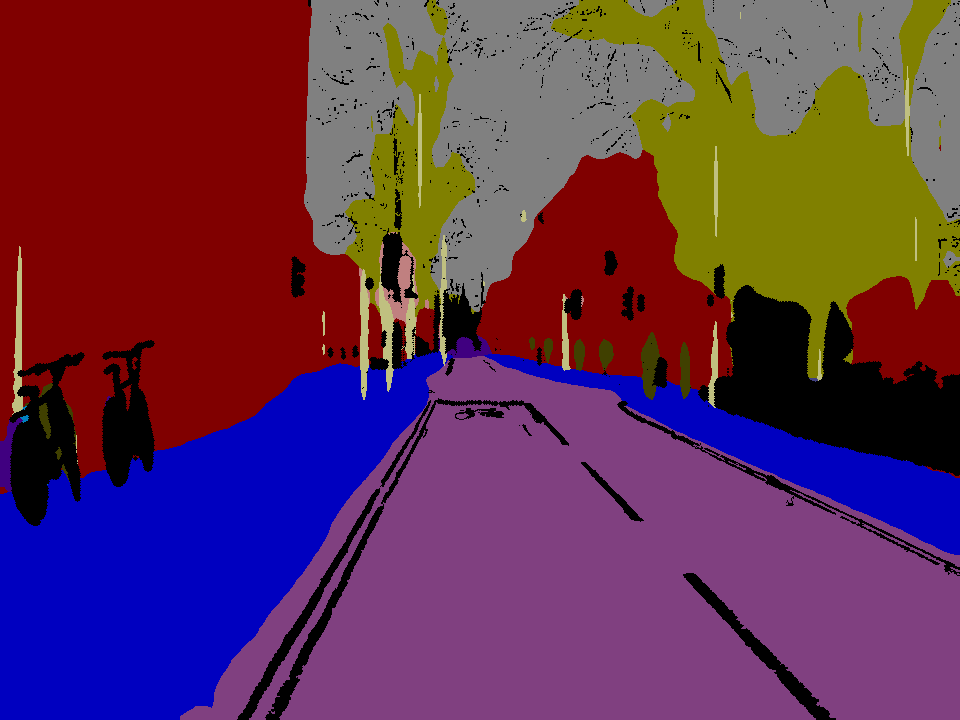}
\end{subfigure}
\begin{subfigure}[t]{0.12\linewidth}
    \includegraphics[width=\linewidth]{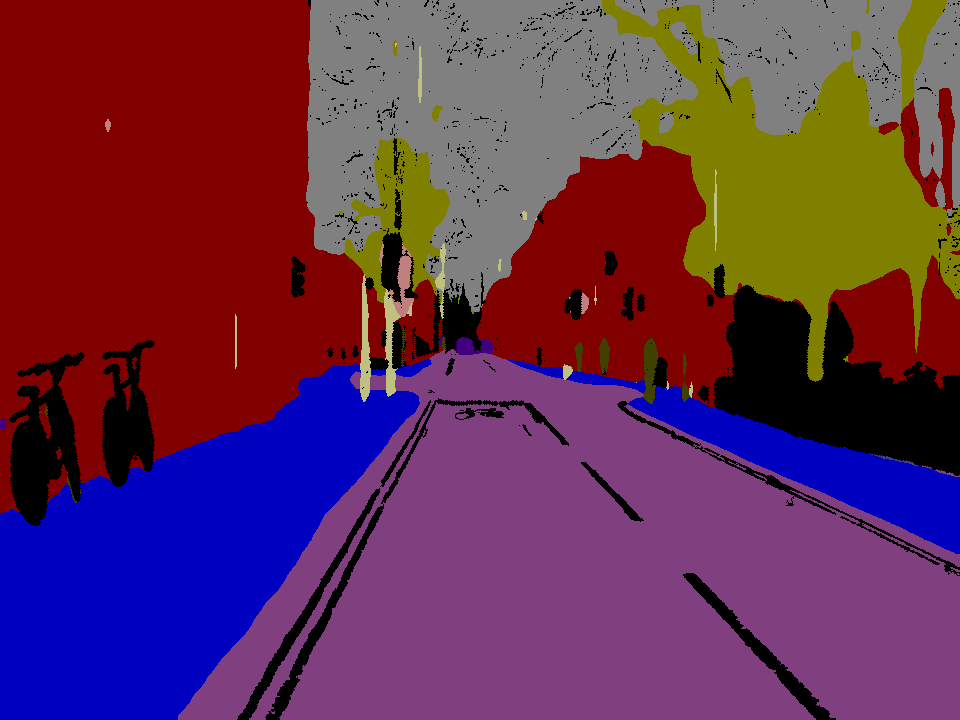}
\end{subfigure}
\begin{subfigure}[t]{0.12\linewidth}
    \includegraphics[width=\linewidth]{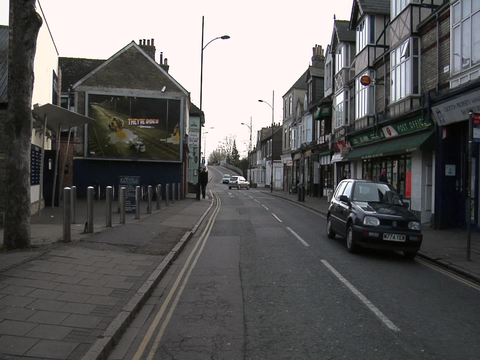}
\end{subfigure}
\begin{subfigure}[t]{0.12\linewidth}
    \includegraphics[width=\linewidth]{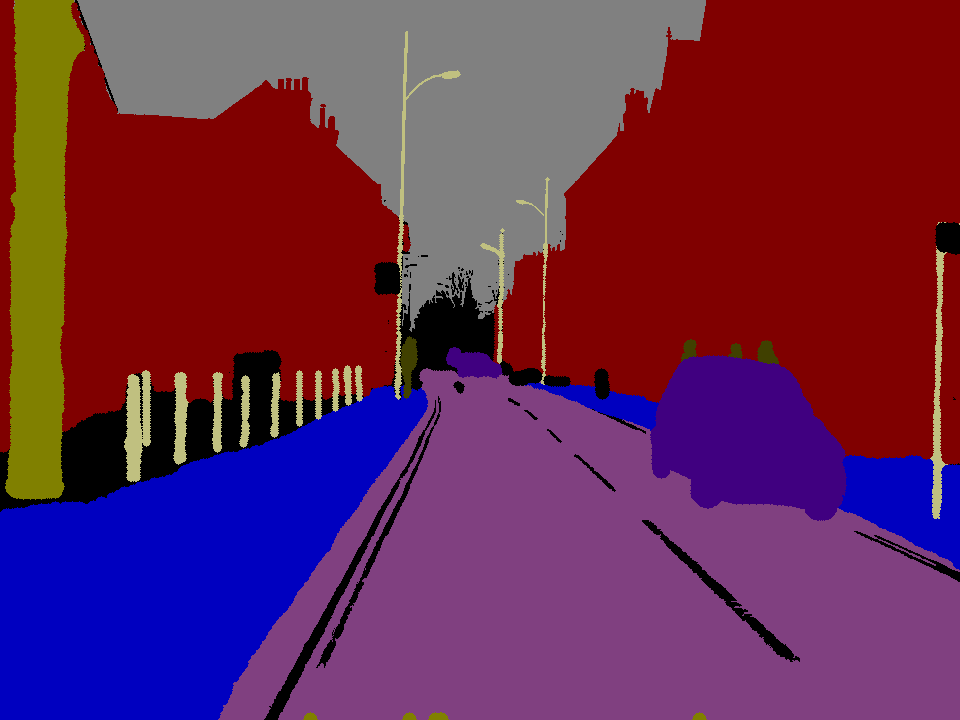}
\end{subfigure}
\begin{subfigure}[t]{0.12\linewidth}
    \includegraphics[width=\linewidth]{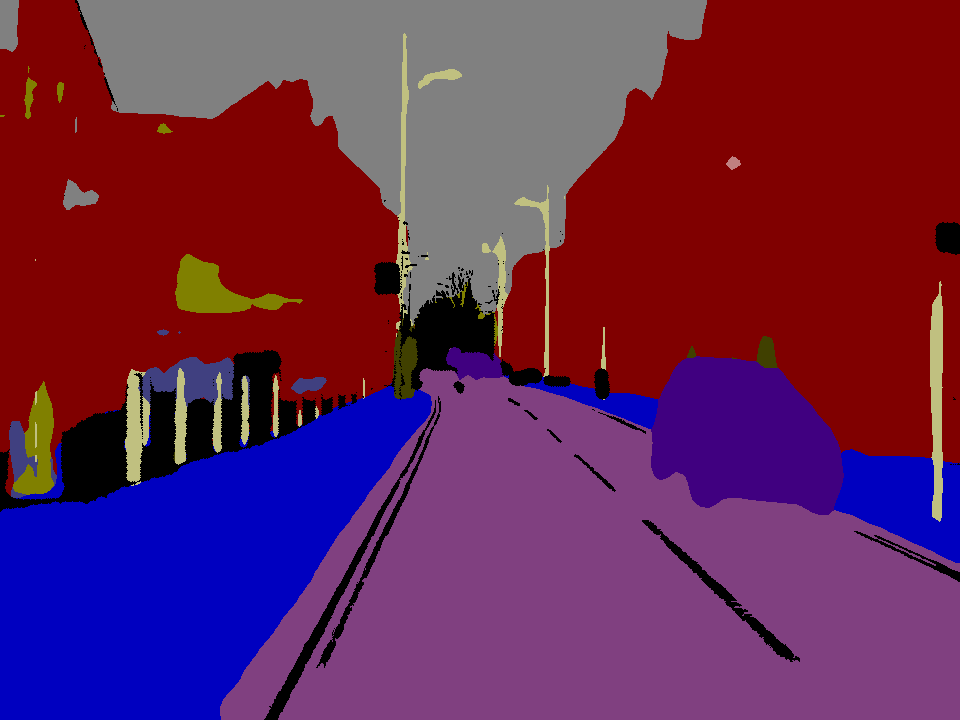}
\end{subfigure}
\begin{subfigure}[t]{0.12\linewidth}
    \includegraphics[width=\linewidth]{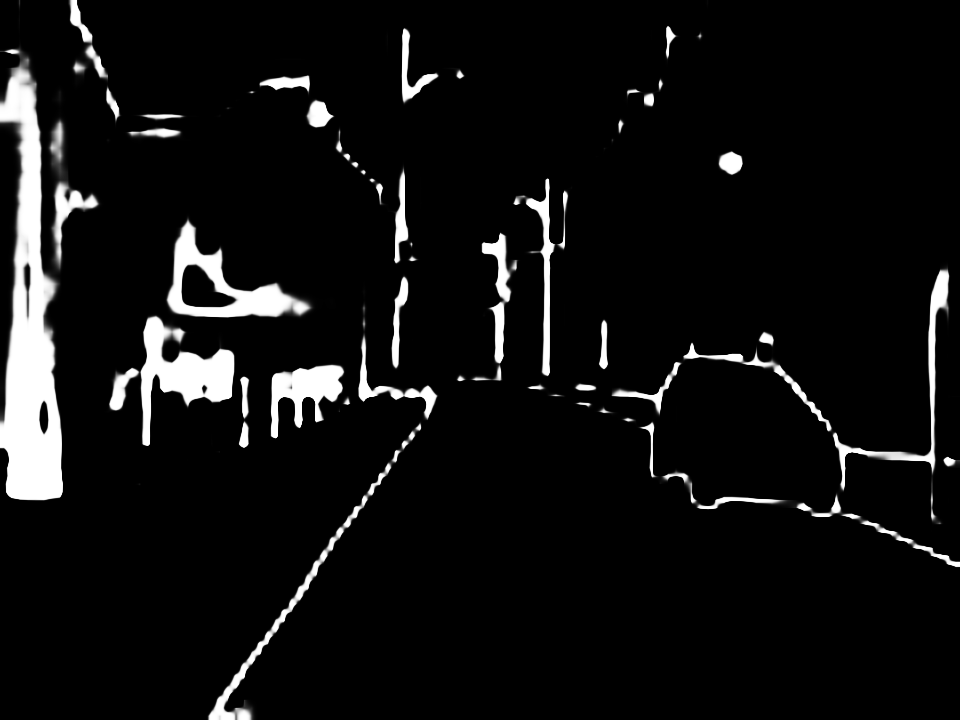}
\end{subfigure}
\begin{subfigure}[t]{0.12\linewidth}
    \includegraphics[width=\linewidth]{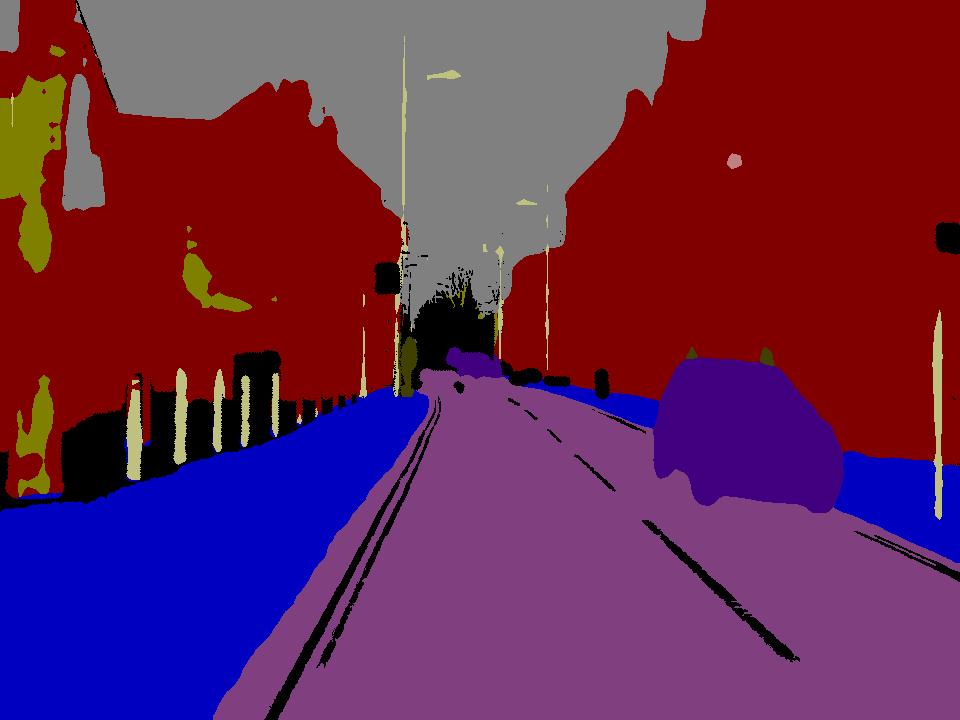}
\end{subfigure}
\begin{subfigure}[t]{0.12\linewidth}
    \includegraphics[width=\linewidth]{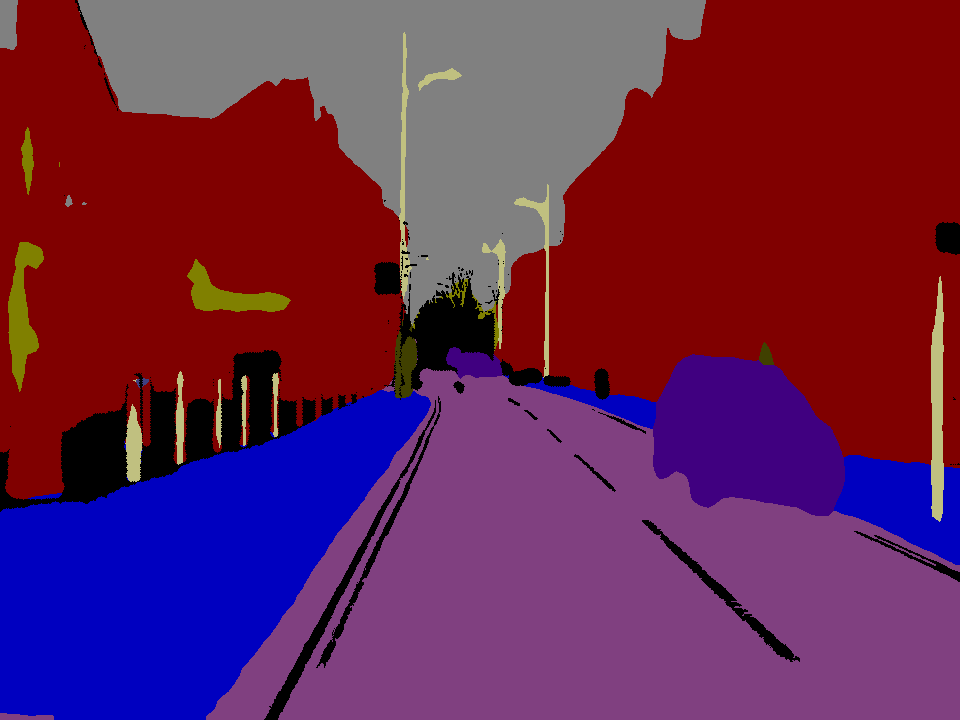}
\end{subfigure}
\begin{subfigure}[t]{0.12\linewidth}
    \includegraphics[width=\linewidth]{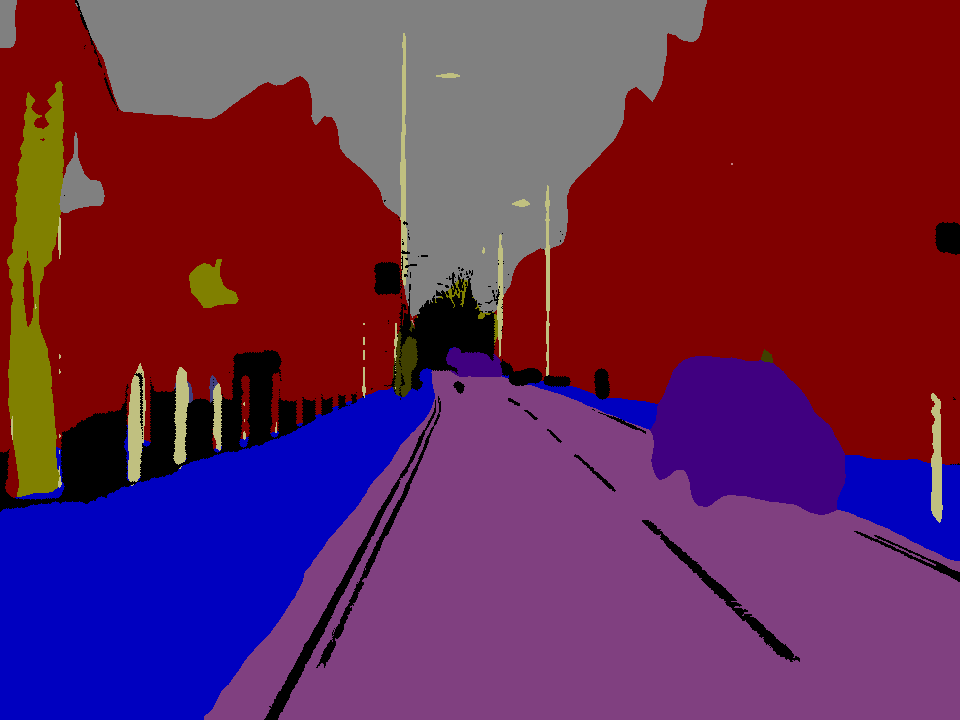}
\end{subfigure}
\begin{subfigure}[t]{0.12\linewidth}
    \includegraphics[width=\linewidth]{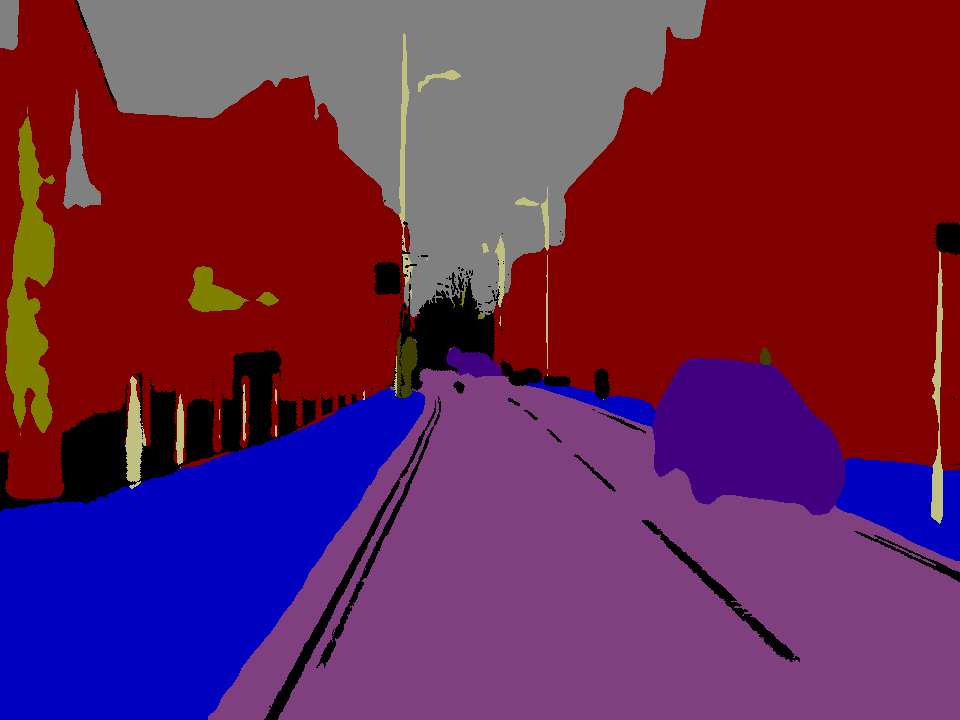}
\end{subfigure}
\begin{subfigure}[t]{0.12\linewidth}
    \includegraphics[width=\linewidth]{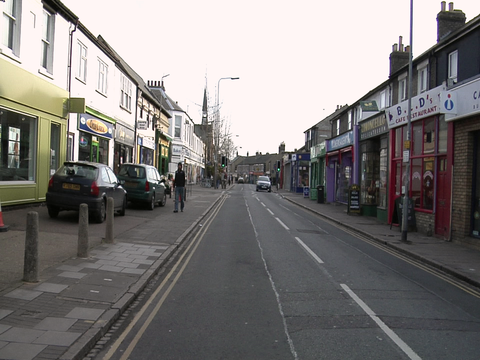}
\end{subfigure}
\begin{subfigure}[t]{0.12\linewidth}
    \includegraphics[width=\linewidth]{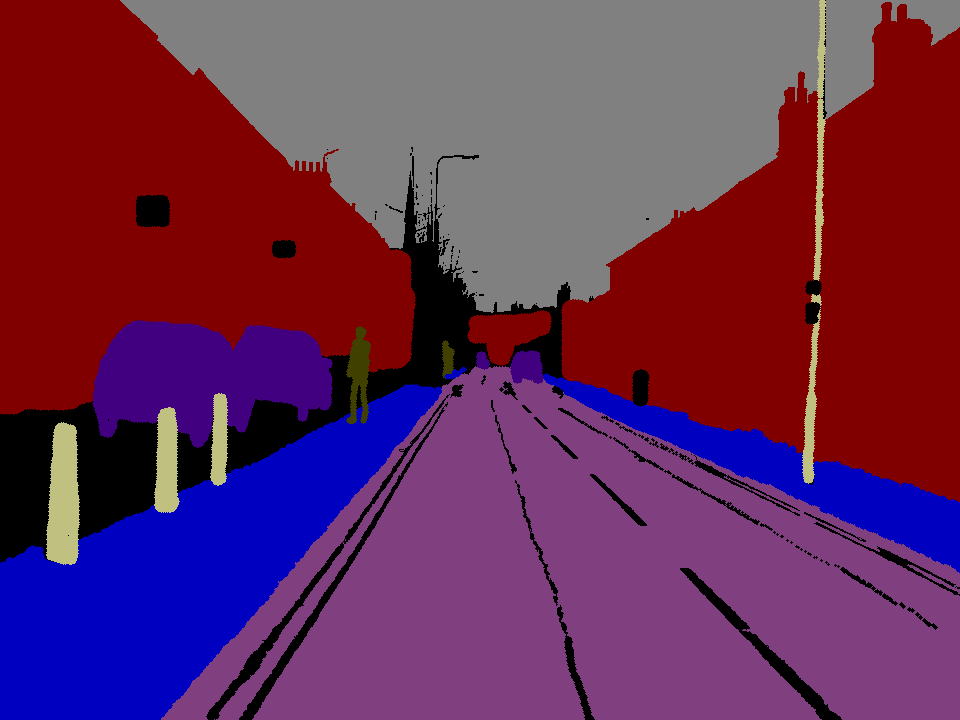}
\end{subfigure}
\begin{subfigure}[t]{0.12\linewidth}
    \includegraphics[width=\linewidth]{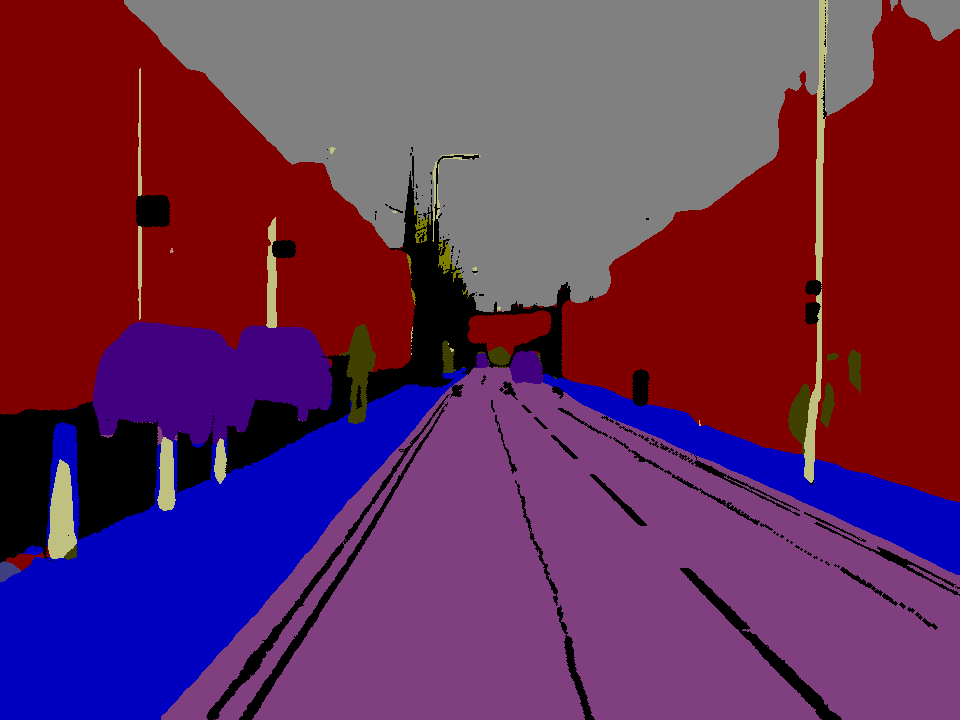}
\end{subfigure}
\begin{subfigure}[t]{0.12\linewidth}
    \includegraphics[width=\linewidth]{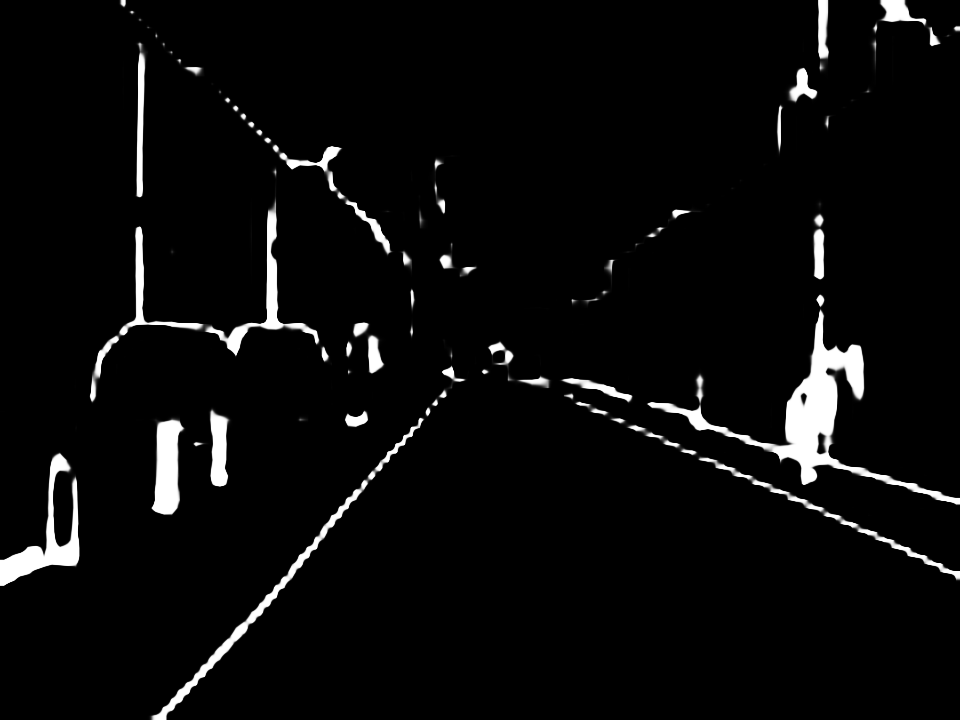}
\end{subfigure}
\begin{subfigure}[t]{0.12\linewidth}
    \includegraphics[width=\linewidth]{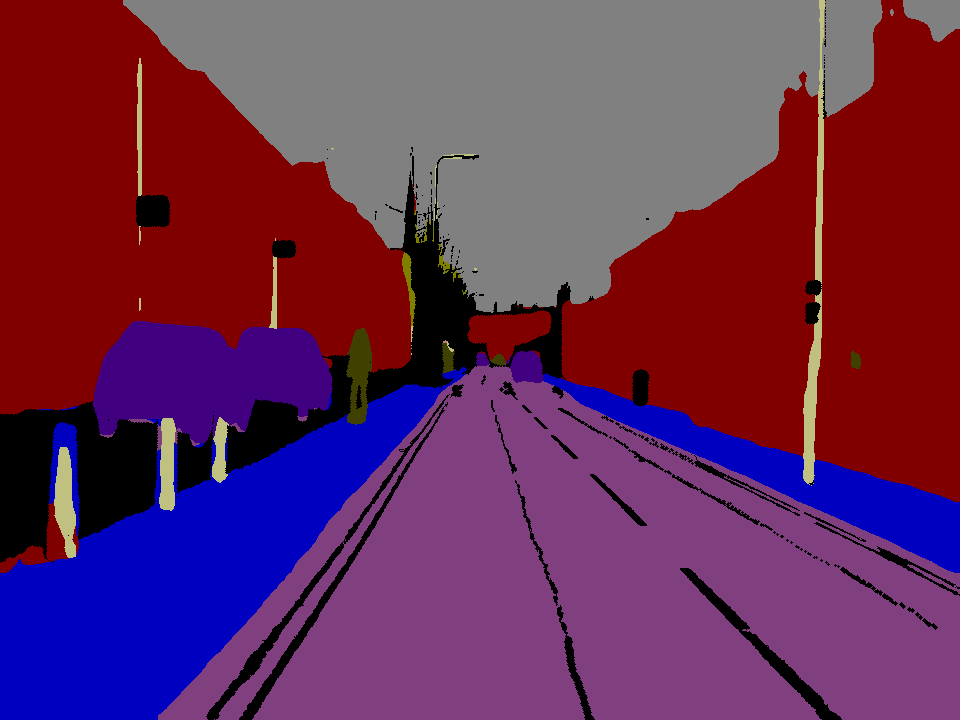}
\end{subfigure}
\begin{subfigure}[t]{0.12\linewidth}
    \includegraphics[width=\linewidth]{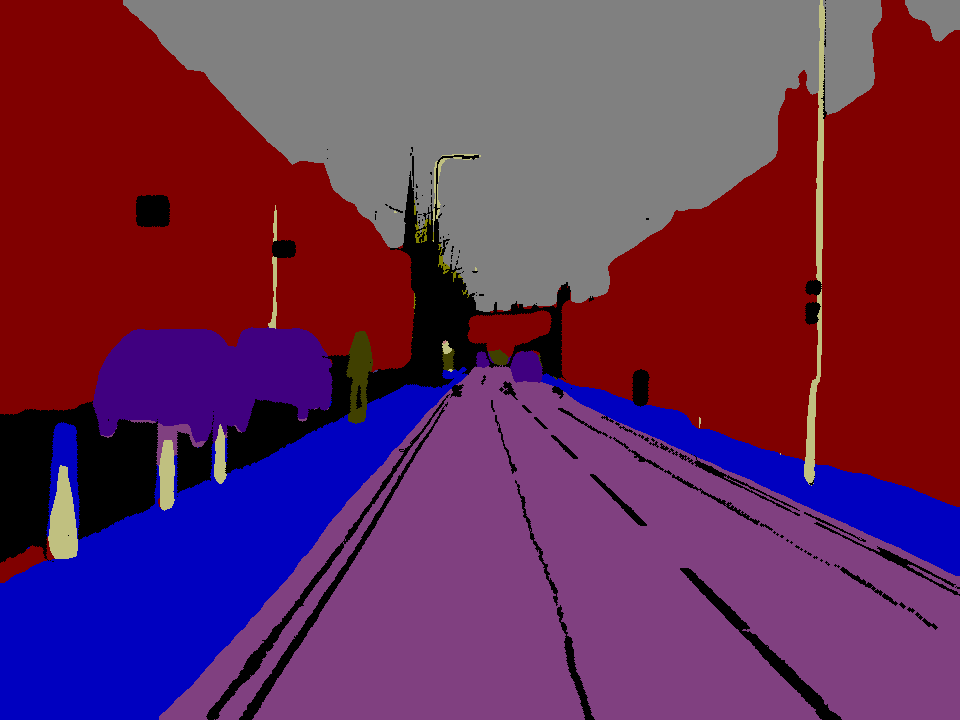}
\end{subfigure}
\begin{subfigure}[t]{0.12\linewidth}
    \includegraphics[width=\linewidth]{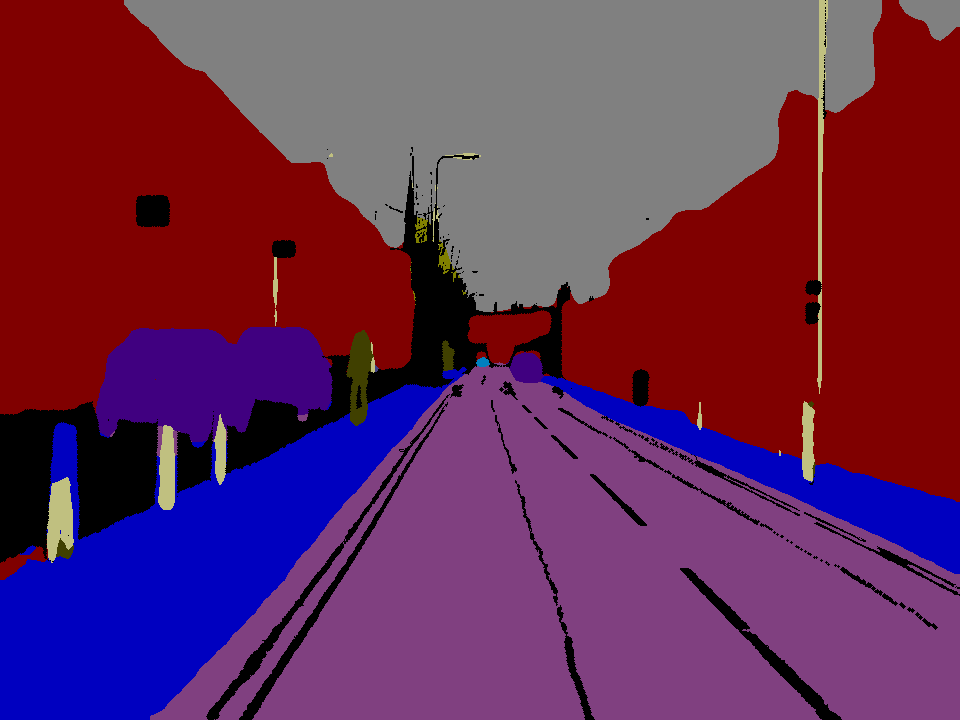}
\end{subfigure}
\begin{subfigure}[t]{0.12\linewidth}
    \includegraphics[width=\linewidth]{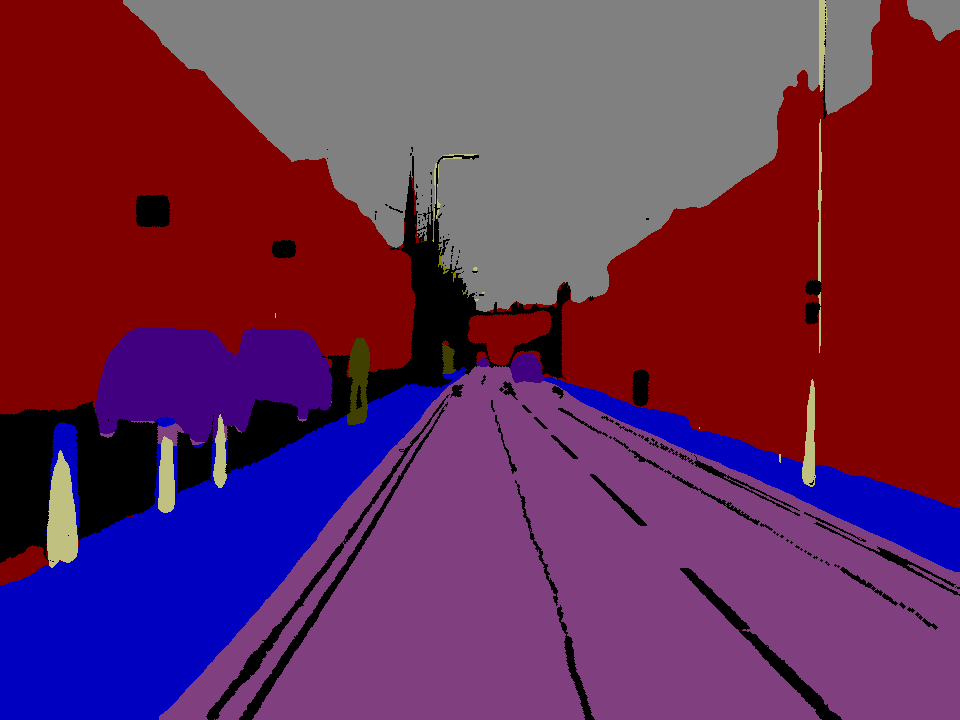}
\end{subfigure}

\caption{From left to right: RGB-image, ground-truth, CE, betting-map, focal loss, CE + adv, EL-GAN, gambling nets. The betting map is a prediction with as input the RGB image and the CE prediction. Results are for the Camvid \cite{cordts2016cityscapes} test set with PSPNet  \cite{zhao2017pyramid}. Best visible zoomed-in on a screen.}
\label{extra_qual_cam_psp}
\end{figure}
\end{landscape}
\nocite{*}
\newsavebox\mytempbib
\bibliographystyle{ieee}
\savebox\mytempbib{\parbox{\textwidth}{\bibliography{egbib}}}

\end{document}